\documentclass[letterpaper]{article} 
\usepackage{aaai2026}  
\usepackage{times}  
\usepackage{helvet}  
\usepackage{courier}  
\usepackage[hyphens]{url}  
\usepackage{graphicx} 
\usepackage{natbib}  
\usepackage{caption} 
\usepackage{subcaption}
\usepackage{algorithm}
\usepackage{algorithmic}

\usepackage{newfloat}
\usepackage{listings}
\DeclareCaptionStyle{ruled}{labelfont=normalfont,labelsep=colon,strut=off} 
\floatstyle{ruled}
\newfloat{listing}{tb}{lst}{}
\floatname{listing}{Listing}
\title{Consistency-based Abductive Reasoning over Perceptual Errors of Multiple Pre-trained Models in Novel Environments}
\author{
	Mario Leiva\textsuperscript{\rm 1},
	Noel Ngu\textsuperscript{\rm 2},
	Joshua Shay Kricheli\textsuperscript{\rm 3},
	Aditya Taparia\textsuperscript{\rm 2},
	Ransalu Senanayake\textsuperscript{\rm 2}, \\
	Paulo Shakarian\textsuperscript{\rm 3},
	Nathaniel D. Bastian\textsuperscript{\rm 4},
	John Corcoran\textsuperscript{\rm 5}, and
	Gerardo Simari\textsuperscript{\rm 1}
}
\affiliations{
	\textsuperscript{\rm 1}DCIC, Universidad Nacional del Sur (UNS) \& ICIC (UNS-CONICET), Bah\'ia Blanca, Argentina\\
	\textsuperscript{\rm 2}Arizona State University, Tempe, AZ USA\\
	\textsuperscript{\rm 3}Syracuse University, Syracuse, NY USA\\
	\textsuperscript{\rm 4}United States Military Academy, West Point, NY USA \\
	\textsuperscript{\rm 5}Systems Planning \& Analysis, Alexandria, VA USA \\	
		
mario.leiva@cs.uns.edu.ar,
\{nngu2, ataparia, ransalu\}@asu.edu,
\{jkrichel, pashakar\}@syr.edu,
nathaniel.bastian@westpoint.edu,
jack.fd.corcoran@gmail.com, 
gis@cs.uns.edu.ar
}

\usepackage{bibentry}

\usepackage{amsmath}
\usepackage{booktabs}
\usepackage{multirow}

\def\pred{\mathit{pred}}

\newcommand{\supplementaryfigure}[3]{%
    \begin{figure}[h!]
        \centering
        \includegraphics[width=1\textwidth]{#1} 
        \caption{#2}
        \label{#3}
    \end{figure}
}
\begin{document}

\maketitle

\begin{abstract}
The deployment of pre-trained perception models in novel environments often leads to
performance degradation due to distributional shifts. Although recent artificial intelligence approaches for metacognition use logical rules to characterize and filter model errors, improving
precision often comes at the cost of reduced recall. This paper addresses the 
hypothesis that leveraging multiple pre-trained models can mitigate this recall 
reduction. We formulate the challenge of identifying and managing conflicting
predictions from various models as a consistency-based abduction problem, building on the idea of abductive learning (ABL) but applying it to test-time instead of training. The input predictions and the learned error detection rules derived from each model are encoded in a logic program. 
We then seek an abductive explanation---a subset of model
predictions---that maximizes prediction coverage while ensuring the rate of logical
inconsistencies (derived from domain constraints) remains below a specified
threshold. We propose two algorithms for this knowledge representation task: an exact method based on 
Integer Programming (IP) and an efficient Heuristic Search (HS). Through extensive 
experiments on a simulated aerial imagery dataset featuring controlled, complex 
distributional shifts, we demonstrate that our abduction-based framework outperforms
individual models and standard ensemble baselines, achieving, for instance, average relative improvements of approximately 13.6\% in F1-score and 16.6\% in accuracy across 15 diverse test datasets when compared to the best individual model. 
Our results validate the use of consistency-based abduction as an effective mechanism to robustly integrate knowledge from multiple imperfect models in
challenging, novel scenarios.
\end{abstract}

\begin{links}
    \link{Code}{github.com/lab-v2/EDCR_PyReason_AirSim}
    \link{Extended version}{https://arxiv.org/abs/2505.19361}
\end{links}

\section{Introduction}
\label{sec:introduction}

The use of pre-trained models is very common in tasks that require perception data, such as classification and object detection in images and video~\cite{han2021pre,parisi2022unsurprising}.
Another scenario in which differences arise is when we know we will be deploying in different environments because we are using the models that we have available---we refer to these issues as {\em deployment in novel environments}. 
As a specific example, consider emergency response, where perception models examining a disaster must contend with unforeseen environmental changes even when trained on data for a similar region.  Another example is an NGO providing aid to a remote location where training data was unavailable.  In both cases, we can be assured that the environment in which the perception models operate is \textit{novel} with respect to what they were trained on.

Psychologists have shown that humans deal with novelty through metacognition~\cite{thompson2009dual} by leveraging the dual Type 1 / Type 2 processing~\cite{evans13} (i.e., ``dual process theory''~\cite{WASON1974141} popularized by \citep{kahneman_thinking_2012}).  In particular, a collection of autonomous ``Type 1'' systems perceives information that may also lead to a ``metacognitive cue'' that triggers additional ``Type 2'' reasoning.  Following the renewed interest in metacognitive artificial intelligence (AI)~\cite{wei24,johnson2024imaginingbuildingwisemachines}, recent work has shown that we can learn rules that provide metacognitive cues about machine learning model failure~\cite{kricheli2024error}---that work, however, uses a single model and does not provide further reasoning resulting from the metacognitive cue.  Meanwhile, work on abductive learning~\cite{dai2019bridging} allows for adjustment to a single machine learning model at training time based on abductive feedback.  In this work, we use cues provided by logic programs modeling the failures of multiple models in an abductive framework at inference time to allow for enhanced perception in novel environments.

\begin{figure}[t]
    \centering
    \includegraphics[width=\linewidth]{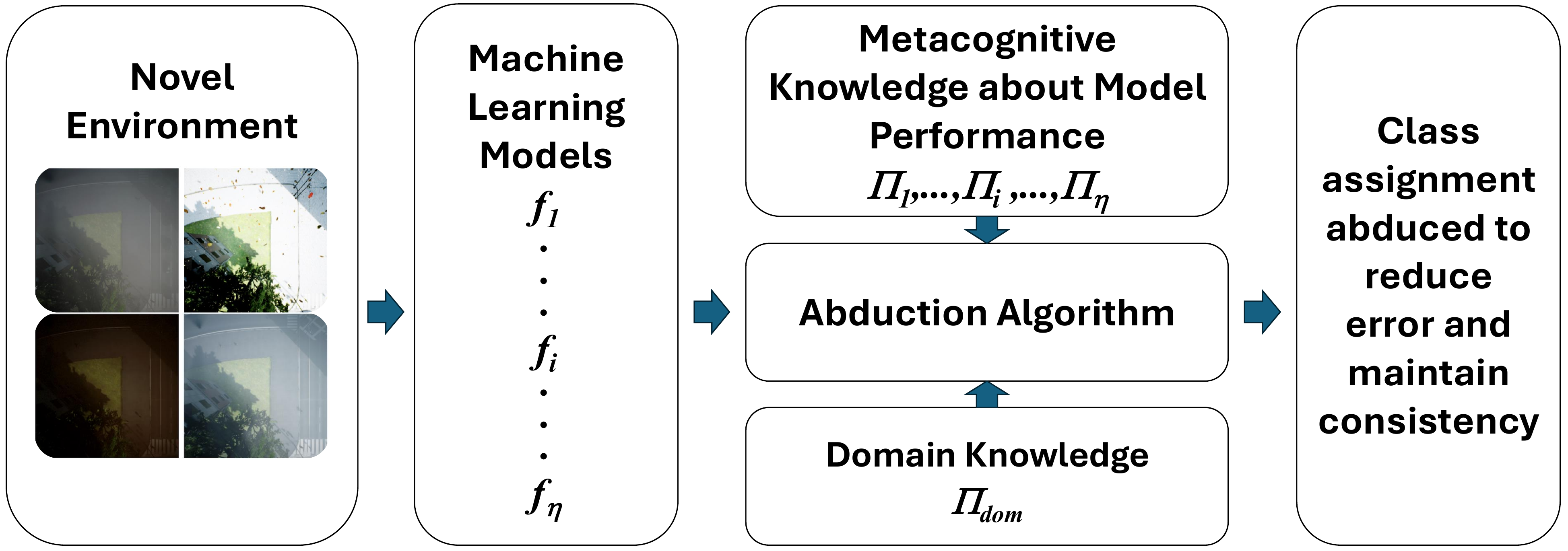}
    \caption{Overview of our Consistency-based Abductive Reasoning Approach. Here $\eta$ machine learning models perceive a novel environment.  Their results are considered with domain knowledge and metacognitive information about the models (learned independently and with the same training data) to abduce a set of results that exhibit consistency and reduce perceptual errors.}
    \label{fig:intro}
\end{figure}

In this work, our working hypothesis is that by deploying more than one model we are able to at least partially address this drawback; this is the same underlying principle behind standard approaches in machine learning for developing ensembles of models, but our approach goes beyond such standard practices since we apply novel metacognitive AI techniques.
In particular, as shown in Figure~\ref{fig:intro}, we leverage existing rule learning techniques to derive a logic program consisting of metacognitive rules across a set of perception models, and frame the task of identifying errors across all models as a {\em consistency-based abduction problem}.
We then show that such error identification problems can be posed as integer programs, and provide a scalable heuristic algorithm to solve the abductive reasoning task. Noteworthy in our approach is that the logic program is created by rules learned for each perceptual model based on their training data---so there is no a priori knowledge of test data (i.e., no leakage). Further, the rules for the individual models are learned independently from each other, so we assume there is no existing knowledge of how the models perform together.

Finally, we present a thorough experimental evaluation using an extended, highly-controlled aerial imagery dataset with diverse distributions~\cite{ngu2025multipledistributionshift}. Our results demonstrate that our abductive-based reasoning approach---by effectively managing inconsistencies while maximizing predictions---achieves superior performance compared to individual models across numerous test datasets.

\section{Related Work}
\label{sec:related_work}

This work is closely related to~\cite{xi2024rulebasederrordetectioncorrection}, where Error Detection and Correction rules are introduced. The main difference with our approach is that they only consider a single model. As we discuss below, here we only leverage error detection rules, though the approach can easily accommodate the use of correction rules as well.

Abductive learning (ABL)~\cite{dai2019bridging} also leverages abduction to reduce perception errors and, like our approach, relies on some domain knowledge (for instance, we require knowledge about consistency of predictions).  However, ABL uses this to improve performance based on model training and assumes that the test environment is not entirely novel. In this work, we relax both assumptions and use abduction only at test time.  We note that a follow-on study, called ``ABL with new concepts'' ($\textit{ABL}_{\textit{nc}}$) \cite{dai2023_newconcepts} extends ABL---in a manner similar to EDCR---with the ability to identify previously ``unknown'' concepts, which EDCR~\cite{xi2024rulebasederrordetectioncorrection} was also shown to have.  Like ABL, this approach is also focused on using abduction at model training and not inference.  In this work, we do not extend the concept scheme (though approaches like ABL/$\textit{ABL}_{\textit{nc}}$, EDCR, and HDC could all potentially be helpful) but rather change the distribution of perceptual data by which such concepts are extracted.
We note that early work on abduction often focused on the diagnosis of errors and faults~\cite{poole89}; however, to our knowledge this has not been applied to perception tasks at test time.  This early work inspired our use of abduction to identify perception errors.

Recently, a concept known as ``test-time training'' (TTT) \cite{sun19ttt} has gained traction in the machine learning literature and has shown importance in reasoning tasks as well~\cite{ttt2024}. In this approach, the neural model is trained in a manner such that it performs self-supervised training during test time to improve inference.  While this shows ability to improve out-of-distribution results, it is also a method designed to improve model training, and was noted to have limitations based on the classes used in the self-supervised training.  We note that such a TTT model could be treated as any other pre-trained model in our framework, and we can even envision different variants of TTT (e.g., with different strategies for the self-supervised portion) better working together by leveraging our results.

Finally, in~\cite{sutor2022gluing}, the authors also explore combining a set of pre-trained models using hyperdimensional computing (HDC). However, their method relies on having a set of training samples from the same target distribution as the test set, enabling the joint training of an HDC ``gluing'' module. In contrast, here we assume that data from the same distribution as the test set is unavailable and that the models are trained independently. Furthermore, their method depends on the output layers of the neural networks and lacks explainability, unlike our rule-based approach.

\section{Consistency-based Abduction Problem}
\label{sec:problem_definition}

\smallskip
\noindent\textbf{Technical Preliminaries.}
We consider the problem of object identification for a given set of perception data $\Omega$~\cite{cheng2016survey} and assume the availability of
a set $\eta$ of perception models $\mathcal{F} = \{f_1, f_2, \dots, f_\eta\}$, where each model generates a set of predictions over a shared set of $m$ object classes $\mathcal{C} = \{c_1, ..., c_m\}$.  Under the unique name assumption\footnote{See Section~1 of the appendix in the extended version for the implementation details of how we employ this assumption.}, every object $\omega \in \Omega$ identified by one or more models has at least one associated fact of the form $f_i(\omega)=c_j$, meaning that model $i$ has identified object $\omega$ as belonging to class $c_j$.  We will denote the set of these facts (hereafter ``observations'') as $O$.

We note that, as we are working with multiple models, there may be differences in which class is assigned a particular object.  Further, some of the models may make mistakes.  Hence, we introduce the predicate $\textit{accept}$ where for some model $f_i$ and class $c$, $\textit{accept}(i,c)$ is true if we wish to accept model $f_i$'s results when it returns class $c$.  We denote the set of all acceptance atoms with $\mathcal{H}$, and typically refer to a subset of this set as a ``hypothesis'', denoted $H$.  
Our goal is to find a set $H$ that meets certain criteria (to be described).

A key piece of our framework is leveraging the ability to identify metacognitive cues.  As such, we assume that each model $f_i$ has an associated logic program $\Pi_i$ consisting of rules that, when fired, provide such metacognitive cues about errors.  These logic programs can be learned from the model training data as per prior work~\cite{kricheli2024error,xi2024rulebasederrordetectioncorrection} and are of the following form:
\begin{eqnarray*}
    \textit{error}(i,c,\omega) \leftarrow (f_i(\omega)=c) \wedge \textit{cue}(\omega)
\end{eqnarray*}
Intuitively, if metacognitive cue is present for object $\omega$ and it was classified as class $c$ by model $f_i$, then we can assume that there was an error made in the assignment.  In \cite{kricheli2024error}, the metacognitive cues (referred to as ``conditions'' in that work) are selected from a set of candidate conditions.  We include details of how we applied their learning algorithm in extended version.

In addition to each of the $\Pi_i$'s, we assume a logic program $\Pi_\textit{helper}$ that consists of rules of the following form:
\[
\small
\textit{assign}(c,\omega) \leftarrow \neg \textit{error}(i,c,\omega) \wedge (f_i(\omega)=c) \wedge \textit{accept}(i,c)
\]
In other words, if we accept the results of a given class from a model and no error is reported for that model-class pair, then we can assign object $\omega$ class $c$. 
Another set of rules $\Pi_\textit{dom}$ specifies domain knowledge.  In this work, we are primarily concerned about integrity constraints that prevent a given object from being classified with conflicting classes.  In this paper, rules in $\Pi_\textit{dom}$ are of the form:
\begin{eqnarray}
    \label{rule-dom}
\neg \textit{assign}(c',\omega) \leftarrow \textit{assign}(c,\omega)
\end{eqnarray}
We use symbol $\Pi$ to denote 
$\Pi_\textit{dom} \cup \Pi_\textit{helper} \cup (\bigcup_i\Pi_i)$.  
Further, leveraging simple stratification and the limited use of negations of $\textit{error}$ and 
$\textit{assign}$ predicates, we can implement this in a tractable monotonic logic.

\smallskip
\noindent\textbf{Abduction Problem.}  We build a consistency-based abduction problem~\cite{eiter95} whereby a set of hypotheses $H$ is found that is consistent with a set of observations $O$ and domain knowledge $\Pi$.  
We use logic programs as our representation language (specifically, the framework of~\cite{aditya2023pyreason}, though others such as Prolog and Datalog are possible).  
Here, $H$ and $O$ are expressed as atomic facts, while $\Pi$ is a set of logical rules.  For some universe of hypothesis atoms $\mathcal{H}$, we wish to find $H \subseteq \mathcal{H}$ such that $H\cup O \cup \Pi$ is consistent.

From a practical perspective, we may wish to allow for some (small) amount of inconsistency in the resulting output with respect to $\Pi_\textit{dom}$.  For a given hypothesis $H$, we define $\textit{Inc}(H)$ as the normalized number of ground rules in $\Pi_\textit{dom}$ not entailed by $(H \cup O \cup \Pi) \setminus \Pi_\textit{dom}$. 
Given the rule format for $\Pi_\textit{dom}$ (cf.\ Eq.~\ref{rule-dom}), computing this is trivial (doing so in other languages is left to future work). We use the symbol $\delta \in [0,1]$ to denote an acceptable threshold for this value.

As the set of possible solutions is large, and many such solutions could be consistent with $O,\Pi\setminus\Pi_\textit{dom}$, we employ a notion of \textit{parsimony} as usual with abduction~\cite{reggia91}.  Our parsimony function will maximize the number of atoms of the form $\textit{assign}(c,\omega)$ entailed by the minimal model of 
$(H \cup O \cup \Pi) \setminus \Pi_\textit{dom}$; we denote this quantity $\textit{Pred}(H)$.  
The reason for maximization (as opposed to minimization) is threefold: 
$(i)$~by maximizing assigned output, we minimize the suspected errors (as the models are presumably well trained);
$(ii)$~we generally seek to maximize recall; and 
$(iii)$~the construction of rules $\Pi_\textit{helper}$ minimizes the impact of over-assignment of model results (as assignment occurs based not only on acceptance, but having at least one model perceive the object as belonging to the class and that it does so error-free).  With this in mind, we can frame the following optimization problem:
\[
\max_{H\in\mathcal{H}} \ \textit{Pred}(H)
\]
subject to:
\[
\textit{Inc}(H) \leq \delta, \quad \delta \in [0, 1]
\]
and
\[
(H \cup O \cup \Pi) \setminus \Pi_\textit{dom} \quad\text{is consistent}
\]
We again note that $\delta$ can be used to gauge the amount of inconsistency with domain knowledge ($\Pi_\textit{dom}$) as opposed to inconsistency with metacognitive knowledge ($\bigcup_i \Pi_i$).  There could be variance in the amount of metacognitive cues triggered by a given $\Pi_i$, and a user may want to vary this as well.  Fortunately, the approach we used for metacognitive rule learning~\cite{kricheli2024error,xi2024rulebasederrordetectioncorrection} provides an intuitive hyperparameter (denoted with $\epsilon$) that can readily be interpreted as the expected reduction in recall experienced by disregarding erroneous predictions.  
We examine varying $\epsilon$ as well as using a heuristic to set it automatically.

\smallskip
\noindent\textbf{Integer Program (IP) Formulation.}  To solve our associated optimization, we provide an exact integer programming solution and heuristic algorithm.  We first review the integer program. The goal of the IP is to find an optimal hypothesis $H$---represented by a set of binary decision variables---that maximizes the total number of entailed assignments ($\textit{Pred}(H)$) while ensuring the inconsistencies with our domain knowledge ($\textit{Inc}(H)$) remain below the threshold $\delta$. 

Formally, this can be expressed as
\[
\max_{\textit{}} \sum_{\omega \in \Omega} \sum_{c \in \mathcal{C}} A_{c,\omega},
\]
subject to:
\[
\sum_{\omega \in \Omega} \sum_{(c, c^{\prime}) \in \textit{IC}} \textit{Con}_{\omega,(c,c^{\prime})} \leq \delta,
\]
where
$A_{c,\omega}$ is a binary variable indicating whether object $\omega$ is assigned to class $c$,
$\textit{Con}_{\omega,(c,c^{\prime})}$ is a binary variable indicating a conflict between assignments $c$ and $c^{\prime}$ for object $\omega$.

We further define the following constants and variables:
$\pred_{f,c,\omega}$ is a constant set to 1 if $(\omega, c) \in \textit{assigns}$ and $\pred_c^f(\omega) \in \Gamma^*(\Pi)$, 0 otherwise; variable $X_{\omega,f,c} \in \{0, 1\}$ indicates whether object $\omega$ is considered for model $f$ and 
class~$c$; and variable $\textit{Elim}_{f,c} \in \{0, 1\}$ indicates whether predictions from model~$f$ for class~$c$ are excluded; this last variable directly implements our choice of hypothesis $H$. Setting  $\textit{Elim}_{f,c}=0$ is equivalent to including the atom $\textit{accept}(f,c)$ in our hypothesis $H$, thereby trusting model $f$ for class $c$. Conversely, setting $\textit{Elim}_{f,c}=1$ excludes it.
We can now present the set of constraints.
First, for each $f, c, \omega$ we have constraints of the form:
\begin{eqnarray}
X_{\omega,f,c} & \leq & 1 - \textit{Elim}_{f,c} \\
X_{\omega,f,c} \cdot \pred_{f,c,\omega} & \leq & A_{c,\omega}
\end{eqnarray}
Next, for each $c, \omega$ we have:
\begin{equation}
A_{c,\omega} \leq \sum_f X_{\omega,f,c} \cdot \pred_{f,c,\omega}
\end{equation}
For each $\omega \in \Omega, (c,c^{\prime}) \in IC$:
\begin{equation}
A_{c,\omega} + A_{c^{\prime},\omega} - 1 \leq \textit{Con}_{\omega,(c,c^{\prime})}
\end{equation}
for each $\omega$, we have:
\begin{equation}
\sum_{c \in \mathcal{C}} A_{c,\omega} \geq 1
\end{equation}
Finally:
\begin{equation}
\sum_{\omega \in \Omega} \sum_{(c,c^{\prime}) \in IC} \textit{Con}_{\omega,(c,c^{\prime})} \leq \delta
\end{equation}
They respectively ensure the elimination of invalid predictions,
consistency between predictions and assignments,
upper bounds on assignments per object,
that conflicts are adequately managed,
that each object is assigned at least one class, and that
the global conflict threshold holds.

The IP formulation described by Equations (3)--(8) translates into a model with a number of variables and constraints dependent on the number of unique objects ($N$), models ($\mathcal{|F|}$), classes ($\mathcal{|C|}$), and integrity constraints ($|\textit{IC}|$). Specifically, the IP model involves decision variables for assignments ($A_{c,\omega}$), conflict indicators ($\textit{Con}_{\omega,(c,c')}$), model-class eliminations ($\textit{Elim}_{f,c}$), and object consideration ($X_{\omega,f,c}$). The total number of variables is primarily driven by the $N \times \mathcal{F} \times \mathcal{C}$ term (associated with $X_{\omega,f,c}$), resulting in an overall count in 
$O(N \cdot |\mathcal{F}| \cdot |\mathcal{C}|)$. 
Similarly, the number of constraints also scales in $O(N \cdot |\mathcal{F}| \cdot |\mathcal{C}|)$.

From a knowledge representation and reasoning perspective, while solving such IP instances is NP-hard in the worst case, the specific structure of our consistency-based abduction problem often lends itself to relatively efficient resolution in practice. Our IP formulation is characterized by \textit{binary decision variables}, a \textit{linear objective function} (maximizing valid assignments), and a set of \textit{linear constraints}. Many of these constraints exhibit a degree of \textit{locality} (e.g., defining conflicts $\textit{Con}_{\omega,(c,c')}$ based on assignments $A_{c,\omega}$ for the same object $\omega$, or linking model-class considerations $X_{\omega,f,c}$ to overall assignments). This structured nature, which demonstrated efficient performance within our specific experimental configuration, often allows for practical solutions to be found within reasonable timeframes for problems of the scale explored in our experiments.

\smallskip
\noindent\textbf{Heuristic Search (HS).}  Our Heuristic Search (HS) approach is detailed in Algorithm~\ref{alg:hs_concise}.
Given the set of all raw model predictions $P_{\textit{raw}}$, an inconsistency threshold 
$\delta$, and a set of EDR $\epsilon$ values to evaluate, the algorithm 
greedily builds a hypothesis $H$ (represented in the algorithm as the set of final predictions $S_{\textit{final}}$) by iterating through model-class pairs $(f,c)$.
For each pair, it evaluates all $\epsilon$, generating a filtered 
prediction set $P_{\textit{new}}$ (via an implicit $\textit{GetFilteredPreds}(f, c, \epsilon, P_{\textit{raw}})$ 
function). It selects the $P_{\textit{new}}$ that, when added to the current solution 
$S_{\textit{final}}$, maximizes the size of the resulting candidate set $S_{\textit{candidate}} = 
S_{\textit{final}} \cup P_{\textit{new}}$, while ensuring that $\textit{ComputeInconsistency}(S_{\textit{candidate}}) \leq 
\delta$. This chosen $P_{\textit{new}}$ for the current $(f,c)$ pair is then added to 
$S_{\textit{final}}$; this is analogous to deciding whether to add the atom $\textit{accept}(f,c)$ to the hypothesis, based on whether this addition maximizes the number of final assignments without violating the inconsistency threshold $\delta$. The HS algorithm has a running time in 
\(O(|\mathcal{F}| \cdot |\mathcal{C}| \cdot |E_{set}|)\), where \(|\mathcal{F}|\) and \(|\mathcal{C}|\) are the numbers of 
models and classes, and \(|E_{set}|\) is the number of evaluated \(\epsilon\) 
values---that is, the cost is polynomial with respect to these key parameters of the input. 
This structured, greedy method  efficiently selects predictions while managing inconsistencies, rendering it suitable for large-scale problem instances.

\begin{algorithm}[t]
\caption{{\small Heuristic Search (HS) for Prediction Optimization}}
\label{alg:hs_concise}
\begin{algorithmic}[1]
\small
\STATE \textbf{Input:}
\STATE \quad $P_\textit{raw}$ (Set of all raw prediction tuples $(o, l, f, c)$)
\STATE \quad $\delta$ (Maximum allowed inconsistency for $S_{\textit{final}}$)
\STATE \quad $E_\textit{set}$ (Set of EDR $\epsilon$ thresholds to evaluate)
\STATE \quad \COMMENT{Implicit: Sets $\mathcal{F}$ (models), $\mathcal{C}$ (classes); Functions $\textit{GetFilteredPreds}(f,c,\epsilon, P_{\textit{raw}})$ and $\textit{CalcIncon}(S)$.}
\STATE \textbf{Output:} $S_{\textit{final}}$ (Optimized set of prediction tuples $(o,l)$)
\STATE $S_\textit{final} \leftarrow \emptyset$
\FOR{each model $f \in \mathcal{F}$ and class $c \in \mathcal{C}$}
    \STATE $P_{\textit{best\_add}} \leftarrow \emptyset$ \COMMENT{Best predictions from current $(f,c)$ to add}
    \STATE $n_{\textit{current\_max}} \leftarrow |S_{\textit{final}}|$ \COMMENT{Max size of $S_{\textit{final}} \cup P_{\textit{new}}$}
    \FOR{each $\epsilon \in E_{\textit{set}}$}
        \STATE $P_{\textit{new}} \leftarrow \textit{GetFilteredPreds}(f, c, \epsilon, P_{\textit{raw}})$
        \STATE $S_{\textit{cand}} \leftarrow S_{\textit{final}} \cup P_{\textit{new}}$
        \IF{\textit{CalcIncon}($S_{\textit{cand}}$) $\leq \delta$ \AND $|S_{\textit{cand}}| > n_{\textit{current\_max}}$}
            \STATE $P_{\textit{best\_add}} \leftarrow P_{\textit{new}}$
            \STATE $n_{\textit{current\_max}} \leftarrow |S_{\textit{cand}}|$
        \ENDIF
    \ENDFOR
    \IF{$P_{\textit{best\_add}} \neq \emptyset$}
        \STATE $S_{\textit{final}} \leftarrow S_{\textit{final}} \cup P_{\textit{best\_add}}$
    \ENDIF
\ENDFOR
\STATE \textbf{return} $S_{\textit{final}}$
\end{algorithmic}
\end{algorithm}

\smallskip
\noindent\textbf{Tie-Breaker (TB) Mechanism.}  To ensure deterministic class assignment per object and resolve ambiguities where multiple labels remain valid after abduction, we apply a Tie-Breaker (TB) heuristic. For any object $\omega$ with several admissible labels, TB selects the pair $(\omega, c)$ proposed by the perception model with the highest confidence. This refinement yields the IP+TB and HS+TB variants in our experiments.

\section{Experimental Setup}
\label{sec:experiments}

\begin{figure*}[t]
    \centering
    \includegraphics[width=0.98\linewidth]{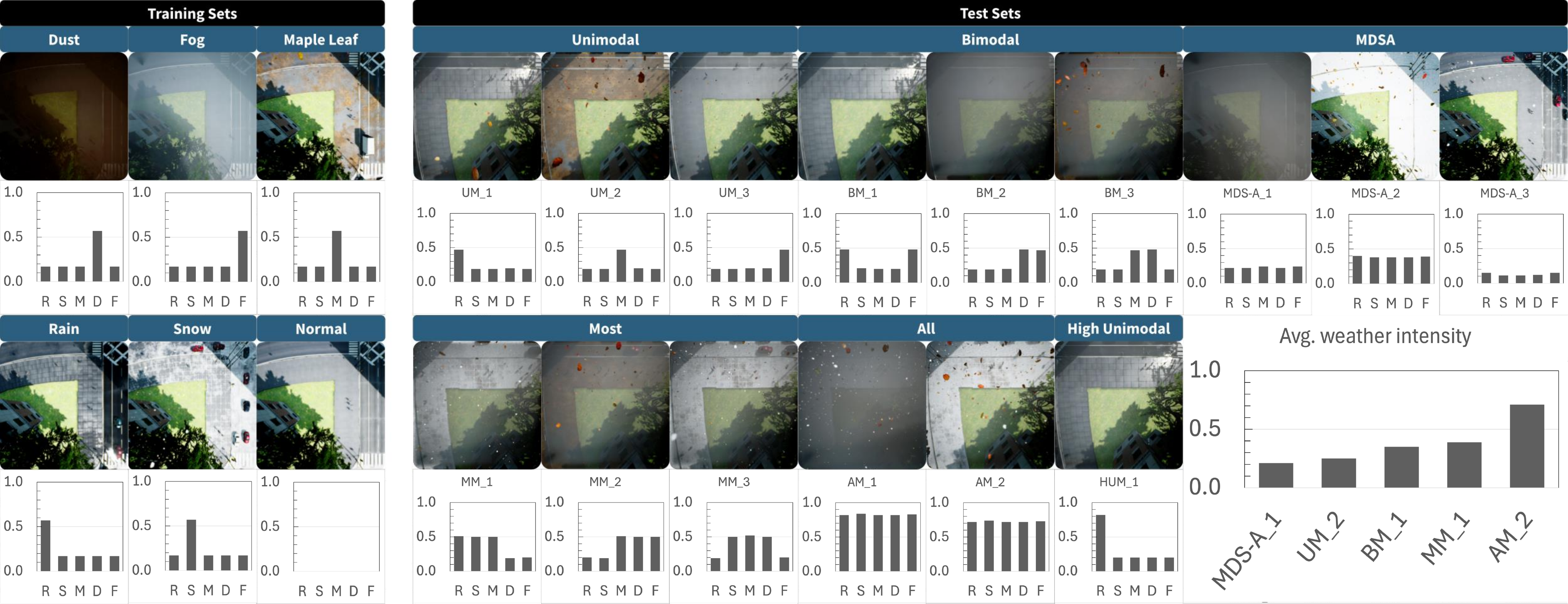}
    \caption{Images captured in the same position in AirSim under various weather conditions along with the distribution of weather conditions of the dataset that it represents. {\em Bottom right}: Histogram showing average intensity in selected datasets.}
    \label{fig:dataset_fid}
\end{figure*}

For evaluating the proposed approaches, we use an extended version of the Multiple 
Distribution Shift -– Aerial (MDS-A) dataset~\cite{ngu2025multipledistributionshift}.
The MDS-A dataset was generated using the AirSim \cite{airsim2017fsr} simulator and designed to analyze the impact of distributional shifts (caused by varying weather conditions) on object detection models in aerial imagery. The dataset consists of images captured in random positions under various weather conditions within a city environment alongside bounding boxes with one of four categories: {\em pedestrians}, {\em vehicles}, {\em nature}, and {\em construction} assigned to each bounding box. 
MDS-A contains six training sets, each simulating a specific weather condition: {\em rain}, {\em snow}, {\em fog}, {\em maple leaves}, {\em dust}, and a {\em baseline} (no weather effects), and three test sets containing a complex mix of weather conditions---Figure~\ref{fig:dataset_fid} illustrates an example of how the intensity distributions of weather effects vary. 

In addition to the existing test sets, we add an augmented test suite comprising 12 new test sets, which introduce a broader range of complex, mixed-weather conditions, further increasing the severity and diversity of distributional shifts. Each entry specifies the intensity level assigned to five different weather conditions, allowing for a systematic exploration of various distributional settings. The suite includes both homogeneous cases (where a single condition dominates) and heterogeneous scenarios involving multiple concurrent weather effects with varying intensities. The id of each dataset refers to the number of weather conditions fitted with the same intensity level: UM (unimodal), BM (bimodal), MM (most models), AM (all models), and HUM (uhigh unimodal). This design aims to evaluate the robustness of the proposed approaches under diverse and increasingly complex distribution shifts.

Six baseline object detection models were trained using the DeTR architecture~\cite{carion2020endtoendobjectdetectiontransformers} 
with a ResNet-50 backbone~\cite{he2015deepresiduallearningimage}, each specialized in one of the weather-specific training datasets. 
The models were intentionally trained independently on their corresponding datasets to emphasize the effects of distributional shifts 
under specific weather conditions. This experimental setup provides a rigorous framework to assess the performance of 
both the integer programming and heuristic approaches under varying inconsistency thresholds and challenging 
weather-induced shifts.

Both the integer programming and heuristic approaches were evaluated using a range of 
inconsistency thresholds: $\delta \in [0.01, 0.1, 0.2, 0.3, 0.4, 0.5, 0.6, 0.7, 0.8, 0.9, 1.0]$. 
For the heuristic method, the trade-off parameter $\epsilon$ was also varied over the same range. 
Initially, all model-class pairs were included in the optimization process, allowing the 
heuristic approach to adaptively determine which combinations most effectively improve 
prediction performance while satisfying the inconsistency constraint. The integer 
programming formulation followed a similar configuration, incorporating $\delta$ as a hard 
constraint to limit the allowable inconsistencies. Each approach was run 50 times on each test sets.

To assess performance, we report the following metrics: \textit{F1-score,
accuracy}, and \textit{execution time}; the latter is measured as a function of the number of 
objects present in each analyzed image. The results are compared against three baselines: 
the majority vote (MV) ensemble method (which selects the most common class prediction 
across models), the best-performing individual model, and the average performance of all models.

All experiments were conducted on a high-performance computing system, leveraging its advanced 
computational capabilities. Specifically, we used two configurations: 
(1)~high-memory node: a Dell PowerEdge R6525 equipped with AMD EPYC 7713 64-Core processors and 2TB of RAM, and 
(2)~GPU node: a Dell PowerEdge R7525 with AMD EPYC 7413 24-Core processors, 512GB of RAM, and three NVIDIA 
A30 GPUs. 
The logical deduction process, which applies the learned EDR to raw model outputs to identify predictions and errors, was implemented using PyReason. 
Subsequently, the IP solutions were implemented using the PuLP library for optimization. 

\section{Results and Discussion}
\label{sec:results_discussion}

We now discuss the results of the empirical evaluation of our proposed consistency-based abductive reasoning approaches for integrating predictions from multiple models in novel environments. We evaluate their effectiveness and ensure final consistency, including variants that incorporate a tie-break (TB) refinement (IP+TB and HS+TB). 
The key performance metrics, namely \emph{F1-score} and \emph{Accuracy}, are calculated across the suite of~15 test datasets, encompassing diverse distributional shifts as described in the experimental setup section. 
Figure~\ref{tab:main_results_ablation} (left) summarizes these primary results, providing a comprehensive overview of how each method performs under various challenging conditions.

\begin{figure*}[t]
    \centering
    \scriptsize
    \begin{minipage}[t]{0.49\textwidth}
        \centering
        \begin{tabular}{p{30pt}|p{6pt}c|p{6pt}c|p{6pt}c|p{6pt}c|p{6pt}c}
\toprule
\multirow{2}{*}{\textbf{Test Set}} & \multicolumn{2}{c|}{\textbf{Best}} & \multicolumn{2}{c|}{\textbf{Avg.}} & \multicolumn{2}{c|}{\textbf{MV}} & \multicolumn{2}{c|}{\textbf{IP+TB}} & \multicolumn{2}{c}{\textbf{HS+TB}} \\
 & F1 & Acc & F1 & Acc & F1 & Acc & F1 & Acc & F1 & Acc \\
\midrule
MDS-A\_1 & \underline{0.57} & \underline{0.40} & 0.52 & 0.36 & 0.28 & 0.34 & \textbf{0.58} & \textbf{0.41} & \textbf{0.58} & \textbf{0.41} \\
MDS-A\_2 & \underline{0.33} & \underline{0.20} & 0.29 & 0.17 & 0.26 & \textbf{0.22} & \textbf{0.37} & \textbf{0.22} & 0.32 & 0.19 \\
MDS-A\_3 & 0.54 & 0.37 & 0.49 & 0.33 & 0.39 & 0.29 & \textbf{0.56} & \textbf{0.39} & \underline{0.55} & \underline{0.38} \\
UM\_1 & 0.54 & 0.37 & 0.47 & 0.31 & 0.26 & 0.23 & \textbf{0.64} & \textbf{0.47} & \underline{0.61} & \underline{0.44} \\
UM\_2 & 0.56 & 0.38 & 0.46 & 0.31 & 0.25 & 0.22 & \textbf{0.64} & \textbf{0.47} & \underline{0.61} & \underline{0.44} \\
UM\_3 & 0.54 & 0.37 & 0.43 & 0.28 & 0.22 & 0.19 & \textbf{0.63} & \textbf{0.46} & \underline{0.59} & \underline{0.42} \\
BM\_1 & \underline{0.42} & \underline{0.27} & 0.33 & 0.20 & 0.19 & 0.16 & \textbf{0.45} & \textbf{0.29} & 0.39 & 0.24 \\
BM\_2 & 0.33 & 0.20 & 0.25 & 0.15 & 0.14 & 0.12 & \textbf{0.37} & \textbf{0.23} & \underline{0.36} & \underline{0.22} \\
BM\_3 & 0.37 & 0.23 & 0.31 & 0.19 & 0.18 & 0.16 & \textbf{0.43} & \textbf{0.27} & \underline{0.40} & \underline{0.25} \\
MM\_1 & \underline{0.46} & \underline{0.30} & 0.40 & 0.25 & 0.22 & 0.21 & \textbf{0.51} & \textbf{0.34} & \underline{0.46} & \underline{0.30} \\
MM\_2 & \underline{0.32} & \underline{0.19} & 0.24 & 0.14 & 0.13 & 0.10 & \textbf{0.36} & \textbf{0.22} & 0.29 & 0.17 \\
MM\_3 & \underline{0.41} & \underline{0.26} & 0.35 & 0.22 & 0.18 & 0.16 & \textbf{0.46} & \textbf{0.30} & 0.39 & 0.24 \\
AM\_1 & \underline{0.18} & \underline{0.10} & 0.12 & 0.07 & 0.05 & 0.04 & \textbf{0.21} & \textbf{0.11} & \underline{0.18} & \underline{0.10} \\
AM\_2 & \underline{0.23} & \underline{0.13} & 0.18 & 0.10 & 0.07 & 0.06 & \textbf{0.28} & \textbf{0.16} & \underline{0.23} & \underline{0.13} \\
HUM\_1 & 0.45 & 0.29 & 0.40 & 0.25 & 0.18 & 0.17 & \textbf{0.57} & \textbf{0.40} & \underline{0.55} & \underline{0.38} \\
\bottomrule
\end{tabular}
    \end{minipage}
    \hfill
    \begin{minipage}[t]{0.49\textwidth}
        \centering
        \begin{tabular}{l|cc|cc}
\toprule
\multirow{2}{*}{\textbf{Test Set}} & \multicolumn{2}{c|}{\textbf{IP (No TB)}} & \multicolumn{2}{c}{\textbf{HS (No TB)}}\\
& F1 (\% Diff) & Acc (\% Diff) & F1 (\% Diff) & Acc (\% Diff)\\
\midrule
MDS-A\_1 & 0.58 (0.0) & 0.41 (0.0) & 0.52 (-10.3\%) & 0.35 (-14.6\%)\\
MDS-A\_2 & 0.37 (0.0) & 0.22 (0.0) & 0.27 (-15.6\%) & 0.16 (-16.7\%)\\
MDS-A\_3 & 0.56 (0.0) & 0.39 (0.0) & 0.49 (-10.9\%) & 0.32 (-15.8\%)\\
UM\_1 & 0.64 (0.0) & 0.47 (0.0) & 0.53 (-13.1\%) & 0.36 (-18.2\%)\\
UM\_2 & 0.64 (0.0) & 0.47 (0.0) & 0.52 (-14.1\%) & 0.35 (-18.8\%)\\
UM\_3 & 0.63 (0.0) & 0.46 (0.0) & 0.52 (-11.9\%) & 0.35 (-16.7\%)\\
BM\_1 & 0.45 (0.0) & 0.29 (0.0) & 0.34 (-11.1\%) & 0.20 (-16.7\%)\\
BM\_2 & 0.37 (0.0) & 0.23 (0.0) & 0.31 (-13.5\%) & 0.19 (-13.6\%)\\
BM\_3 & 0.43 (0.0) & 0.27 (0.0) & 0.34 (-15.0\%) & 0.20 (-20.0\%)\\
MM\_1 & 0.51 (0.0) & 0.34 (0.0) & 0.38 (-15.7\%) & 0.24 (-20.0\%)\\
MM\_2 & 0.36 (0.0) & 0.22 (0.0) & 0.25 (-13.8\%) & 0.14 (-17.6\%)\\
MM\_3 & 0.46 (0.0) & 0.30 (0.0) & 0.33 (-15.4\%) & 0.20 (-16.7\%)\\
AM\_1 & 0.21 (0.0) & 0.11 (0.0) & 0.15 (-16.7\%) & 0.08 (-20.0\%)\\
AM\_2 & 0.28 (0.0) & 0.16 (0.0) & 0.19 (-17.4\%) & 0.11 (-15.4\%)\\
HUM\_1 & 0.57 (0.0) & 0.40 (0.0) & 0.48 (-12.7\%) & 0.32 (-15.8\%)\\
\bottomrule
\end{tabular}
    \end{minipage}
\caption{{\em Left:} Performance (F1 and Accuracy) across all test sets. Best values per test set in bold, the second-best are underlined. \linebreak
{\em Right:}~Ablation Study -- Performance without Tie-Breaker (TB). Values show F1-score or Accuracy for the method without TB, with the percentage difference relative to the corresponding + TB version (w.r.t.\ values on the left, shown in parentheses).
}
\label{tab:main_results_ablation}
\end{figure*}

\smallskip
\noindent\textbf{Overall Performance Analysis.}
Examining the F1-score and Accuracy metrics, the IP+TB method consistently demonstrates superior performance, achieving the highest scores in all cases. 
For instance, on the challenging AM\_1, AM\_2, and HUM\_1 test sets, characterized by significant distributional shifts, IP+TB yields notable improvements over the best-performing individual model, and significantly surpasses the standard Majority Vote (MV) ensemble, which often struggles in these complex scenarios (e.g., F1 scores of 0.21 vs.\ 0.18 vs.\ 0.05 on AM\_1). 
The heuristic approach, HS+TB, also frequently outperforms the baselines, achieving for example the second-best F1 and Accuracy on the UM\_1 test set (F1 0.61, Acc 0.44) and demonstrating strong performance on others like MDS-A\_1. These results highlight the effectiveness of combining consistency-based abduction with a tie-breaking mechanism that selects the highest confidence prediction among inconsistent options. The contribution of this tie-breaker component is further analyzed in the ablation study presented below.

\begin{figure}[t]
    \centering
    \includegraphics[width=0.94\linewidth]{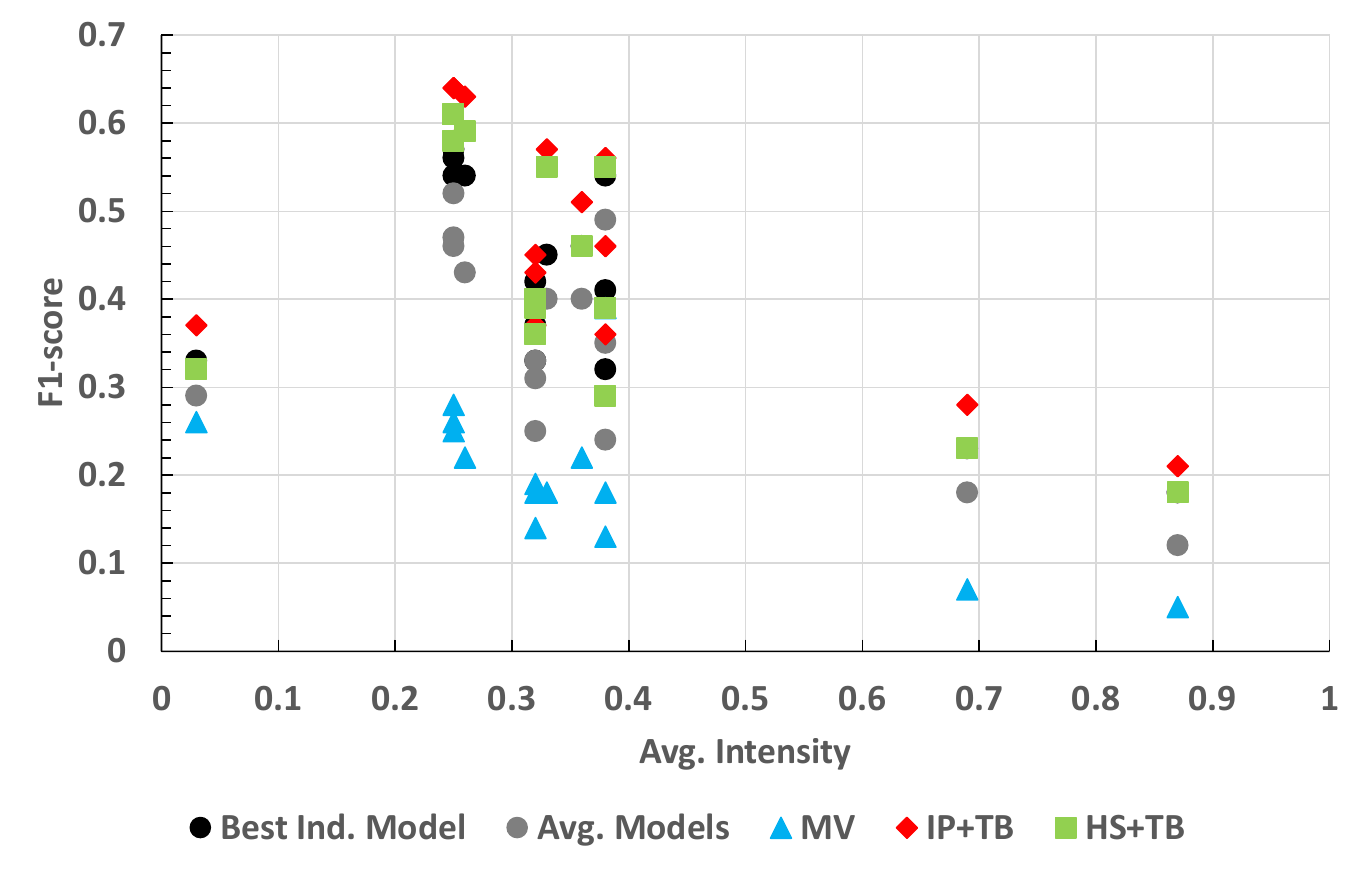}
    \caption{F1-scores for IP+TB and HS+TB vs.\ baselines (Best Ind. Model, Avg. Models, and Maj.\ Vote) across the 15 test datasets under increasing average weather intensity.}
    \label{fig:f1_grouptest}
\end{figure}

\smallskip
\noindent\textbf{Environmental Analysis.}  To assess the robustness of our proposed methods against increasing environmental challenge, Figure~\ref{fig:f1_grouptest} displays the F1-score for all 15 test datasets plotted against their average environmental intensity. Each point represents a unique test set, with different markers indicating the performance of IP+TB, HS+TB, and the baseline methods (Best Individual Model, Average Models, and Majority Vote). The average environmental intensity for each dataset (x-axis) is a normalized measure reflecting the severity of simulated conditions.

As observable in Figure~\ref{fig:f1_grouptest}, while there is a general trend of decreasing F1-scores for all methods as the average environmental intensity increases—indicating the inherent difficulty of operating in more severe novel environments—our IP+TB approach (represented by red diamonds) consistently achieves the highest F1-scores across the entire spectrum of intensities. In nearly all instances, IP+TB surpasses the Best Individual Model, and significantly outperforms the Average Models, Majority Vote, and our Heuristic Search (HS+TB) approach in terms of F1-score. This consistent superiority, irrespective of the environmental intensity level, underscores the effectiveness of IP+TB in robustly integrating model predictions and managing inconsistencies. For instance, even at higher intensity levels where all methods experience performance degradation, IP+TB maintains a clear advantage. The same analysis for Accuracy shows comparable trends, and is provided in the  extended version.

These findings demonstrate that while environmental novelty impacts all approaches, our consistency-based abductive reasoning, particularly when implemented via Integer Programming (IP+TB), provides a robust performance advantage in F1-score over baseline methods and our heuristic alternative across a wide range of challenging conditions.

\begin{figure*}[] 
\centering
\includegraphics[width=0.33\linewidth]{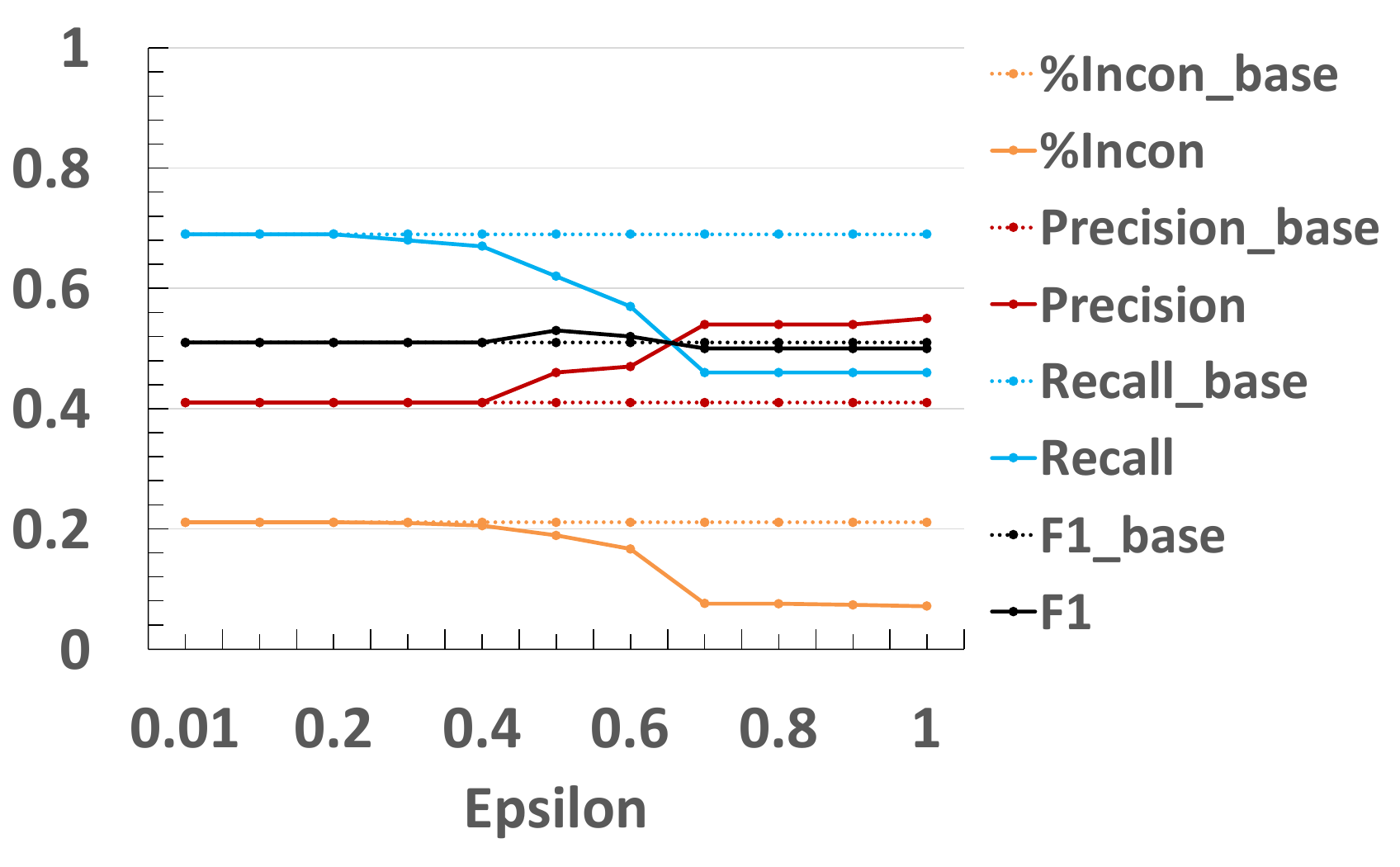}
\includegraphics[width=0.33\linewidth]{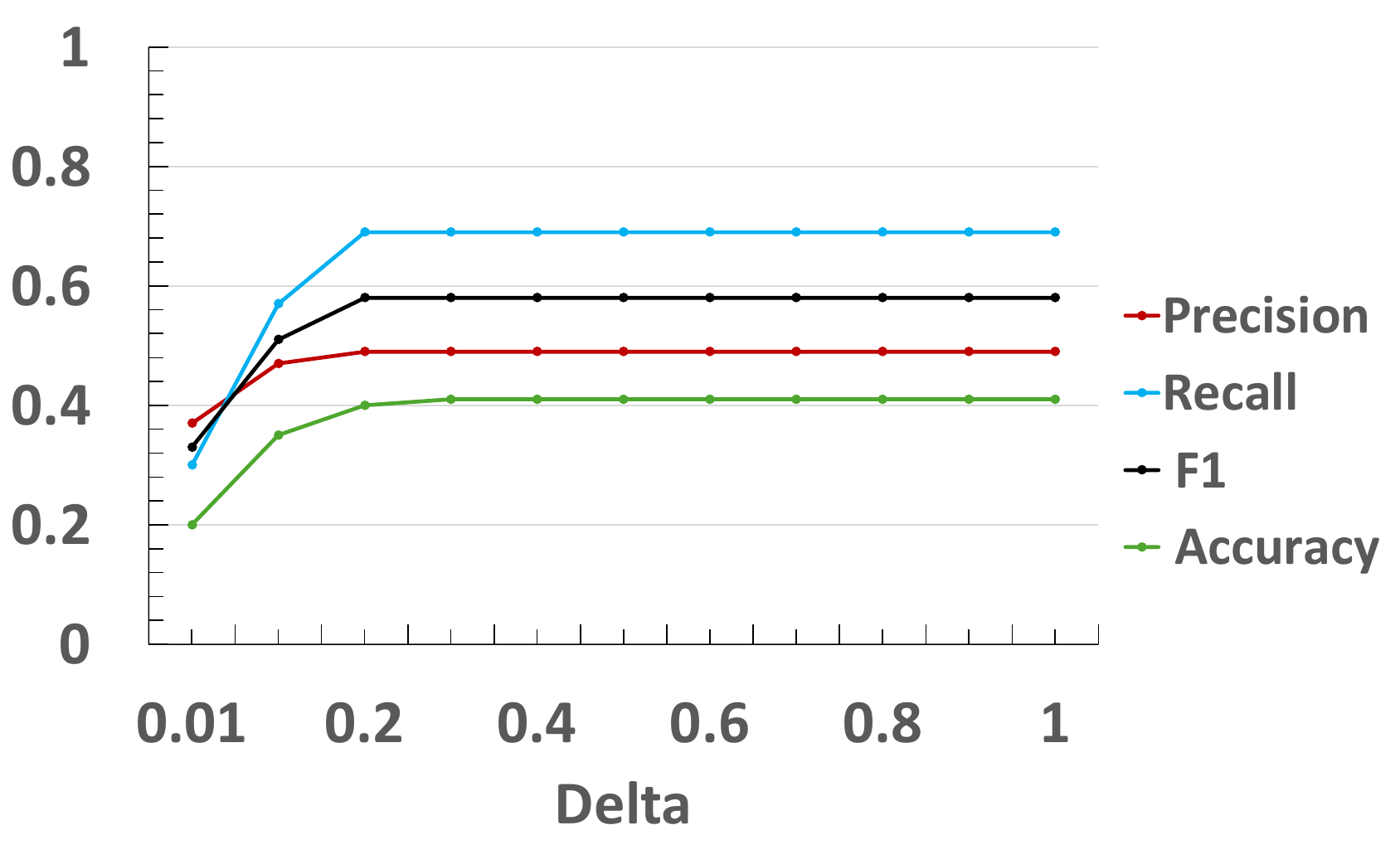}
\includegraphics[width=0.33\linewidth]{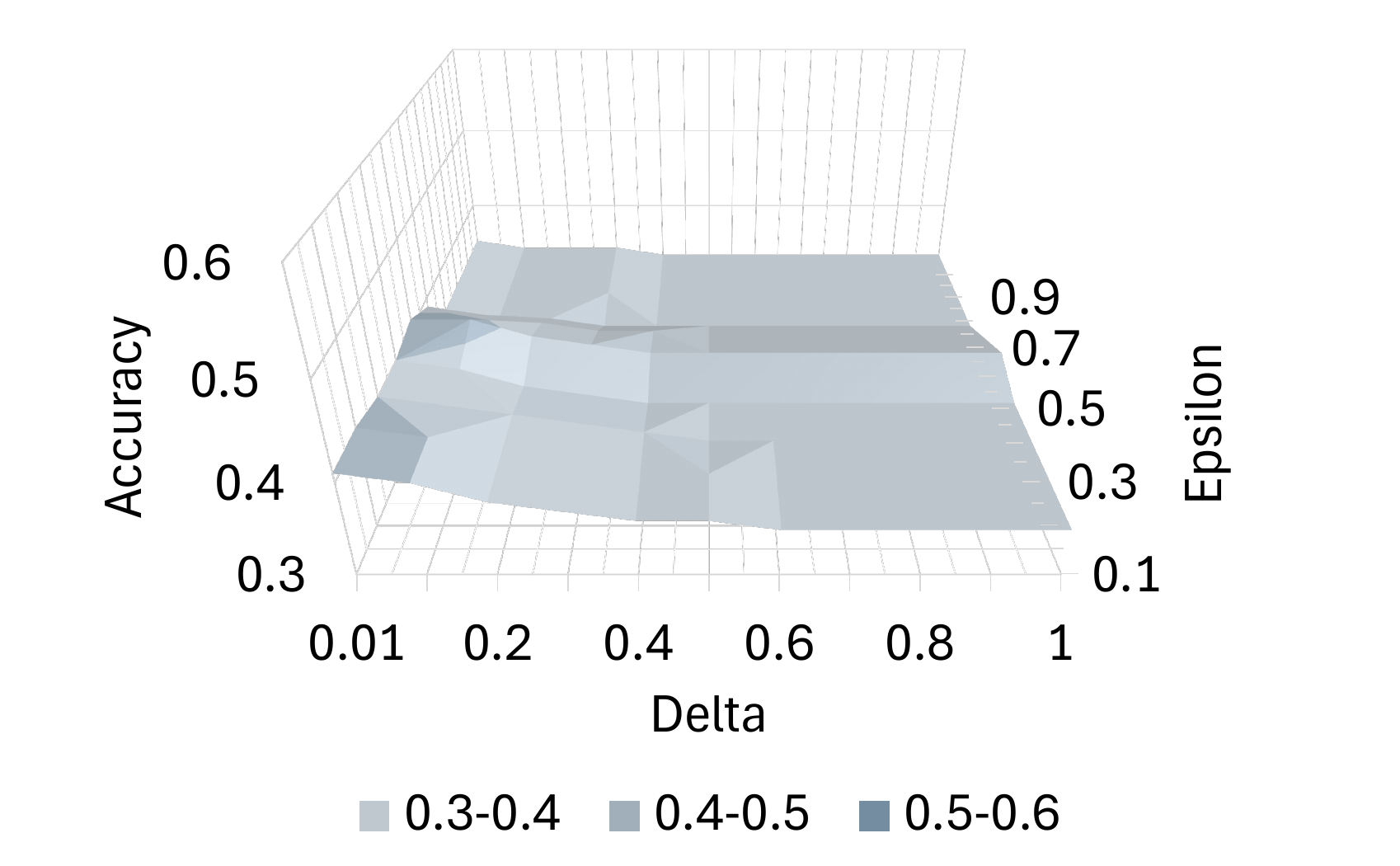}
\caption{Hyperparameter sensitivity for the MDS-A\_1 test set.
{\em Left}: Performance metrics and inconsistency rates of the Error Detection Rule (EDR) stage across varying $\epsilon$ values.
{\em Center}: Heuristic Search (HS+TB) performance, as a function of the $\delta$ inconsistency threshold. 
{\em Right}: IP+TB Accuracy depicted as a surface plot, varying $\delta$ and internal~$\epsilon$.
}
\label{fig:hyperparamter_sens}
\end{figure*}

\smallskip
\noindent\textbf{Tie Breaker Ablation.}  To evaluate the Tie-Breaker (TB) mechanism's role in ensuring consistent final predictions and its effect on performance, we carried out an ablation study comparing results with TB (Figure~\ref{tab:main_results_ablation}-left) and without TB (summarized in Figure~\ref{tab:main_results_ablation}-right, showing percentage differences). For the Integer Programming (IP) approach, removing the TB resulted in a 0.0\% difference in performance across all metrics and datasets.
This is a notable outcome, suggesting that for the chosen optimal $\delta$ values (typically in the range of~0.1 to~0.3, as per our sensitivity analysis), the IP optimization, in its pursuit of maximizing valid assignments inherently converged to solutions that were already fully consistent in terms of providing a single, unambiguous label per object, or where any minor ambiguities were resolvable by the TB without affecting the F1-score or Accuracy. 
Thus, the IP approach effectively achieved a high degree of output consistency by primarily leveraging other aspects of the formulation, such as the elimination of less reliable model-class pairs, potentially rendering the solution more consistent than strictly required by the explicit $\delta$ budget for conflicting assignments. 
In contrast, the Heuristic Search (HS) performance degraded consistently without the TB. As shown by the negative percentage differences in Figure~\ref{tab:main_results_ablation} (right), the F1-score for HS dropped by 10\% to over 17\% across various datasets when the TB was removed, with similar reductions in Accuracy. We remind the reader that the tie-breaker heuristic selects the highest confidence prediction when an inconsistency is present.

\smallskip
\noindent\textbf{EDR Rule Strictness.}  We also analyze the sensitivity of the initial Error Detection Rule (EDR) learning 
stage to its recall reduction threshold~$\epsilon$. This parameter controls the aggressiveness 
of filtering potentially erroneous predictions from individual models before the main 
abduction process. Figure~\ref{fig:hyperparamter_sens} (left)
illustrates the typical impact of varying $\epsilon$ on precision, recall, F1-score, and the inconsistency rate (i.e., conflicting predictions) for the MDS-A\_1 test set.

As expected, increasing $\epsilon$ generally leads to higher 
precision but lower recall, demonstrating the inherent trade-off controlled by this 
parameter. Notably, this stricter filtering also tends to reduce the 
baseline level of inconsistency among the remaining predictions. This analysis 
highlights how tuning $\epsilon$ shapes the pool of candidate predictions subsequently processed by our IP+TB and HS+TB abduction methods aiming to maximize valid assignments under the
global inconsistency constraint $\delta$. Detailed sensitivity plots for all test sets are provided in extended version.

\smallskip
\noindent\textbf{Hyperparameter Sensitivity.}  We analyzed the sensitivity of our approaches to the maximum inconsistency 
threshold hyperparameter $\delta$, which applies to IP+TB and HS+TB, and the internal recall parameter $\epsilon$ used within the IP+TB optimization. Figure~\ref{fig:hyperparamter_sens} (right) illustrates this analysis for the MDS-A\_1 test set. The 3D plot of IP+TB show that the F1 score and accuracy reach their maximum value in the ranges of~$\delta$ between~0.1 and~0.3, and for $\epsilon$ in~0.1 and~0.5. 
Similarly, the performance of HS+TB (center) stabilizes quickly as $\delta$ increases above a small initial value (e.g., 0.2), which is associated with the maximum level of inconsistency present in the initial test set (Figure~\ref{fig:hyperparamter_sens}, left); 
detailed sensitivity plots for all test sets are provided in the extended version.

\smallskip
\noindent\textbf{Running Time Analysis and Complexity.} A key practical difference between our approaches is computational cost. 
Our empirical results, illustrated in a running time plot available in the extended version, align with the theoretical complexities of the methods. 
As expected, our polynomial-time HS+TB approach is significantly faster than the exact IP+TB method. While the IP solver's running time is higher due to its exact nature, it remained tractable for the instances explored here. We observed that the average processing time per object for IP+TB did not increase steeply with more objects. We hypothesize that this is due to a combination of amortized solver overheads and structural dataset characteristics that allow for faster resolution. A detailed study of these issues is future work.

\section{Conclusions and Future Work}
We presented a comprehensive approach to address inconsistencies in model predictions via an abductive reasoning formulation. 
In future work, we plan to focus on the logic program for the deduction process to incorporate more sophisticated rules to infer alternative sets of assignments that address diverse inconsistency scenarios among model predictions. Such enhancements will be applied to new domains and datasets characterized by complex distributions.
Further, we plan to refine our analysis by carrying out a more granular exploration of values of both the $\epsilon$ and $\delta$ parameters to evaluate the impact on solution robustness.
Finally, optimizing runtime efficiency remains a priority to enable scalability and practical application in real-world scenarios.

\section*{Acknowledgments}
This research was supported by the Defense Advanced Research Projects Agency (DARPA) under Cooperative Agreement No. HR00112420370 (MCAI), the U.S. Army Combat Capabilities Development Command Army Research Laboratory under Support Agreement No. USMA 21050, and DARPA under Support Agreement No. USMA 23004. The views expressed in this paper are those of the authors and do not necessarily reflect the official policy or position of the U.S. Military Academy, the U.S. Army, the U.S. Department of Defense, or the U.S. Government.

%
%
\onecolumn
\section*{Appendix}
This document provides supplementary material for the AAAI-26 submission titled ``Consistency-based Abductive Reasoning over Perceptual Errors of Multiple Pre-trained Models in Novel Environments'' (Submission ID: 19865).
It includes some detailed of the implementations and per-testset results:
\begin{enumerate}
    \item Unique Name Assumption Implementation Details
    \item Comparison of all methods (F1 and Accuracy)
    \item Runtime of the IP+TB, HS+TB, and MV approaches.
    \item Performance Across Varying Environmental Intensity (accuracy and Accuracy)
    \item Sensitivity analysis of the Error Detection Rule (EDR) stage to its $\epsilon$ parameter.
    \item Sensitivity analysis of the Heuristic Search (HS+TB) method to the inconsistency threshold $\delta$.
    \item Sensitivity analysis of the Integer Programming (IP+TB) method to the inconsistency threshold $\delta$ and its internal optimization parameter $\epsilon$.
    \item Details of the hyperparameters used to train the object detection models, and solver for IP optimization.
    \item URLs of the datasets and the anonymized code.
\end{enumerate}
For each analysis type, results are presented for all 15 test datasets used in our experiment.

\clearpage
\section{Unique Name Assumption Implementation Details}
In Section \textquotedblleft Consistency-based Abduction Problem,\textquotedblright we introduce the \textbf{Unique Name Assumption (UNA)}, which posits that each real-world object $\omega \in \Omega$ corresponds to a singular, distinct entity. To operationalize this assumption and derive the initial set of \textit{observations} $O$, defined as facts of the form $f_i(\omega) = c_j$ (meaning model $i$ identifies object $\omega$ as class $c_j$), we employ a two-stage bounding box matching mechanism. This mechanism is crucial for linking the raw detections from our multiple pre-trained perception models $F = \{f_1, f_2, \dots, f_\eta\}$ to the objects present in the ground truth for a given image.

\begin{enumerate}
    \item \textbf{Per-Model Primary Assignment:} For each individual perception model $f_i \in F$, and for every ground truth object $\omega$ in the image, we first attempt to find a corresponding bounding box among the detections generated \textbf{by that specific model $f_i$}. The assignment is made to the \textbf{first bounding box} encountered (typically based on a predefined order, e.g., highest confidence score) that exhibits an Intersection Over Union (IoU) overlap \textbf{greater than 90\%} with the ground truth object's bounding box. This step generates the initial $f_i(\omega) = c_j$ facts for each model, which form the basis of the observations $O$.

    \item \textbf{Global Fallback Assignment:} After the primary assignment stage, it is possible that some ground truth objects $\omega$ may remain unassigned by any of the individual models (i.e., no bounding box from any single model met the 90\% IoU threshold for that object). To ensure comprehensive coverage and further adhere to the UNA (where every ground truth object should ideally have an associated prediction), for each such unassigned ground truth object, we search through \textbf{all remaining bounding box detections from \textit{all} models} in $F$. From this aggregated pool of detections, we select the one that yields the \textbf{highest IoU score} with the unassigned ground truth object. This highest-IoU detection is then assigned to the ground truth object, providing a final prediction for it.
\end{enumerate}
This two-tiered approach allows us to robustly establish a unique correspondence between ground truth objects and model predictions, forming the set of observations $O$ upon which our consistency-based abduction framework operates. This implementation detail was deferred to the appendix to maintain focus on the core abductive reasoning framework in the main text.

\clearpage
\section{Comparison of all methods}
\label{sec:app_comp_all_methods}
Table~\ref{tab:supp_comparison_all_methods} shows performance (F1 and Accuracy) across all test sets. Best values per test set in bold, the second-best are underlined.
\begin{table}[h!]
\begin{tabular}{l|lc|lc|lc|ll|lc|lc}
\hline
\multirow{2}{*}{\textbf{Test Set}} & \multicolumn{2}{c|}{\textbf{Best}}                            & \multicolumn{2}{c|}{\textbf{Avg.}} & \multicolumn{2}{c|}{\textbf{MV}} & \multicolumn{2}{c|}{\textbf{TB}} & \multicolumn{2}{c|}{\textbf{IP+TB}} & \multicolumn{2}{c}{\textbf{HS+TB}}                            \\
                                   & F1                            & Acc                           & F1               & Acc             & F1         & Acc                 & F1             & Acc             & F1               & Acc              & F1                            & Acc                           \\ \hline
MDS-A\_1                           & \underline{0.57} & \underline{0.40} & 0.52             & 0.36            & 0.28       & 0.34                &       0.51         &         0.35        & \textbf{0.58}    & \textbf{0.41}    & \textbf{0.58}                 & \textbf{0.41}                 \\
MDS-A\_2                           & \underline{0.33} & \underline{0.20} & 0.29             & 0.17            & 0.26       & \textbf{0.22}       &       0.27         &        0.16         & \textbf{0.37}    & \textbf{0.22}    & 0.32                          & 0.19                          \\
MDS-A\_3                           & 0.54                          & 0.37                          & 0.49             & 0.33            & 0.39       & 0.29                &       0.49         &        0.32         & \textbf{0.56}    & \textbf{0.39}    & \underline{0.55} & \underline{0.38} \\
UM\_1                              & 0.54                          & 0.37                          & 0.47             & 0.31            & 0.26       & 0.23                &     0.53           &         0.36        & \textbf{0.64}    & \textbf{0.47}    & \underline{0.61} & \underline{0.44} \\
UM\_2                              & 0.56                          & 0.38                          & 0.46             & 0.31            & 0.25       & 0.22                &        0.52        &         0.35        & \textbf{0.64}    & \textbf{0.47}    & \underline{0.61} & \underline{0.44} \\
UM\_3                              & 0.54                          & 0.37                          & 0.43             & 0.28            & 0.22       & 0.19                &        0.52        &       0.35          & \textbf{0.63}    & \textbf{0.46}    & \underline{0.59} & \underline{0.42} \\
BM\_1                              & \underline{0.42} & \underline{0.27} & 0.33             & 0.20            & 0.19       & 0.16                &         0.34       &       0.20          & \textbf{0.45}    & \textbf{0.29}    & 0.39                          & 0.24                          \\
BM\_2                              & 0.33                          & 0.20                          & 0.25             & 0.15            & 0.14       & 0.12                &       0.31         &         0.19        & \textbf{0.37}    & \textbf{0.23}    & \underline{0.36} & \underline{0.22} \\
BM\_3                              & 0.37                          & 0.23                          & 0.31             & 0.19            & 0.18       & 0.16                &       0.34         &         0.20        & \textbf{0.43}    & \textbf{0.27}    & \underline{0.40} & \underline{0.25} \\
MM\_1                              & \underline{0.46} & \underline{0.30} & 0.40             & 0.25            & 0.22       & 0.21                &       0.38         &     0.24            & \textbf{0.51}    & \textbf{0.34}    & \underline{0.46} & \underline{0.30} \\
MM\_2                              & \underline{0.32} & \underline{0.19} & 0.24             & 0.14            & 0.13       & 0.10                &       0.25         &      0.14           & \textbf{0.36}    & \textbf{0.22}    & 0.29                          & 0.17                          \\
MM\_3                              & \underline{0.41} & \underline{0.26} & 0.35             & 0.22            & 0.18       & 0.16                &        0.33        &       0.20          & \textbf{0.46}    & \textbf{0.30}    & 0.39                          & 0.24                          \\
AM\_1                              & \underline{0.18} & \underline{0.10} & 0.12             & 0.07            & 0.05       & 0.04                &        0.15        &      0.08           & \textbf{0.21}    & \textbf{0.11}    & \underline{0.18} & \underline{0.10} \\
AM\_2                              & \underline{0.23} & \underline{0.13} & 0.18             & 0.10            & 0.07       & 0.06                &        0.19        &     0.11            & \textbf{0.28}    & \textbf{0.16}    & \underline{0.23} & \underline{0.13} \\
HUM\_1                             & 0.45                          & 0.29                          & 0.40             & 0.25            & 0.18       & 0.17                &        0.48        &       0.32          & \textbf{0.57}    & \textbf{0.40}    & \underline{0.55} & \underline{0.38} \\ \cline{1-13}
\end{tabular}
\caption{}
\label{tab:supp_comparison_all_methods}
\end{table}


\clearpage
\section{Detailed Runtime Analysis per Test Sets}
\label{sec:supp_runtime}

This section presents the detailed runtime performance for the Integer Programming with Tie-Breaker (IP+TB), Heuristic Search with Tie-Breaker (HS+TB) and Majority Vote (MV) methods across all 15 test datasets. Each figure below displays the average processing time per object (in seconds, log scale) as a function of the number of objects for a specific dataset, comparing IP+TB, HS+TB, and the Majority Vote (MV).

\supplementaryfigure{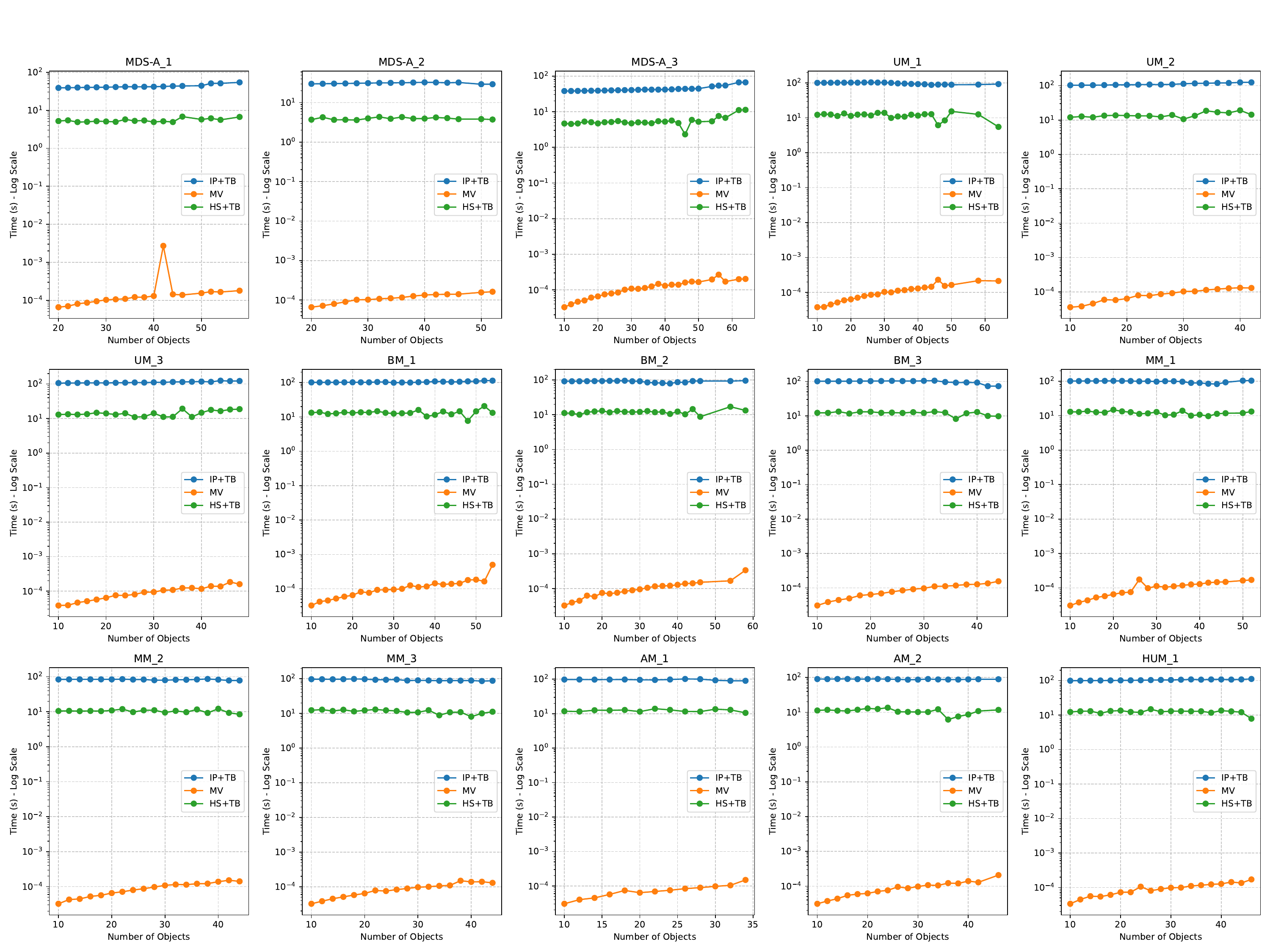}{Detailed runtime per test sets}{fig:runtimes}

\clearpage
\section{Performance Across Varying Environmental Intensity (accuracy and Accuracy)}
\label{sec:supp_runtime}

\begin{figure}[h]
    \centering
    \begin{minipage}[b]{0.48\textwidth}
        \centering
        \includegraphics[width=\textwidth]{f1_vs_avginte.pdf}
    \end{minipage}
    \hfill
    \begin{minipage}[b]{0.48\textwidth}
        \centering
        \includegraphics[width=\textwidth]{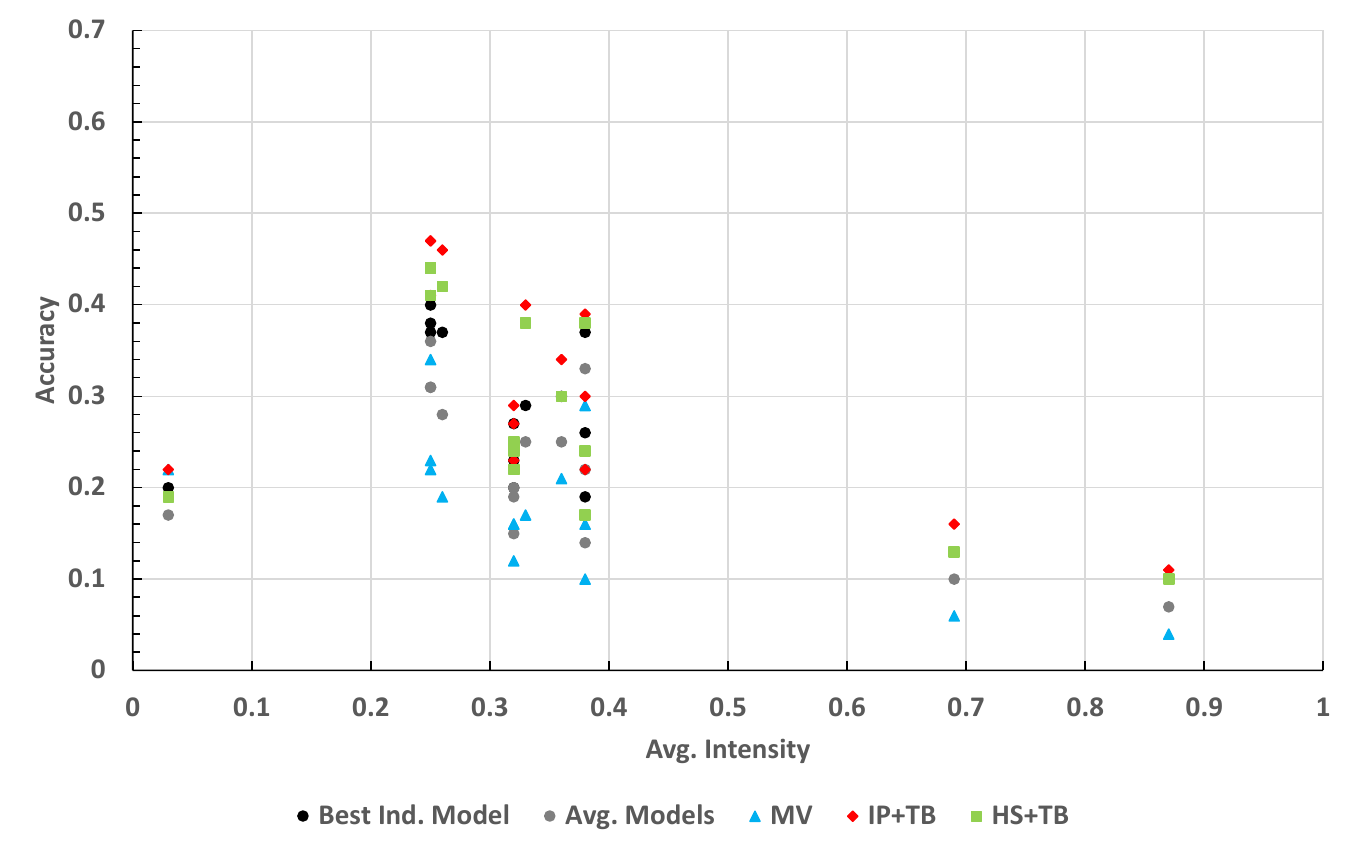}
    \end{minipage}
    \caption{F1-score (left) and Accuracy (right) of the proposed methods (IP+TB, HS+TB) against the baselines under increasing average weather intensity.}
    \label{fig:sup_accuracy_acc_grouptest}
\end{figure}

\clearpage
\section{Detailed EDR Epsilon ($\epsilon$) Sensitivity Analysis per Test Sets}
\label{sec:supp_edr_epsilon}

This section provides a detailed analysis of the Error Detection Rule (EDR) learning stage's sensitivity to its recall reduction threshold, $\epsilon$, for each of the 15 test datasets. The figures illustrate how precision, recall, accuracy-score, and the base inconsistency rate evolve as $\epsilon$ varies. This complements Figure 5 (top-left) of the main paper, which shows this analysis for a single representative test set.

\begin{figure}[h!]
    \centering
    \begin{subfigure}[b]{0.31\textwidth}
        \centering
        \includegraphics[width=\linewidth]{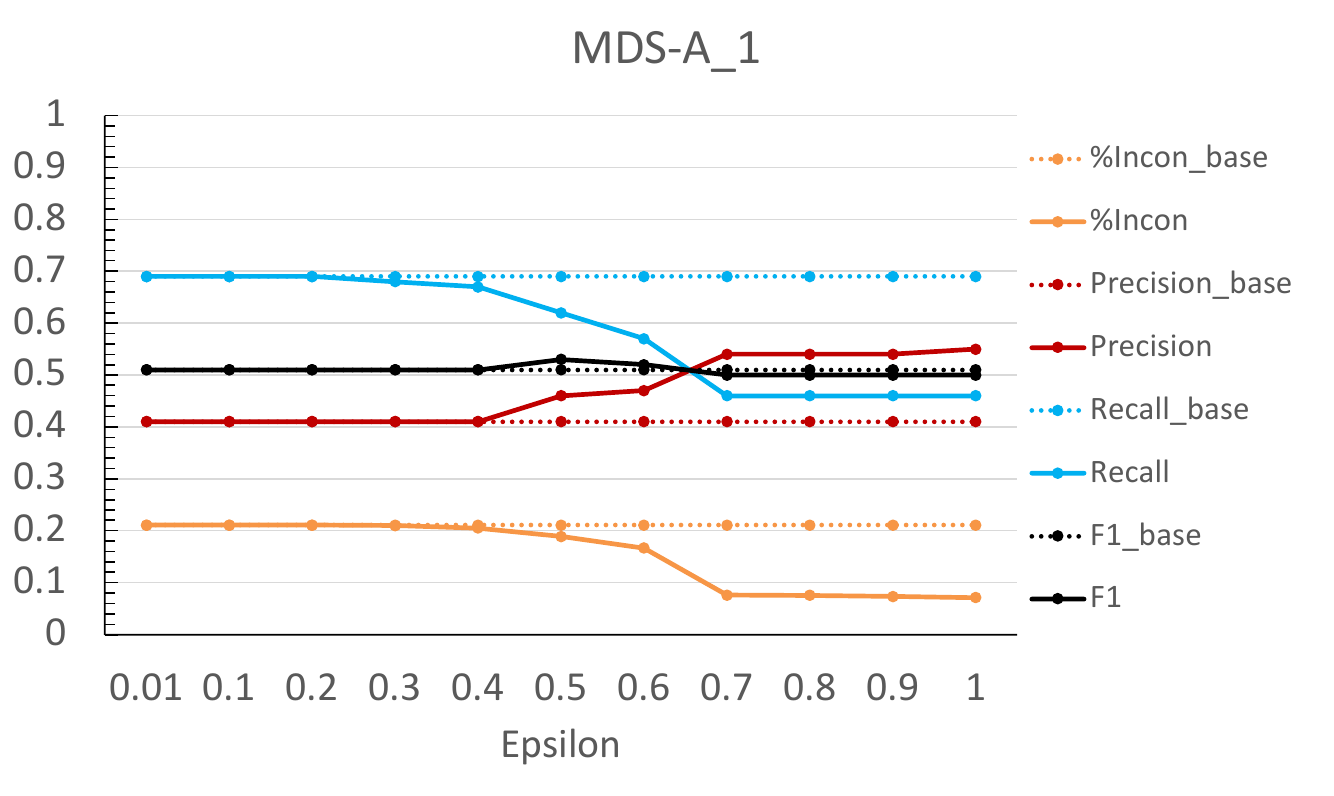}
    \end{subfigure}
    \hfill 
    \begin{subfigure}[b]{0.31\textwidth}
        \centering
        \includegraphics[width=\linewidth]{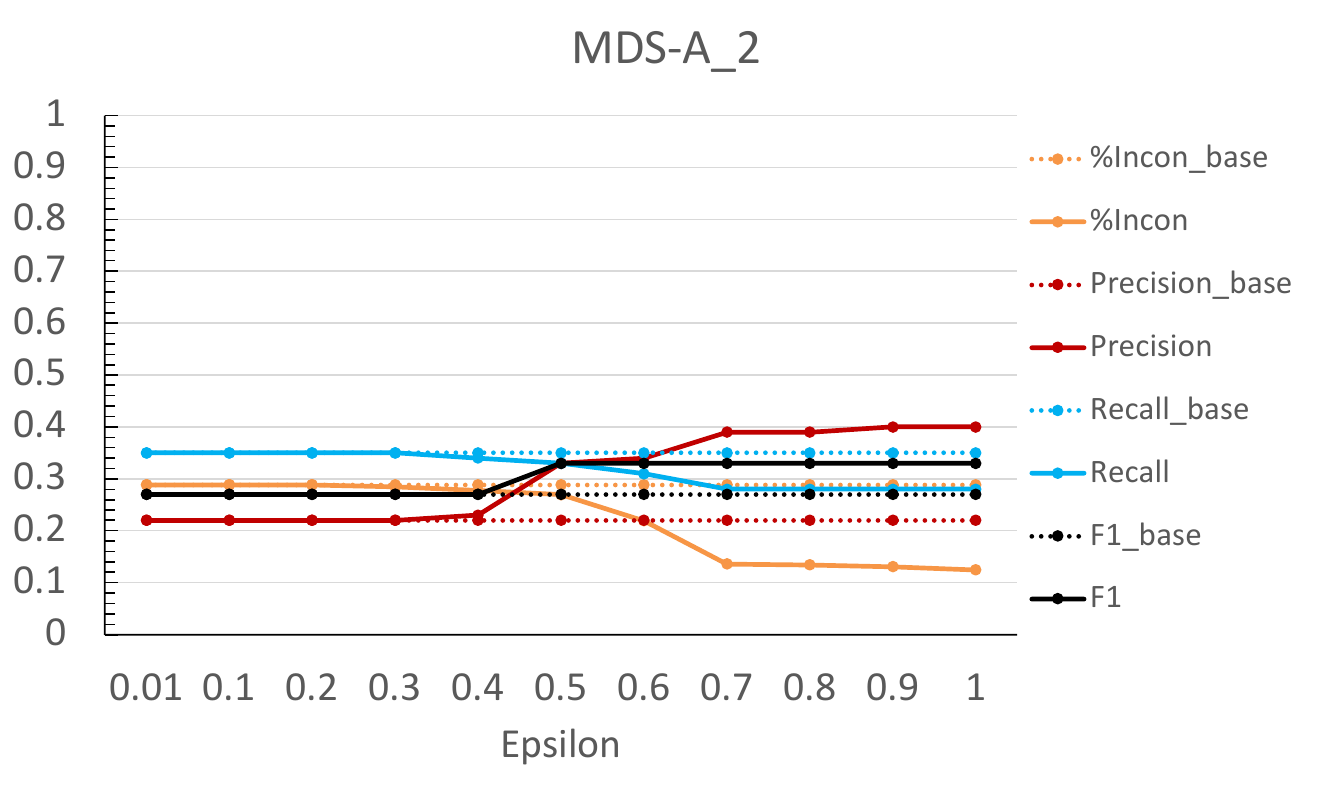}
    \end{subfigure}
    \hfill
    \begin{subfigure}[b]{0.31\textwidth}
        \centering
        \includegraphics[width=\linewidth]{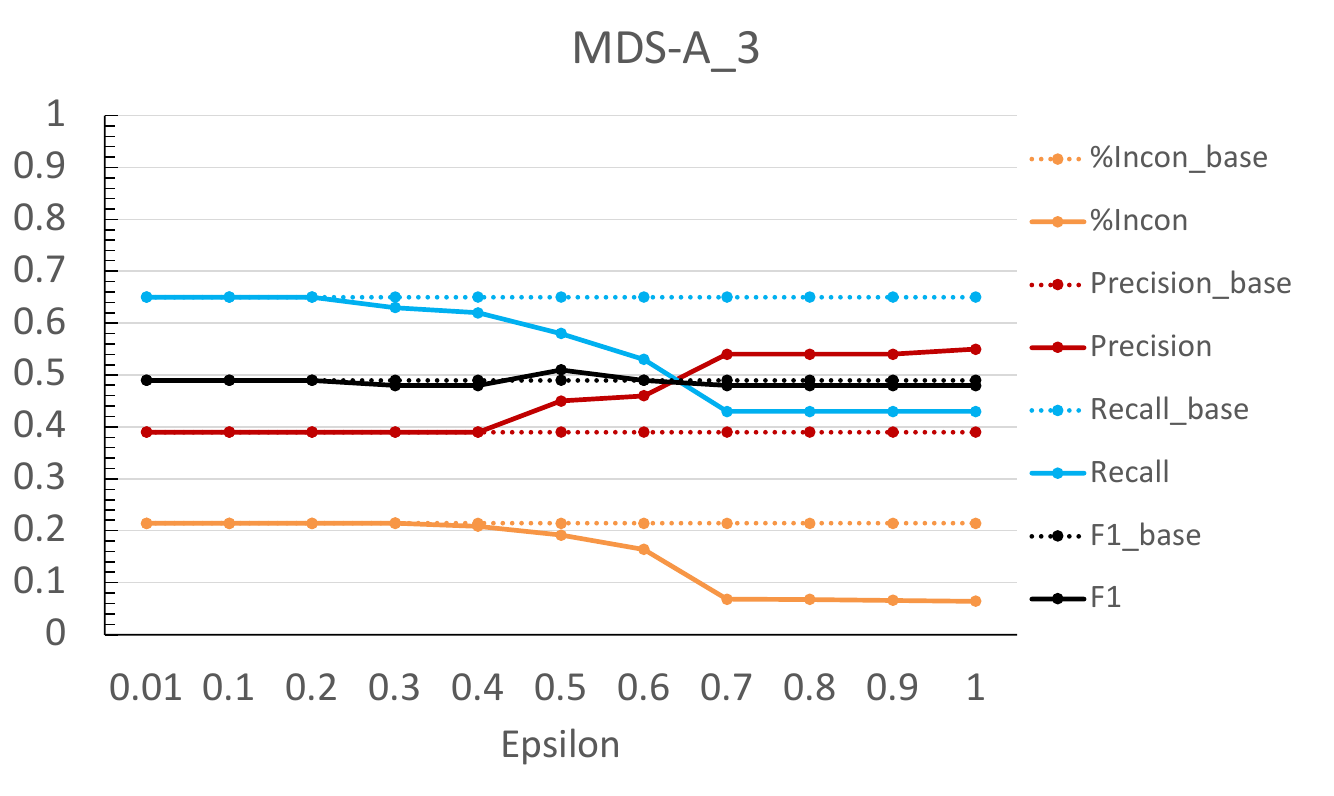}
    \end{subfigure}

    \vspace{\baselineskip} 

    \begin{subfigure}[b]{0.31\textwidth}
        \centering
        \includegraphics[width=\linewidth]{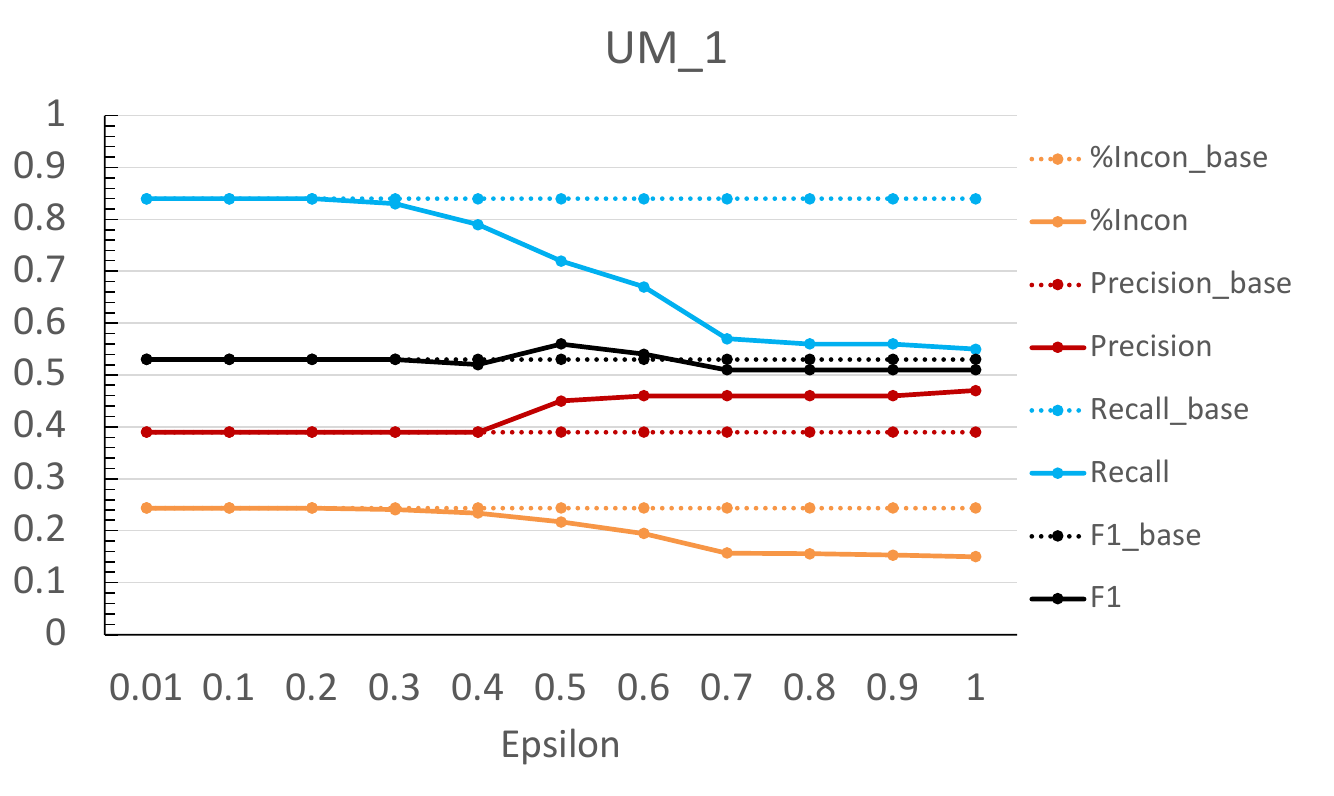}
    \end{subfigure}
    \hfill
    \begin{subfigure}[b]{0.31\textwidth}
        \centering
        \includegraphics[width=\linewidth]{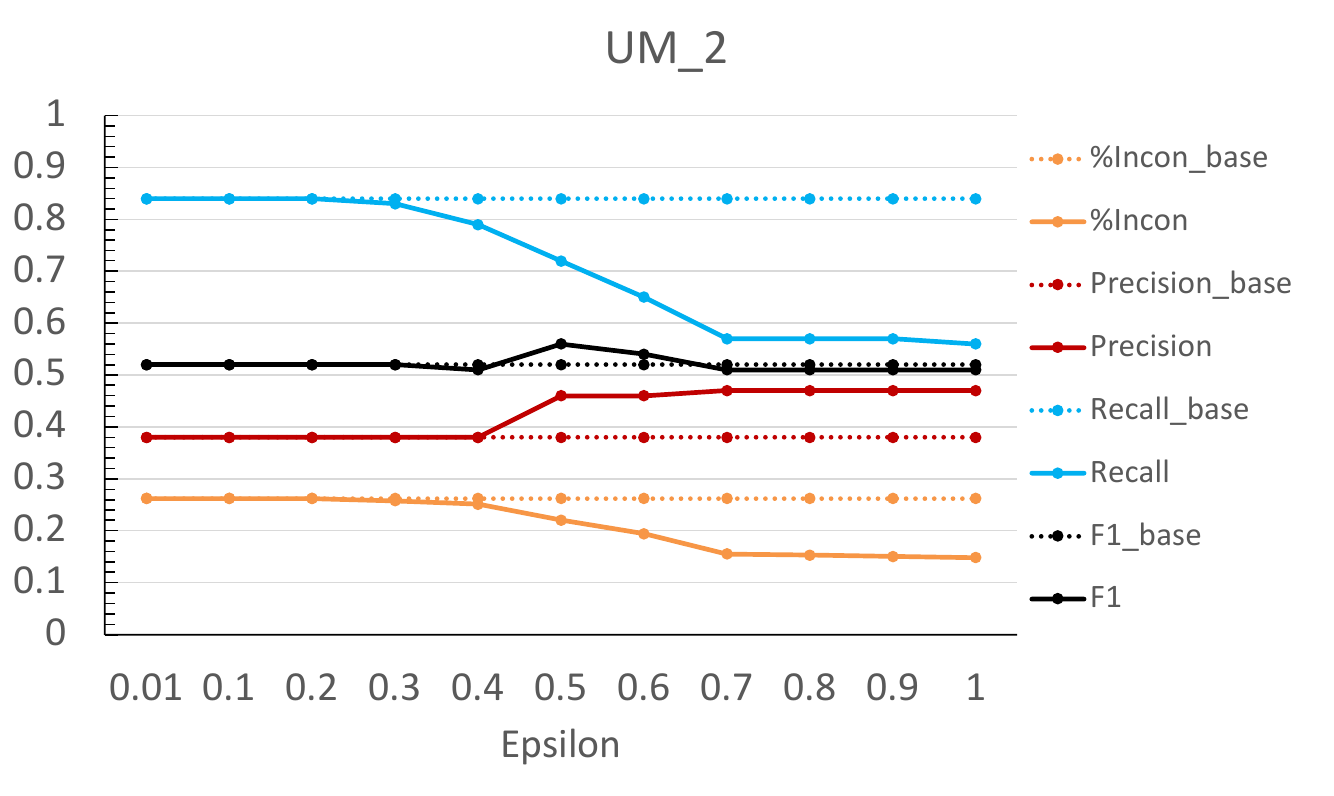}
    \end{subfigure}
    \hfill
    \begin{subfigure}[b]{0.31\textwidth}
        \centering
        \includegraphics[width=\linewidth]{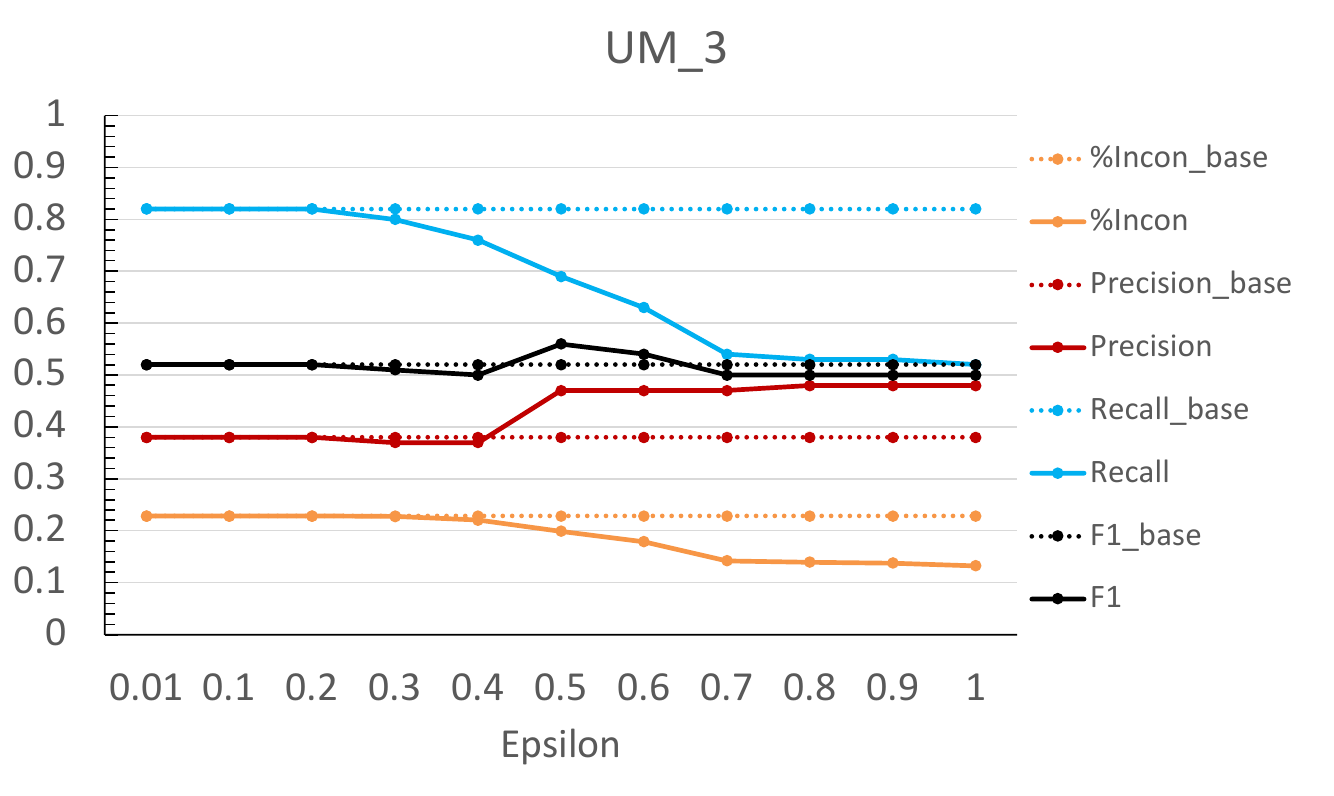}
    \end{subfigure}

    \vspace{\baselineskip}

    \begin{subfigure}[b]{0.31\textwidth}
        \centering
        \includegraphics[width=\linewidth]{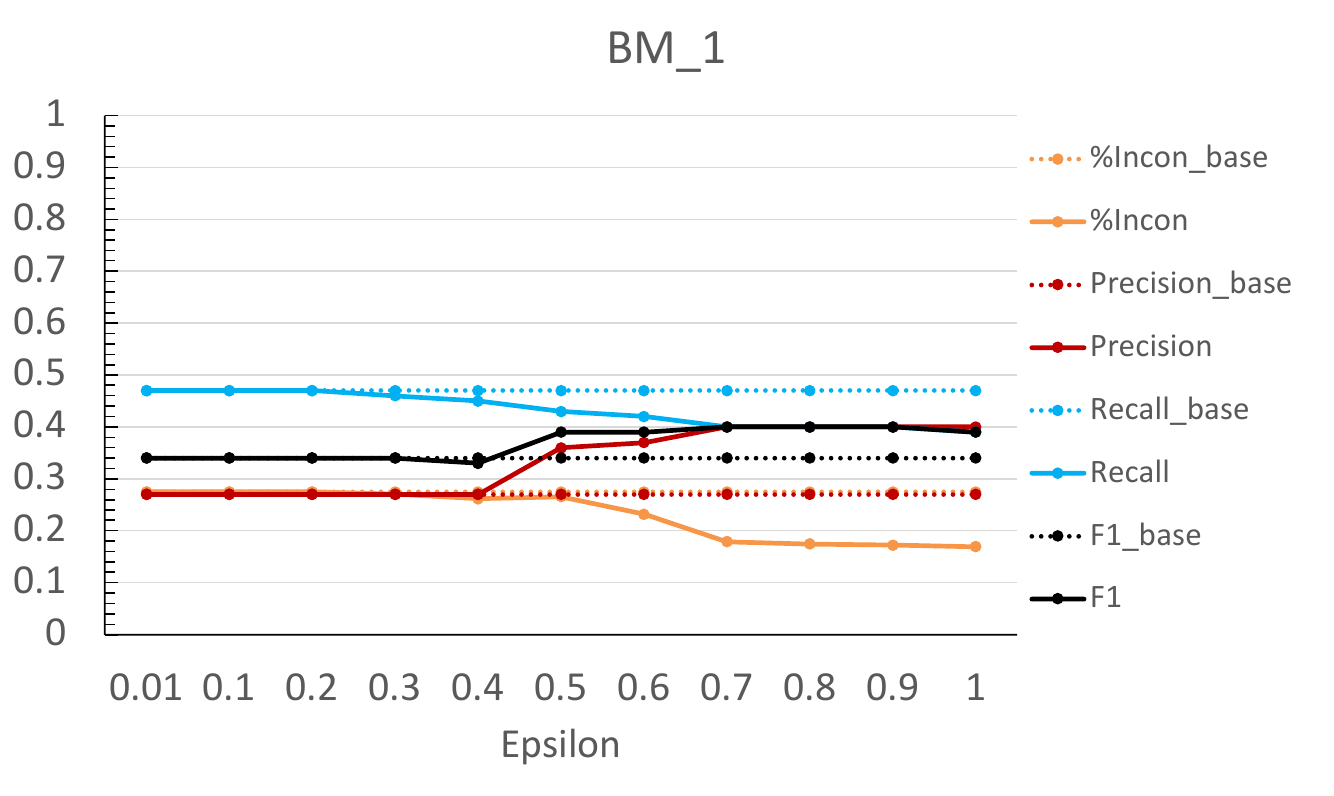}
    \end{subfigure}
    \hfill
    \begin{subfigure}[b]{0.31\textwidth}
        \centering
        \includegraphics[width=\linewidth]{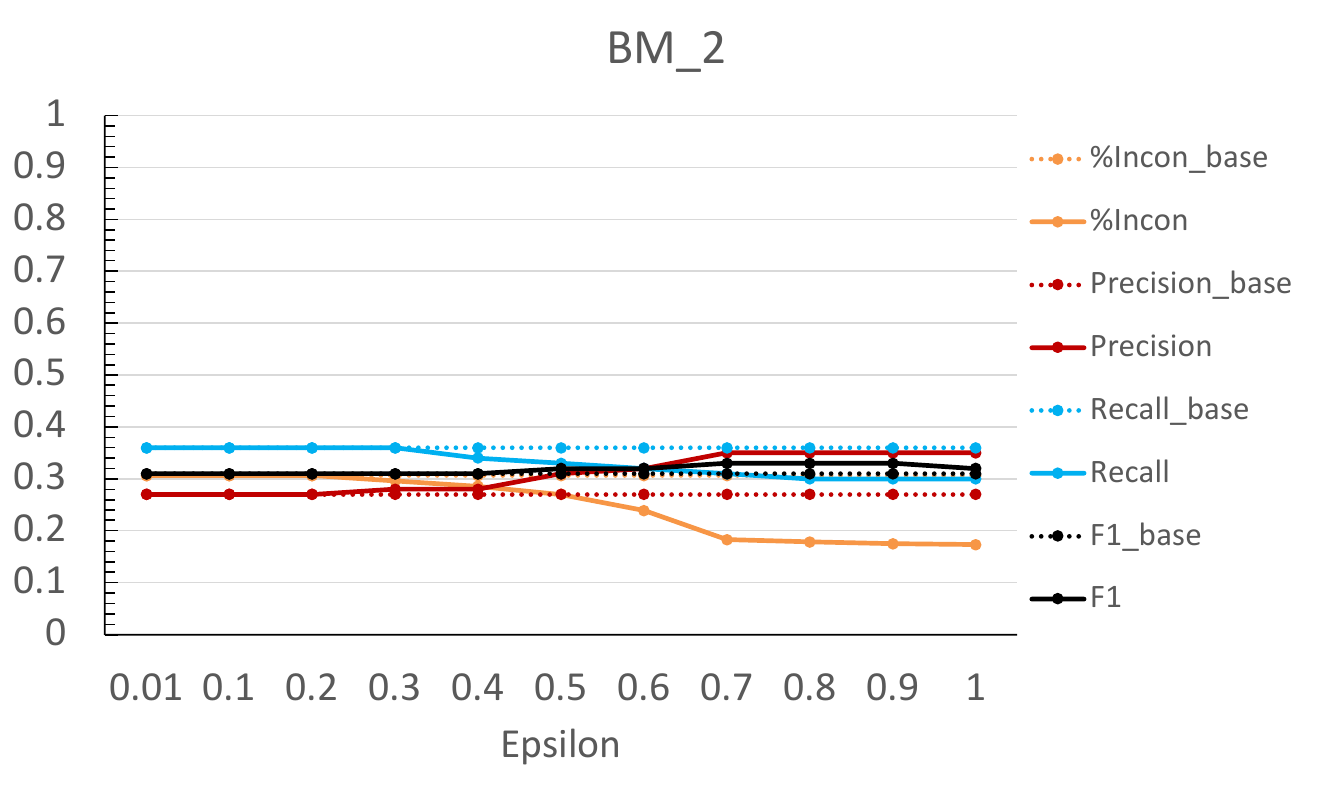}
    \end{subfigure}
    \hfill
    \begin{subfigure}[b]{0.31\textwidth}
        \centering
        \includegraphics[width=\linewidth]{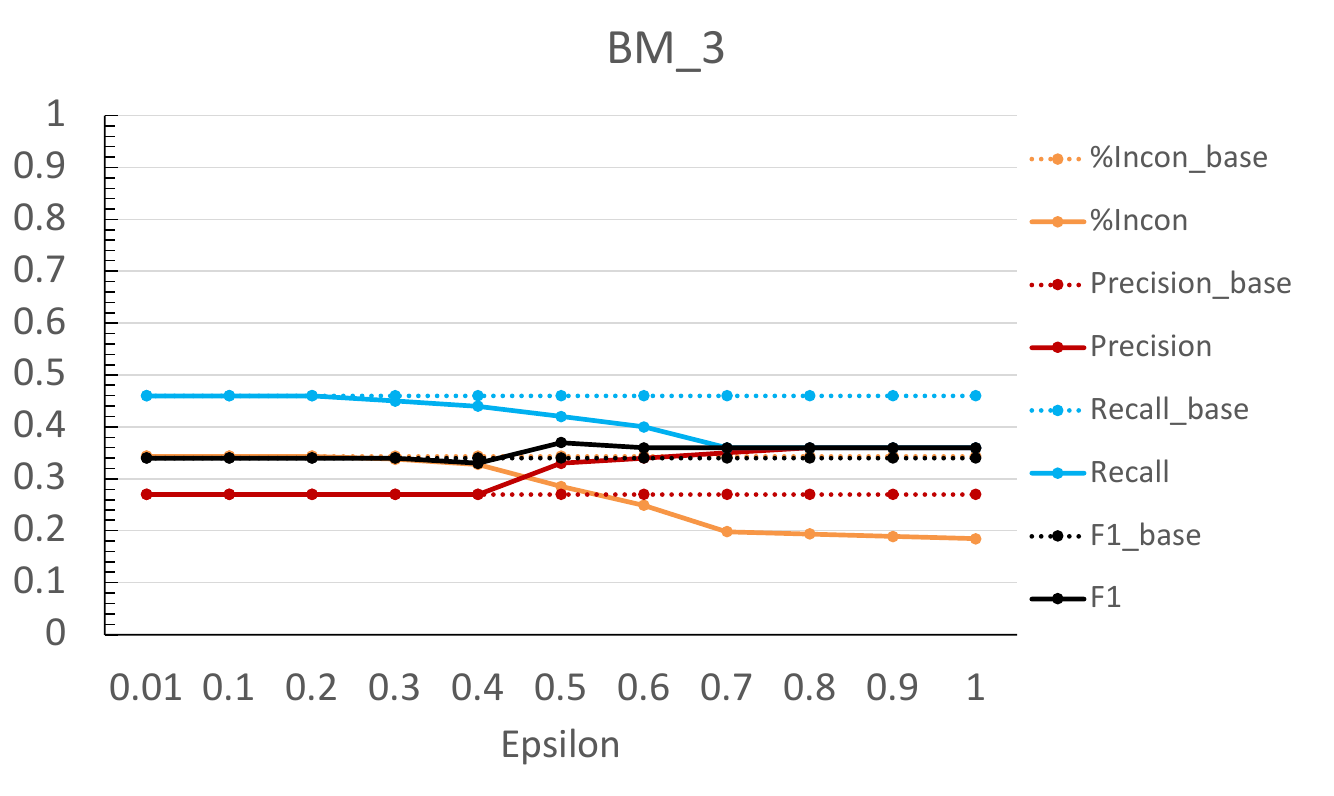}
    \end{subfigure}

    \vspace{\baselineskip}

    \begin{subfigure}[b]{0.31\textwidth}
        \centering
        \includegraphics[width=\linewidth]{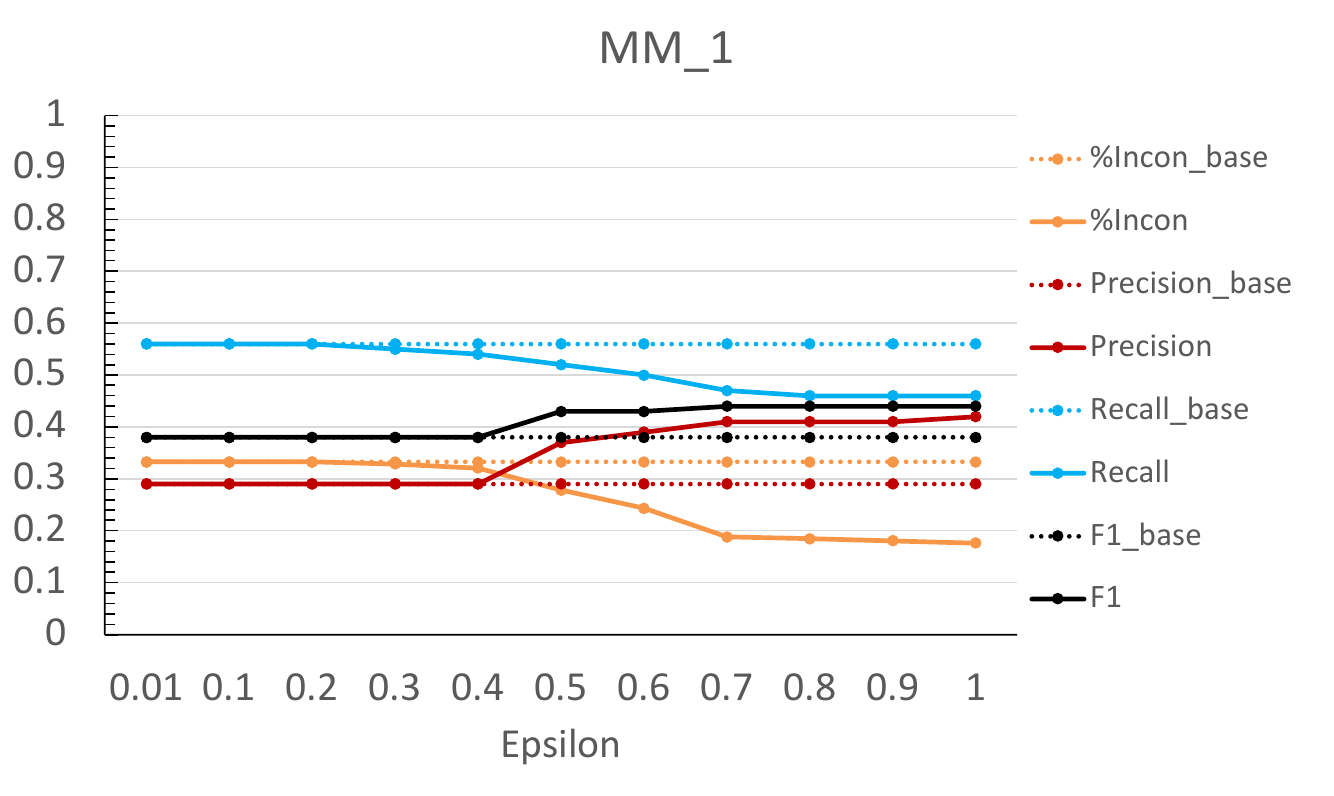}
    \end{subfigure}
    \hfill
    \begin{subfigure}[b]{0.31\textwidth}
        \centering
        \includegraphics[width=\linewidth]{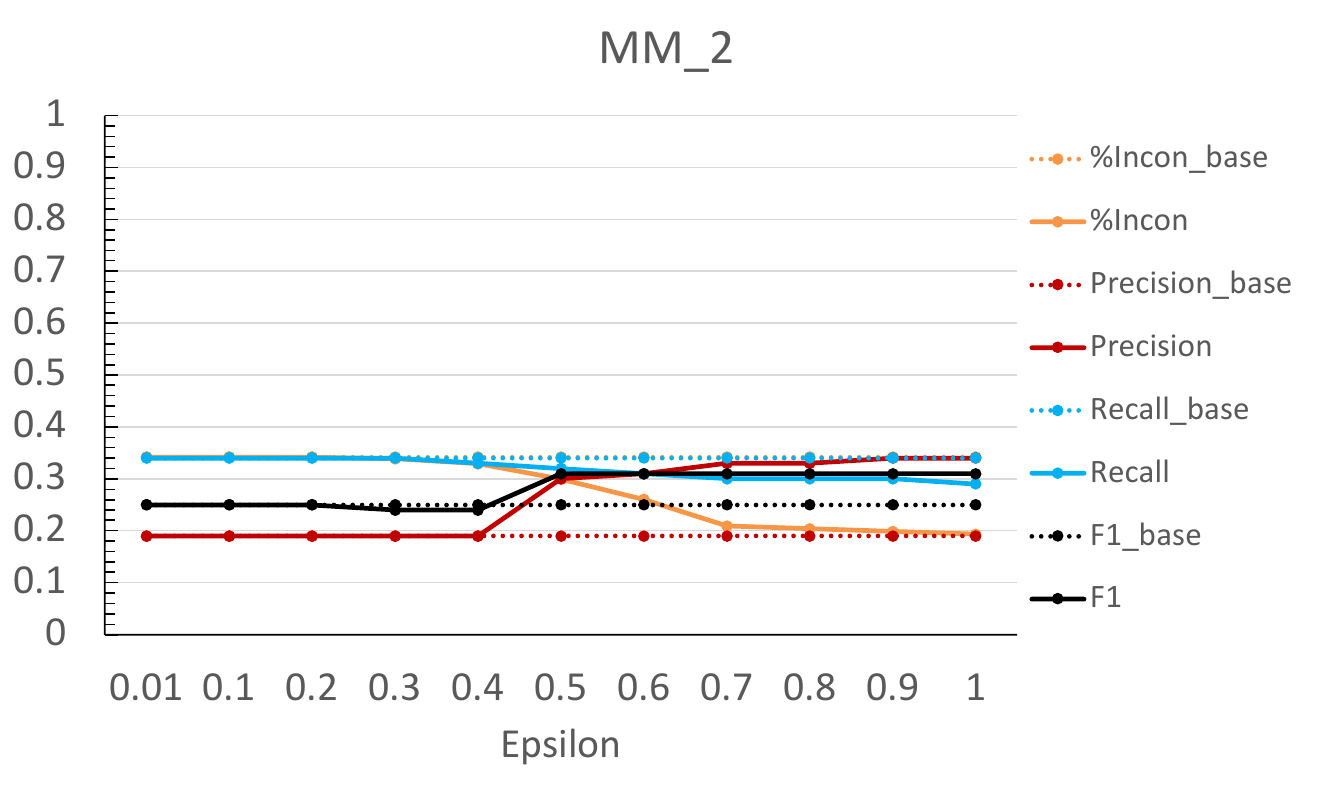}
    \end{subfigure}
    \hfill
    \begin{subfigure}[b]{0.31\textwidth}
        \centering
        \includegraphics[width=\linewidth]{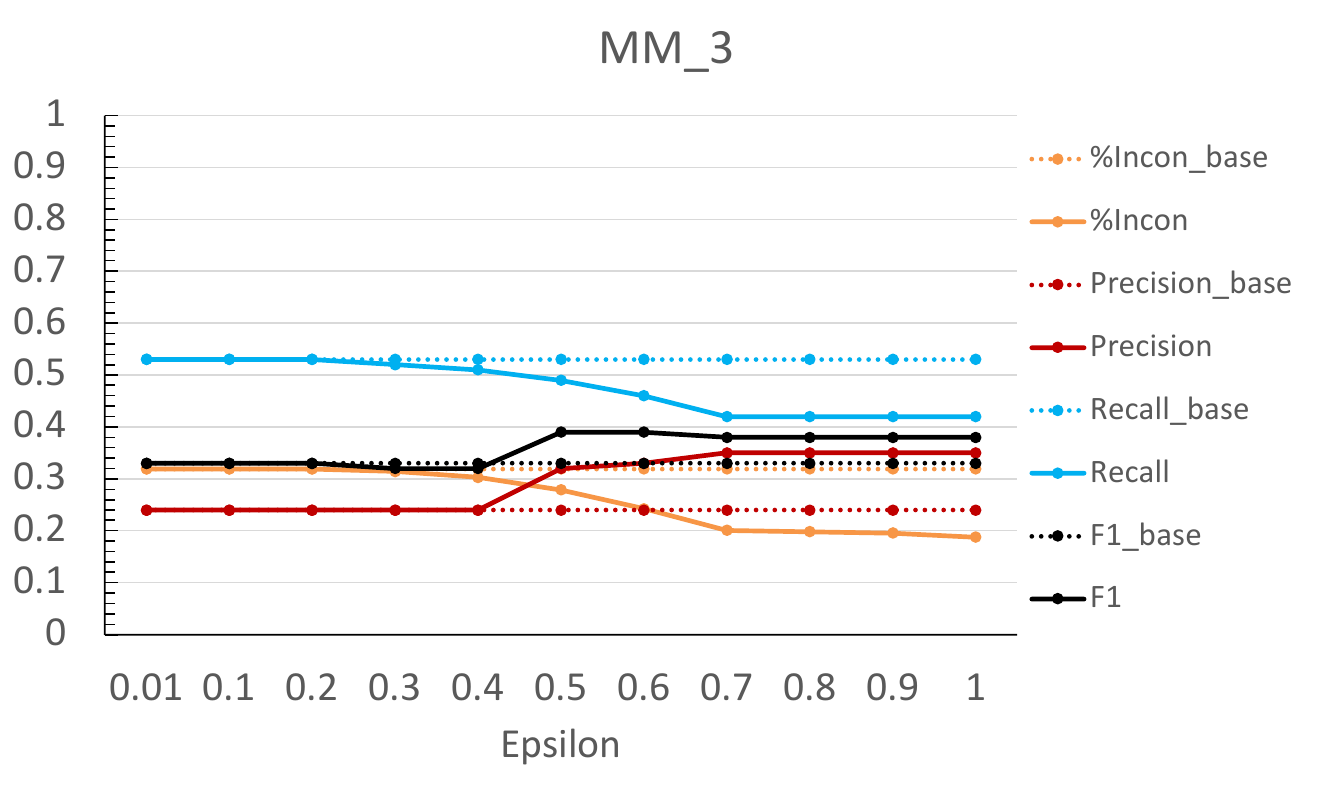}
    \end{subfigure}

    \vspace{\baselineskip}

    \begin{subfigure}[b]{0.31\textwidth}
        \centering
        \includegraphics[width=\linewidth]{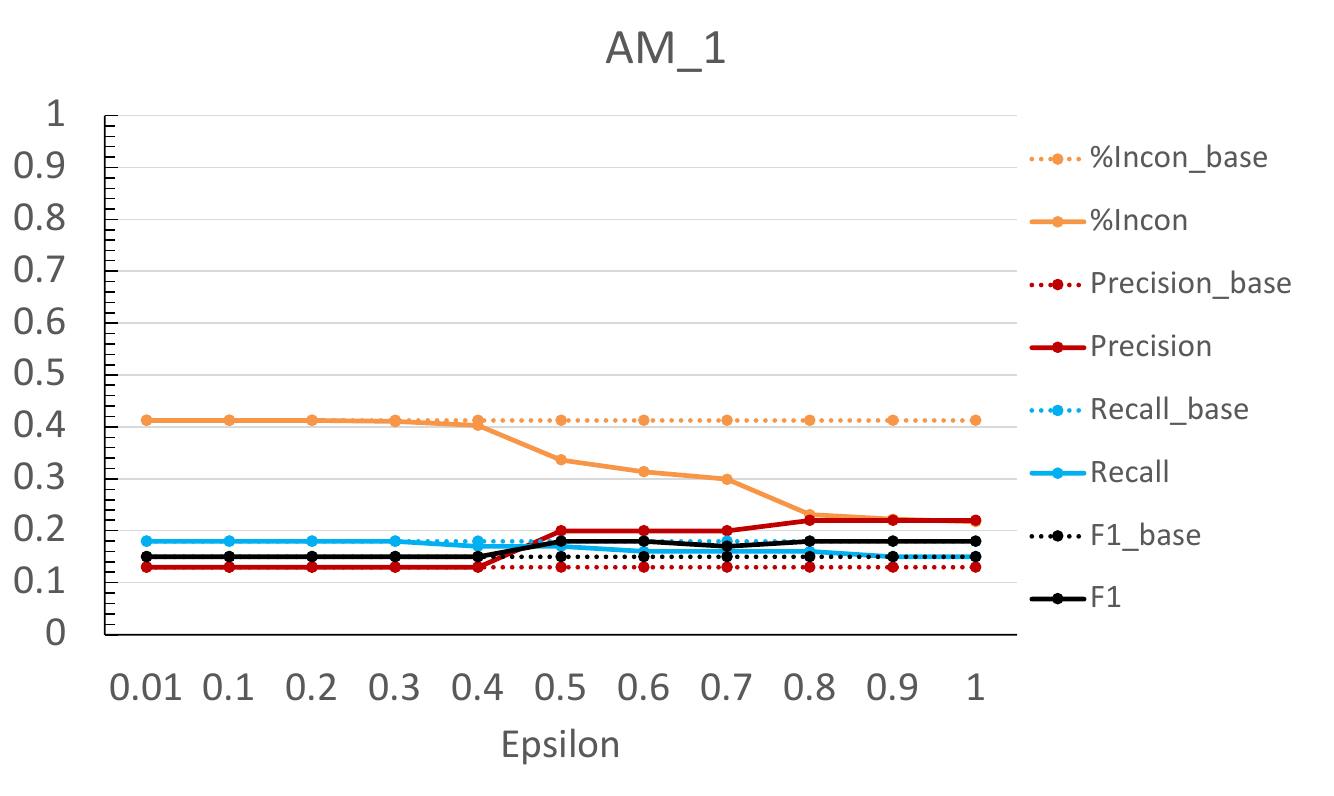}
    \end{subfigure}
    \hfill
    \begin{subfigure}[b]{0.31\textwidth}
        \centering
        \includegraphics[width=\linewidth]{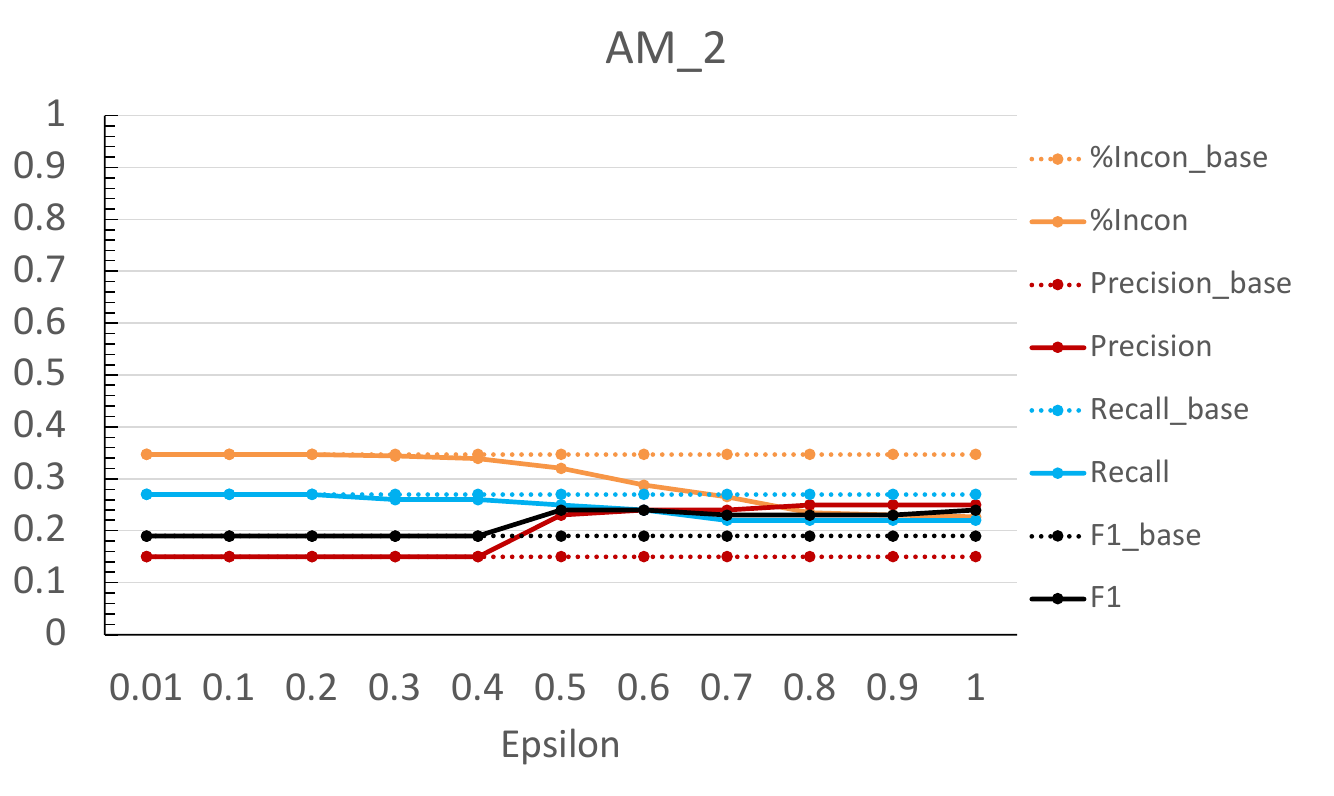}
    \end{subfigure}
    \hfill
    \begin{subfigure}[b]{0.31\textwidth}
        \centering
        \includegraphics[width=\linewidth]{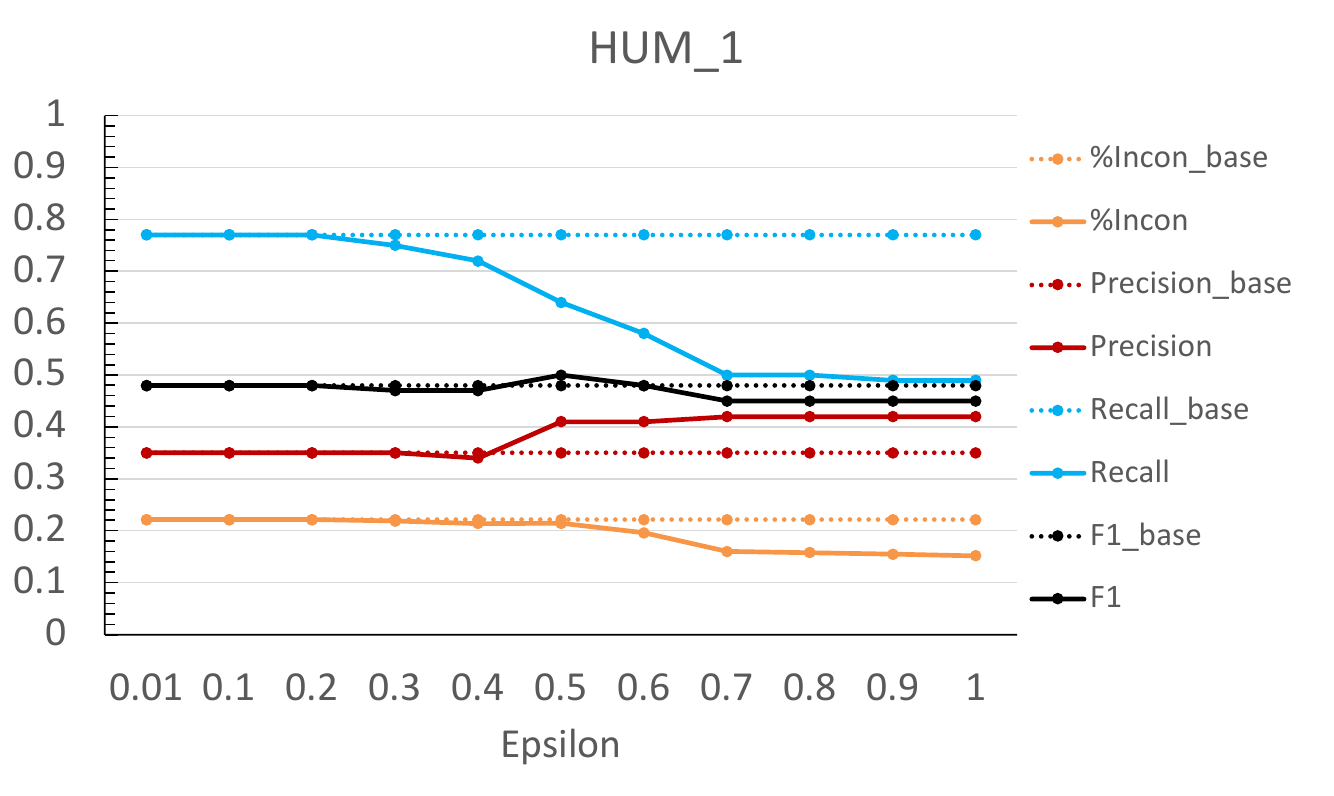}
    \end{subfigure}
    \caption{Detailed EDR $\epsilon$ sensitivity per test sets}
    \label{fig:eps_sens}
\end{figure}

\clearpage
\section{Detailed HS+TB Delta ($\delta$) Sensitivity Analysis per Test Set}
\label{sec:supp_hs_delta}

This section details the sensitivity of the Heuristic Search with Tie-Breaker (HS+TB) method to the maximum inconsistency threshold, $\delta$, for all 15 test datasets. Each figure shows how key performance metrics (e.g., accuracy-score, Accuracy, final inconsistency rate) for HS+TB change as $\delta$ varies. This expands on Figure 5 (top-right) of the main paper.

\begin{figure}[h!]
    \centering
    \begin{subfigure}[b]{0.31\textwidth}
        \centering
        \includegraphics[width=\linewidth]{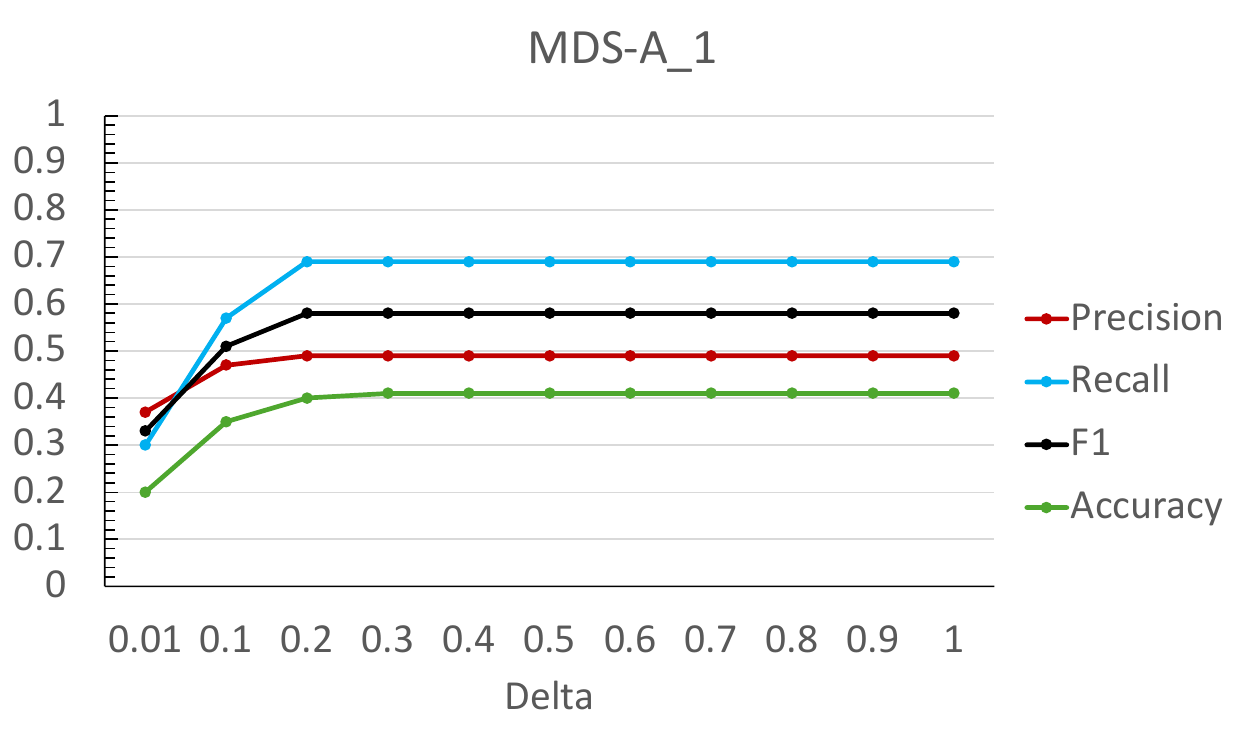}
    \end{subfigure}
    \hfill 
    \begin{subfigure}[b]{0.31\textwidth}
        \centering
        \includegraphics[width=\linewidth]{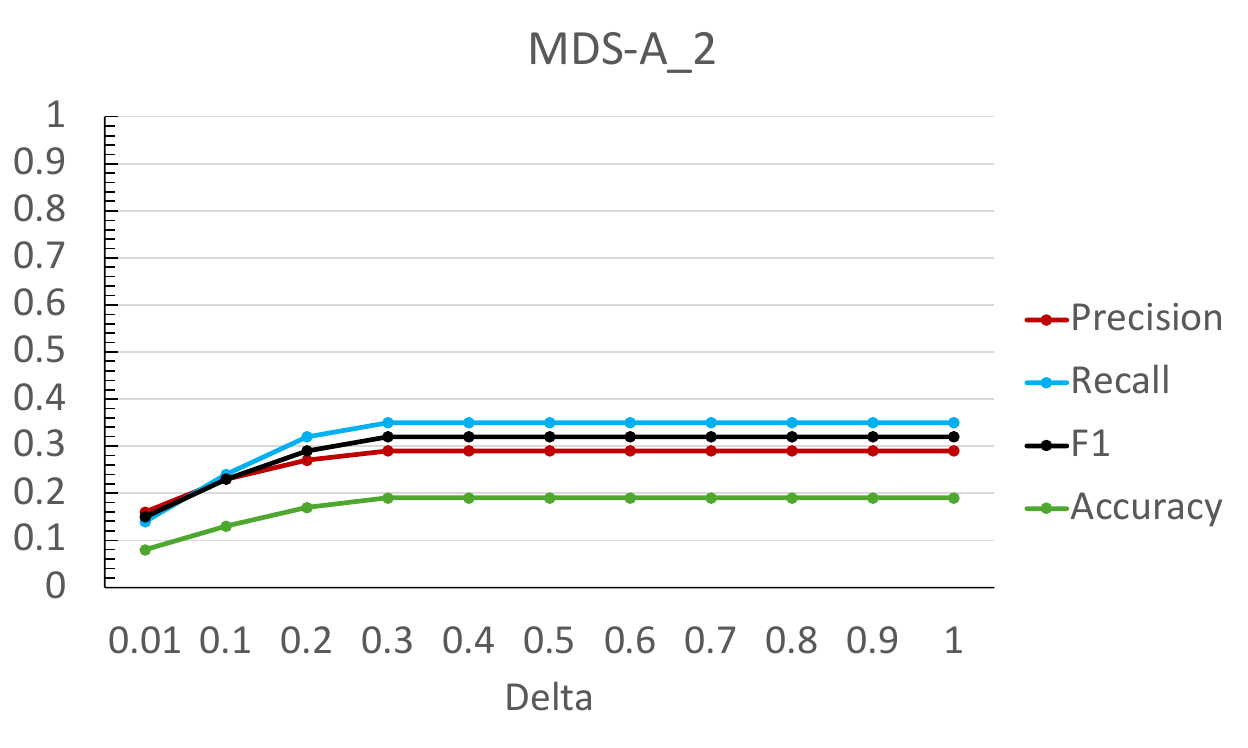}
    \end{subfigure}
    \hfill
    \begin{subfigure}[b]{0.31\textwidth}
        \centering
        \includegraphics[width=\linewidth]{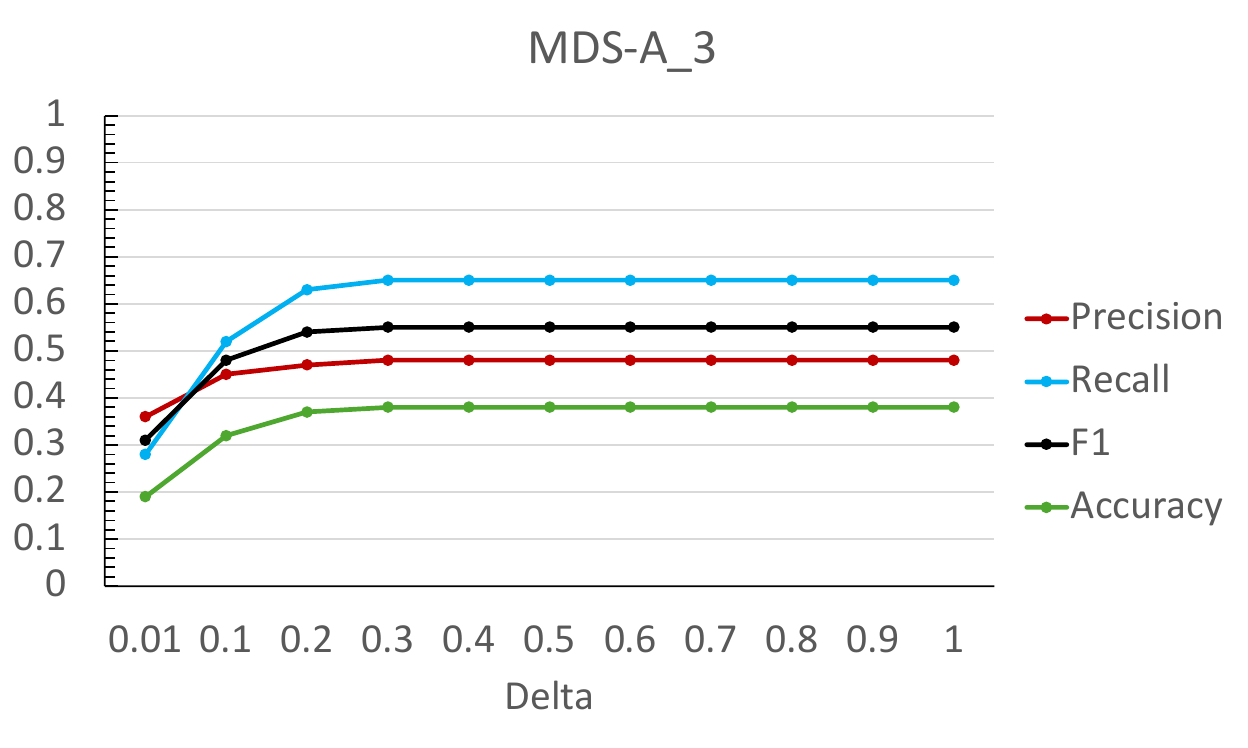}
    \end{subfigure}

    \vspace{\baselineskip} 

    \begin{subfigure}[b]{0.31\textwidth}
        \centering
        \includegraphics[width=\linewidth]{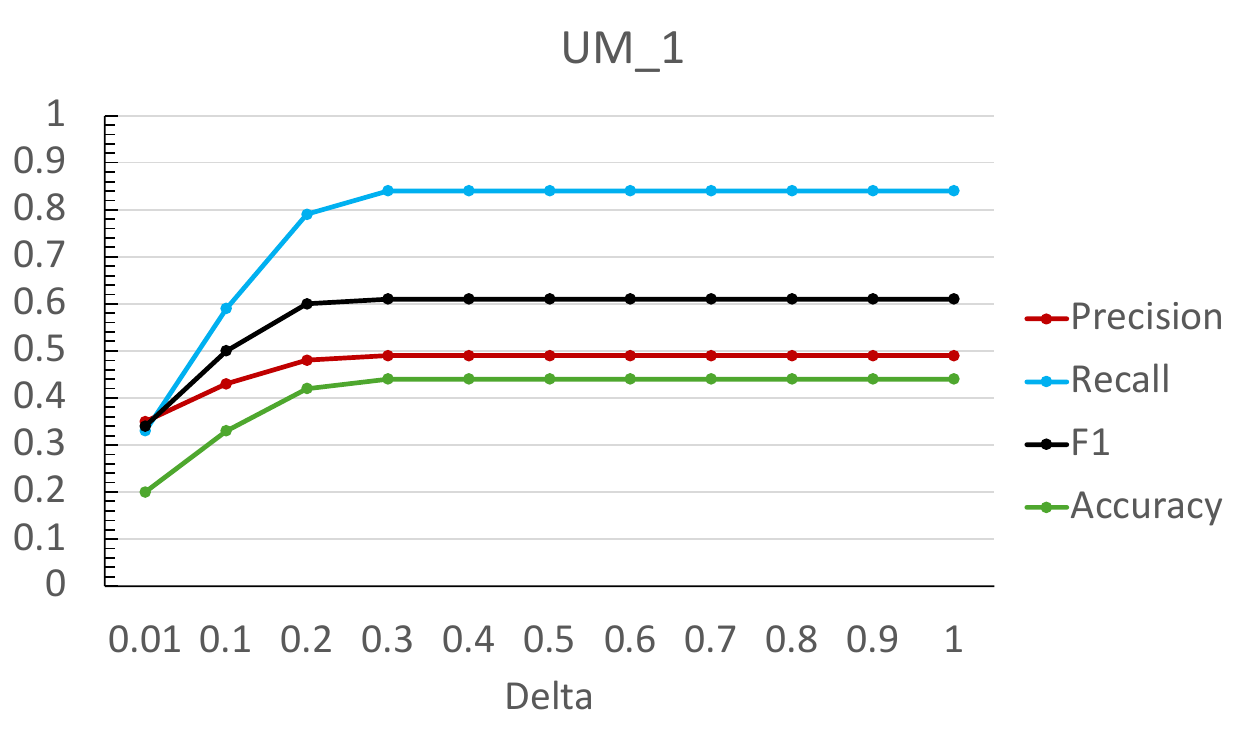}
    \end{subfigure}
    \hfill
    \begin{subfigure}[b]{0.31\textwidth}
        \centering
        \includegraphics[width=\linewidth]{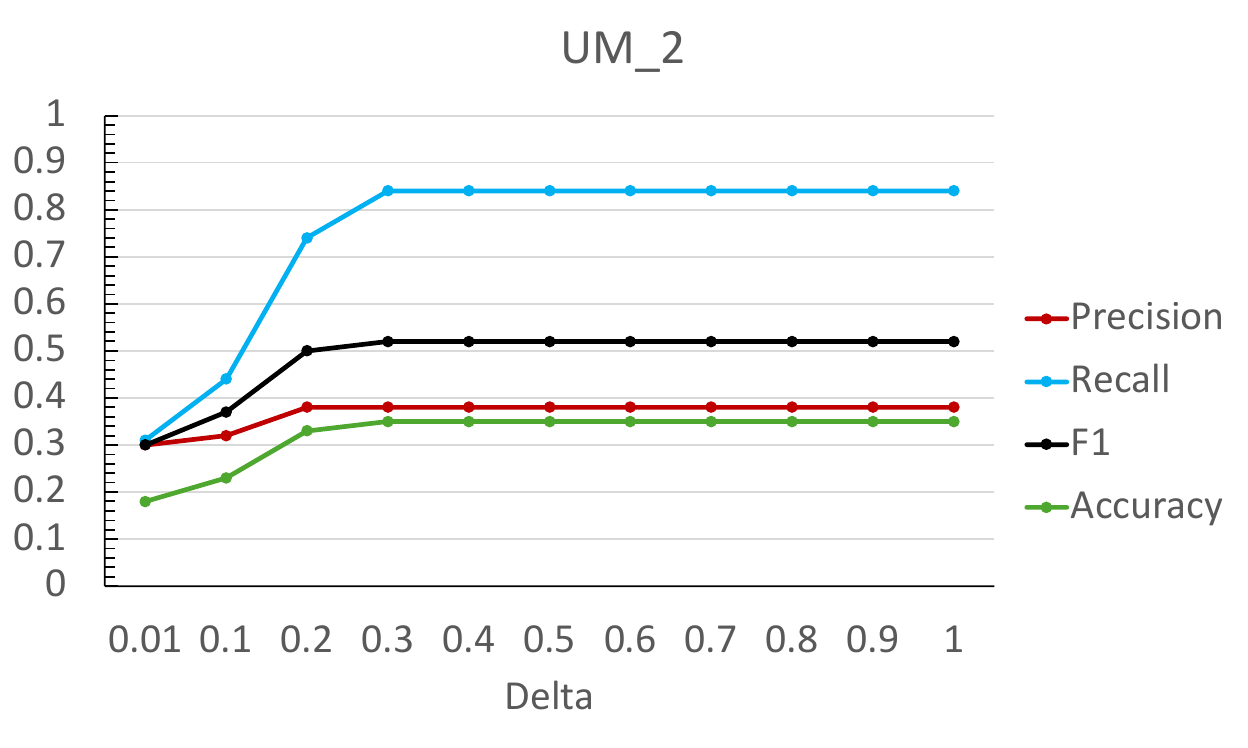}
    \end{subfigure}
    \hfill
    \begin{subfigure}[b]{0.31\textwidth}
        \centering
        \includegraphics[width=\linewidth]{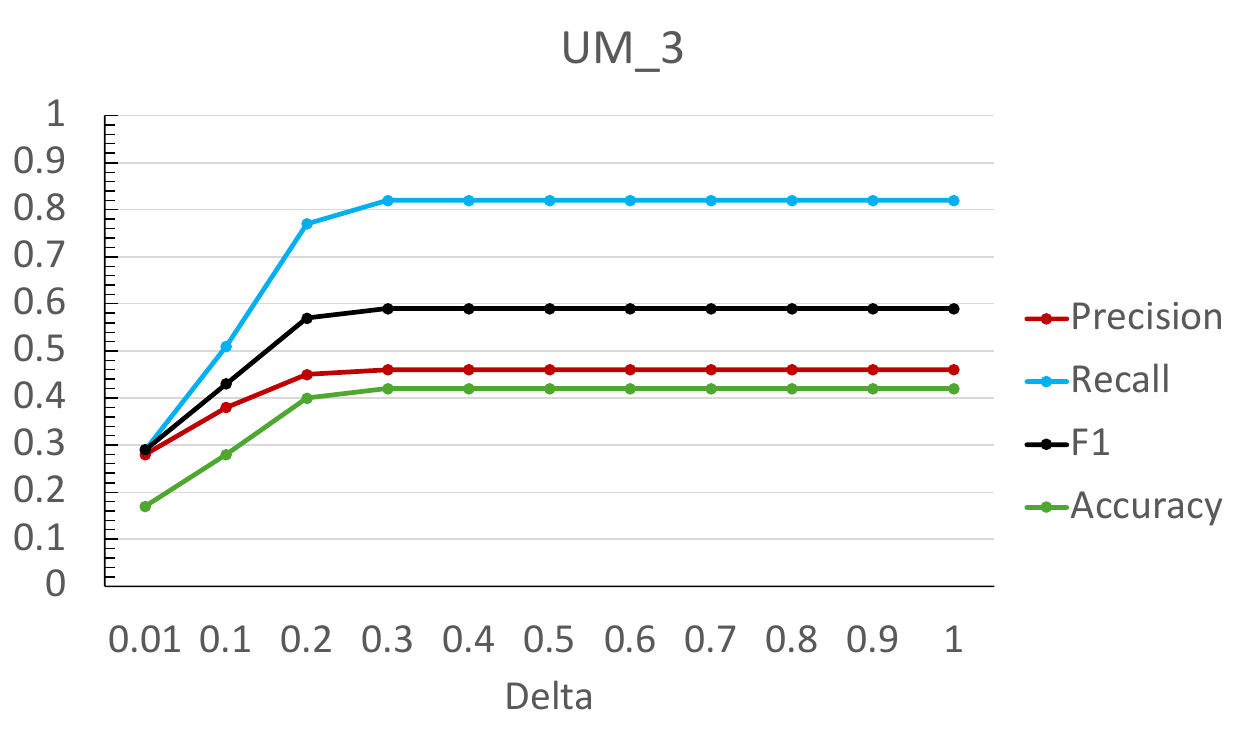}
    \end{subfigure}

    \vspace{\baselineskip}

    \begin{subfigure}[b]{0.31\textwidth}
        \centering
        \includegraphics[width=\linewidth]{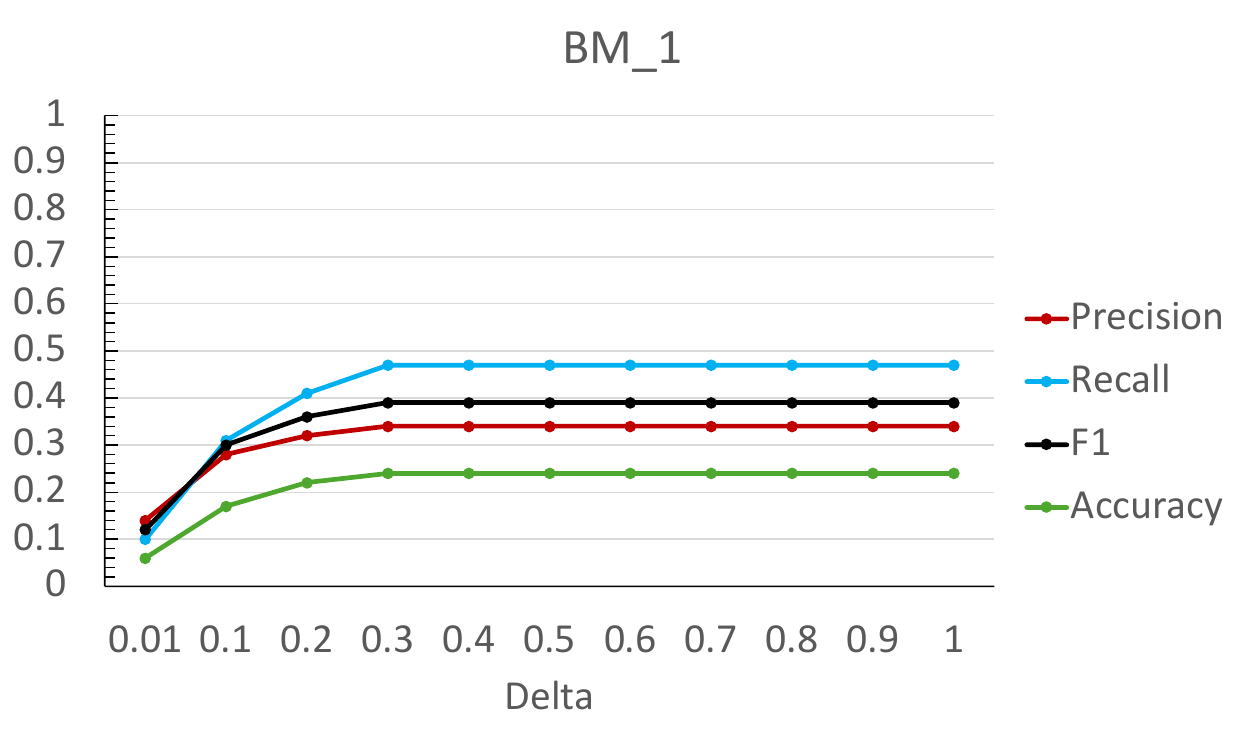}
    \end{subfigure}
    \hfill
    \begin{subfigure}[b]{0.31\textwidth}
        \centering
        \includegraphics[width=\linewidth]{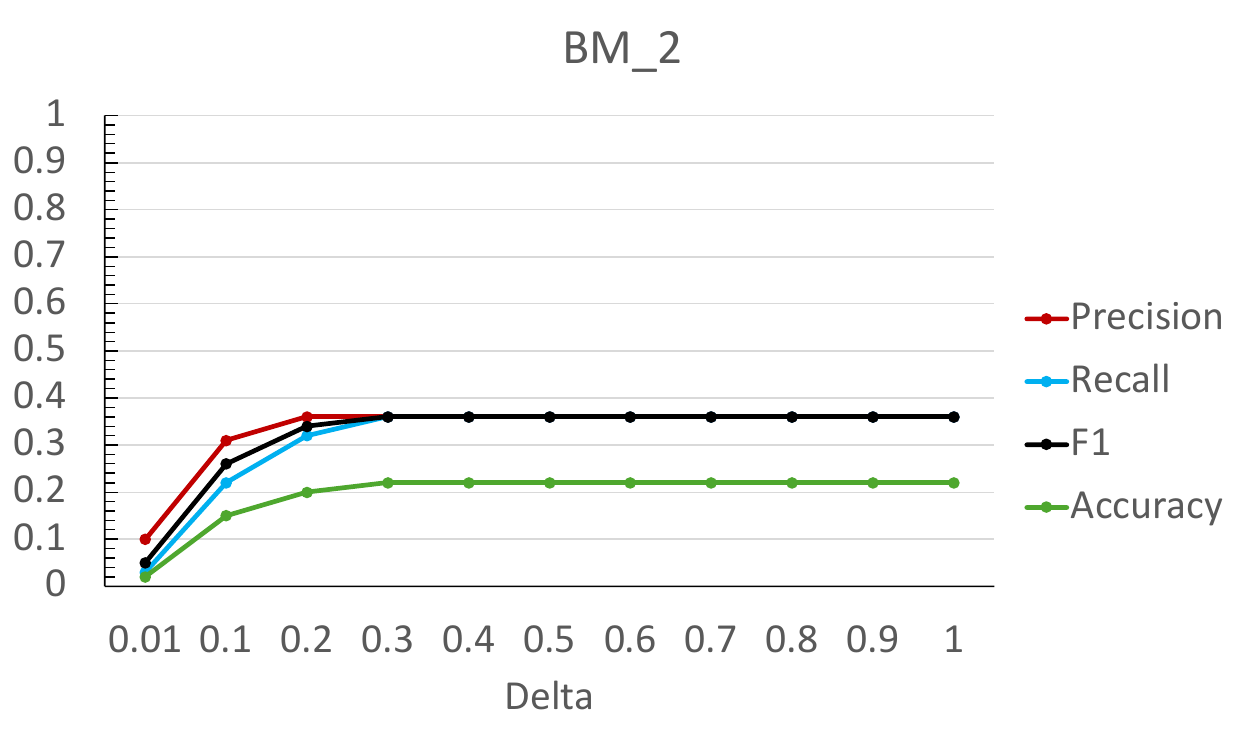}
    \end{subfigure}
    \hfill
    \begin{subfigure}[b]{0.31\textwidth}
        \centering
        \includegraphics[width=\linewidth]{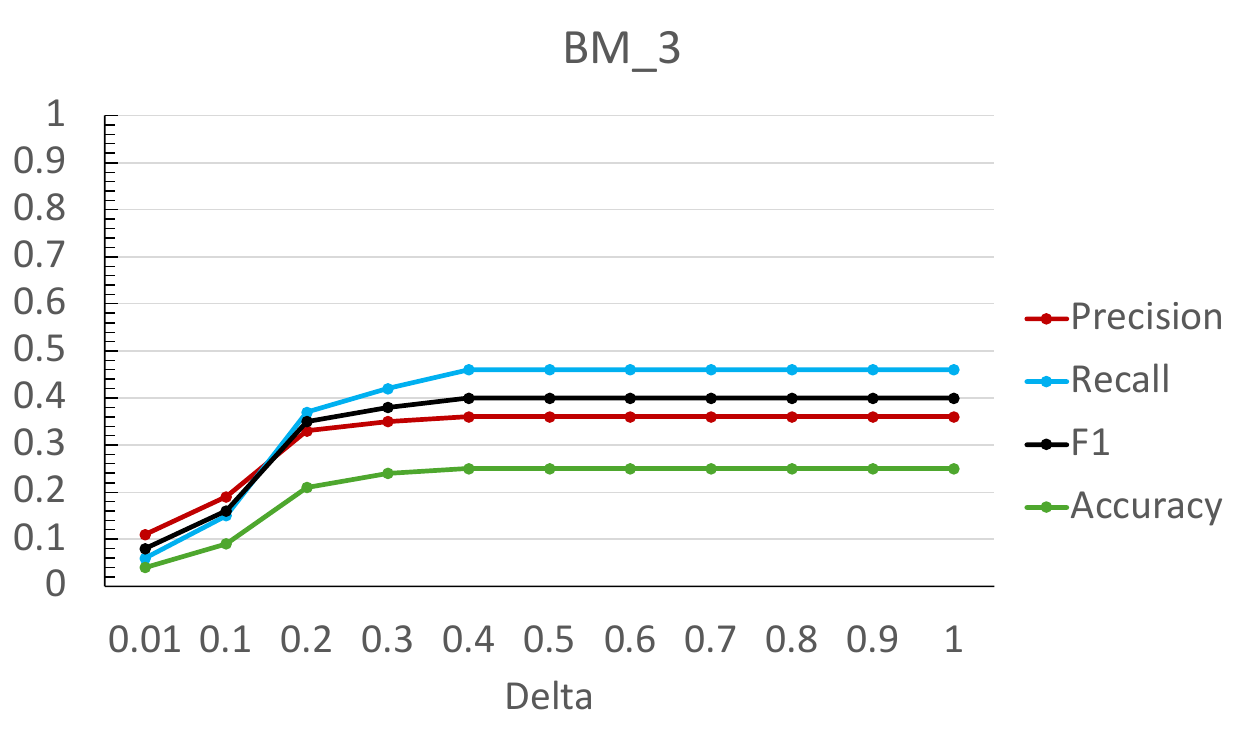}
    \end{subfigure}

    \vspace{\baselineskip}

    \begin{subfigure}[b]{0.31\textwidth}
        \centering
        \includegraphics[width=\linewidth]{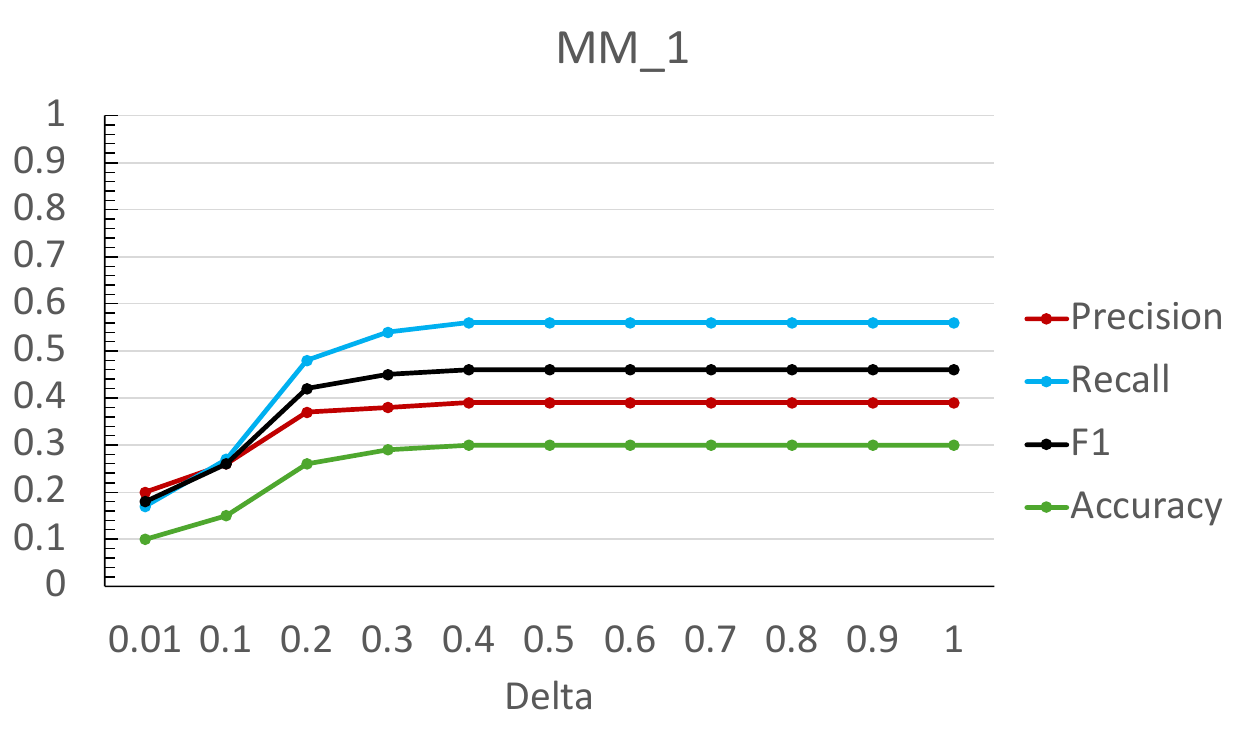}
    \end{subfigure}
    \hfill
    \begin{subfigure}[b]{0.31\textwidth}
        \centering
        \includegraphics[width=\linewidth]{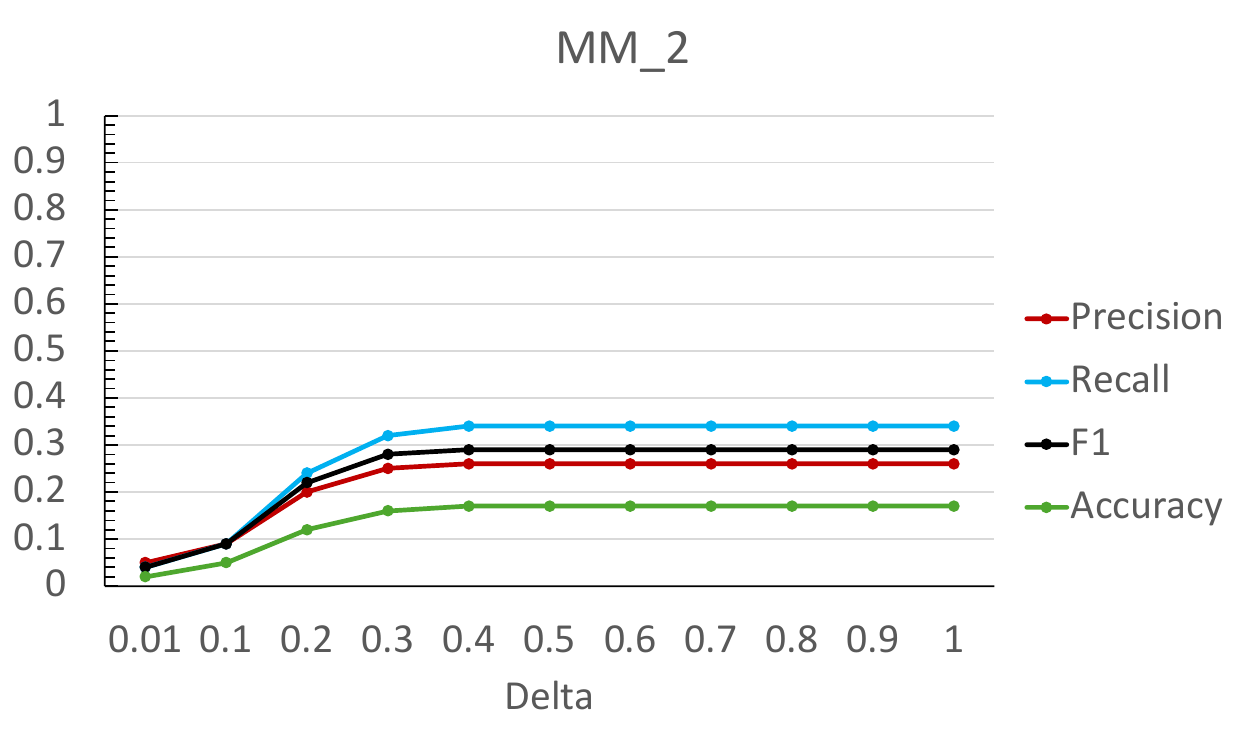}
    \end{subfigure}
    \hfill
    \begin{subfigure}[b]{0.31\textwidth}
        \centering
        \includegraphics[width=\linewidth]{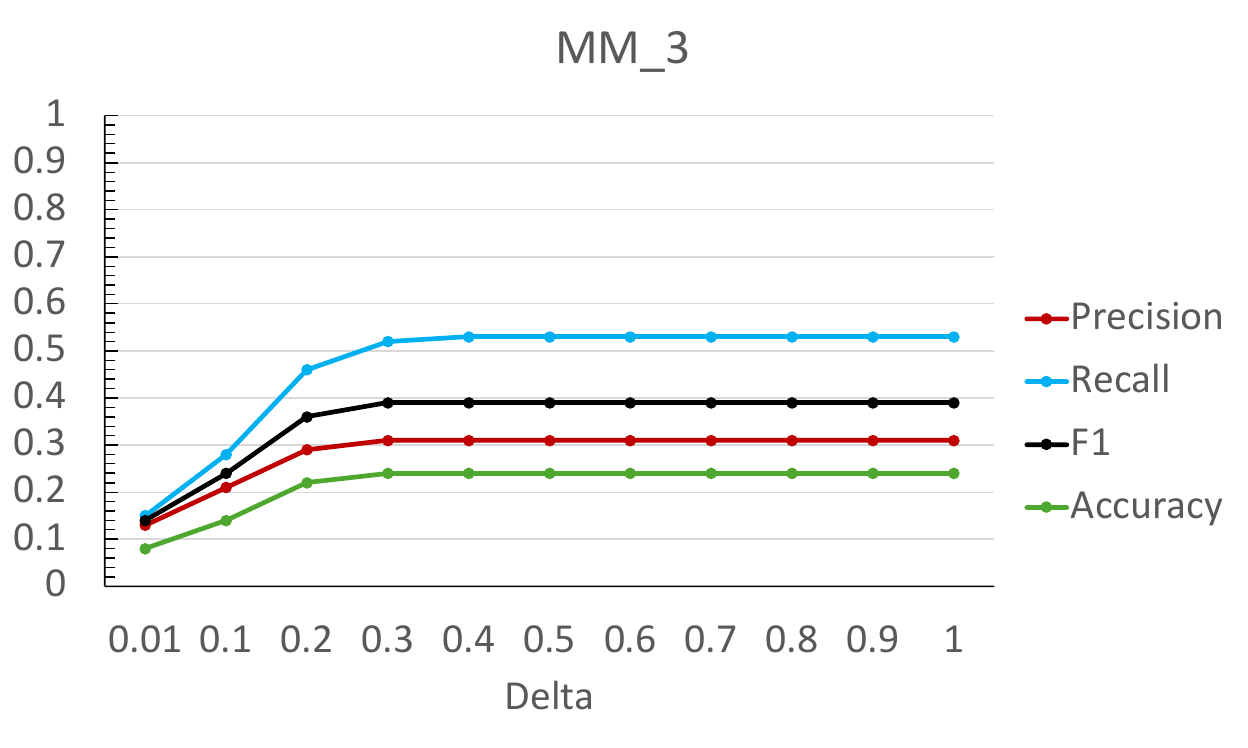}
    \end{subfigure}

    \vspace{\baselineskip}

    \begin{subfigure}[b]{0.31\textwidth}
        \centering
        \includegraphics[width=\linewidth]{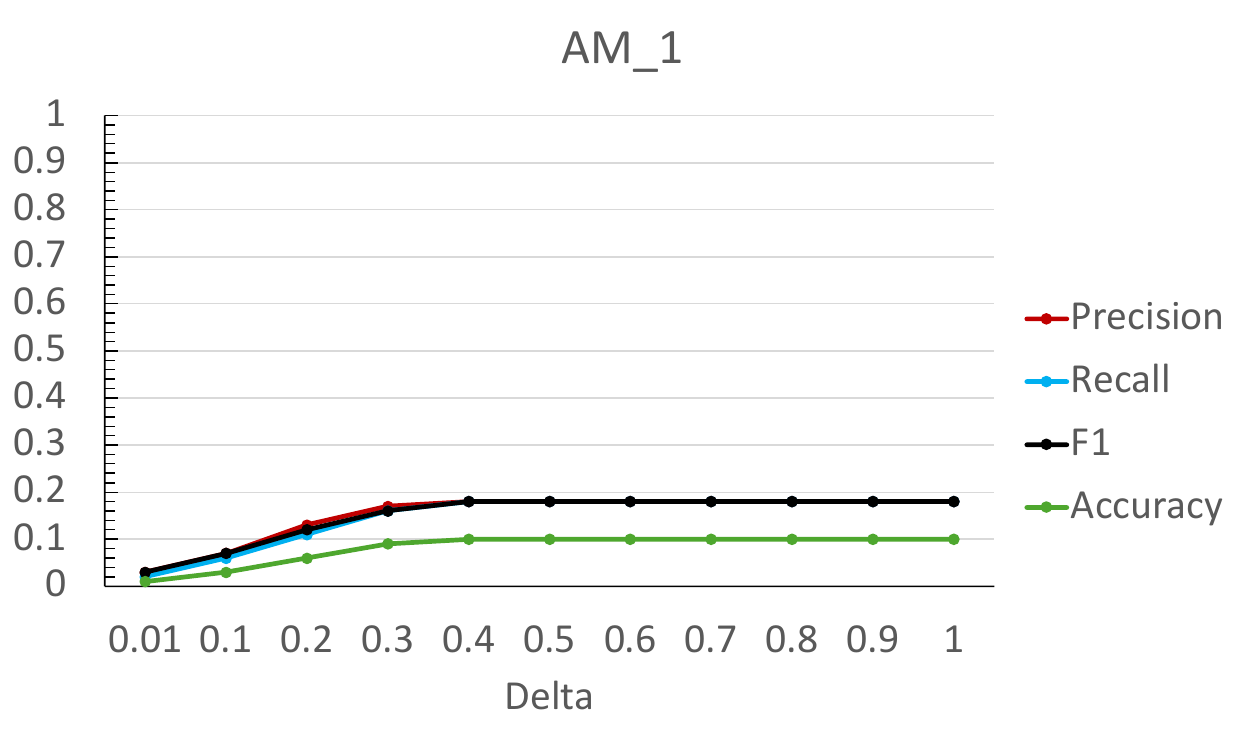}
    \end{subfigure}
    \hfill
    \begin{subfigure}[b]{0.31\textwidth}
        \centering
        \includegraphics[width=\linewidth]{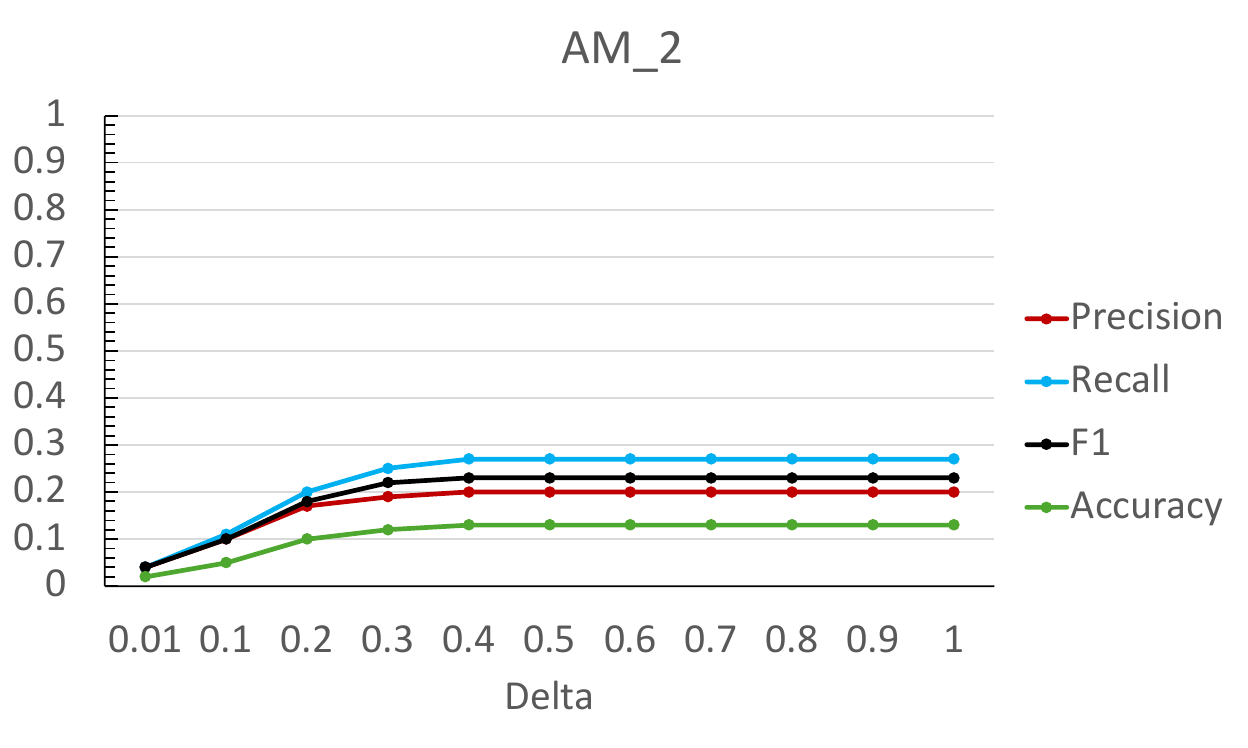}
    \end{subfigure}
    \hfill
    \begin{subfigure}[b]{0.31\textwidth}
        \centering
        \includegraphics[width=\linewidth]{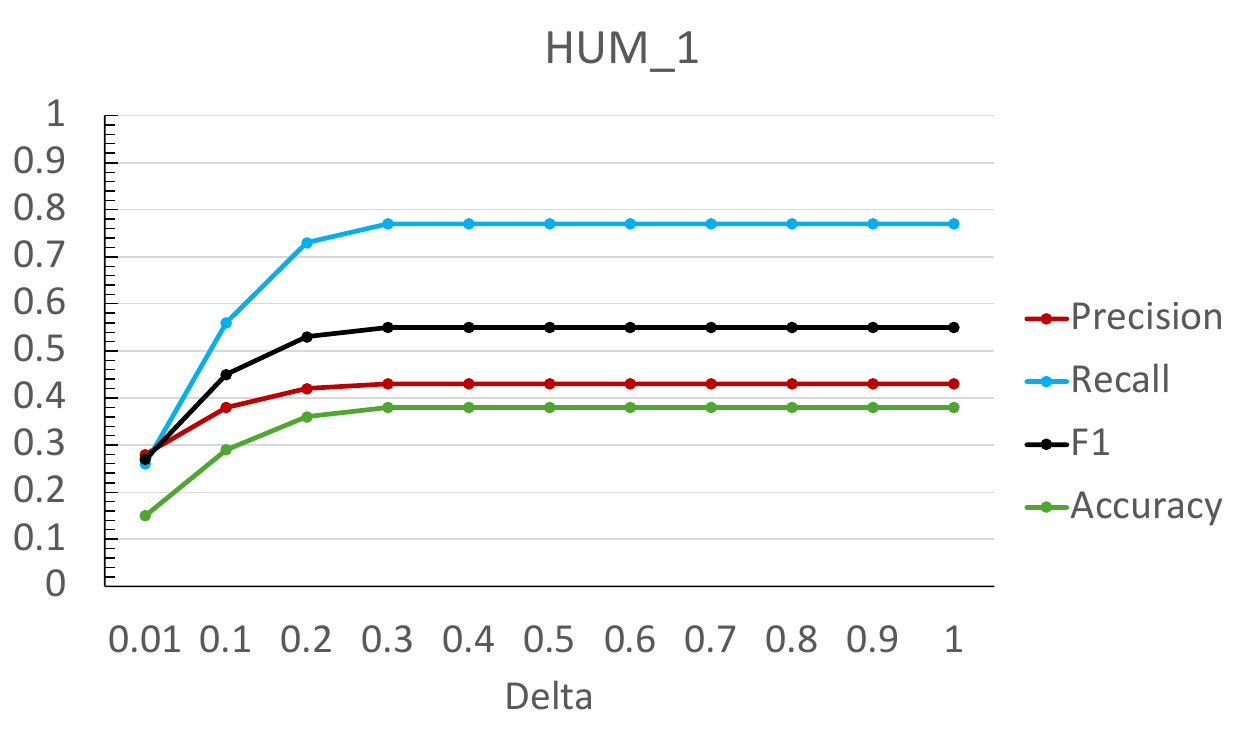}
    \end{subfigure}
    \caption{Detailed HS+TB Delta ($\delta$) sensitivity analysis per test set}
    \label{fig:delta_sens}
\end{figure}

\clearpage

\clearpage
\section{Detailed IP+TB Epsilon-Delta ($\epsilon, \delta$) Sensitivity Analysis per Test Set}
\label{sec:supp_ip_epsilon_delta}

\begin{figure}[htbp!]
    \centering

    \begin{subfigure}[b]{0.19\textwidth} 
        \centering
        \includegraphics[width=\linewidth]{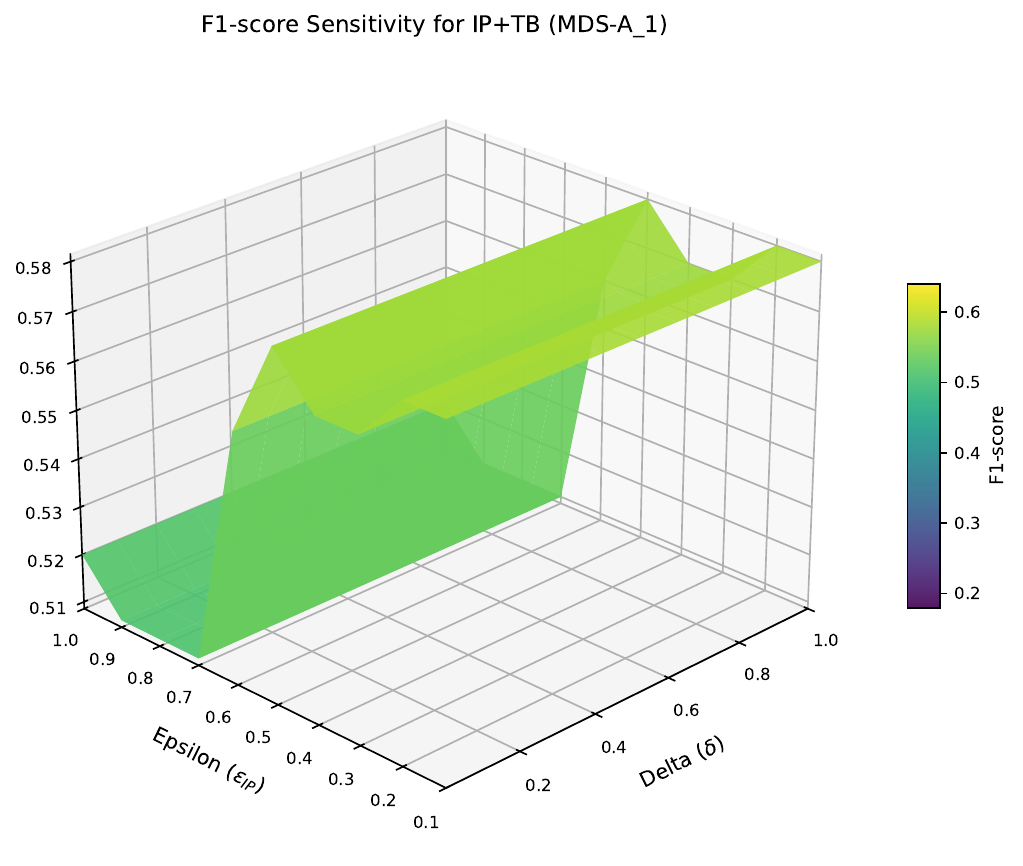}
    \end{subfigure}
    \hfill 
    \begin{subfigure}[b]{0.19\textwidth}
        \centering
        \includegraphics[width=\linewidth]{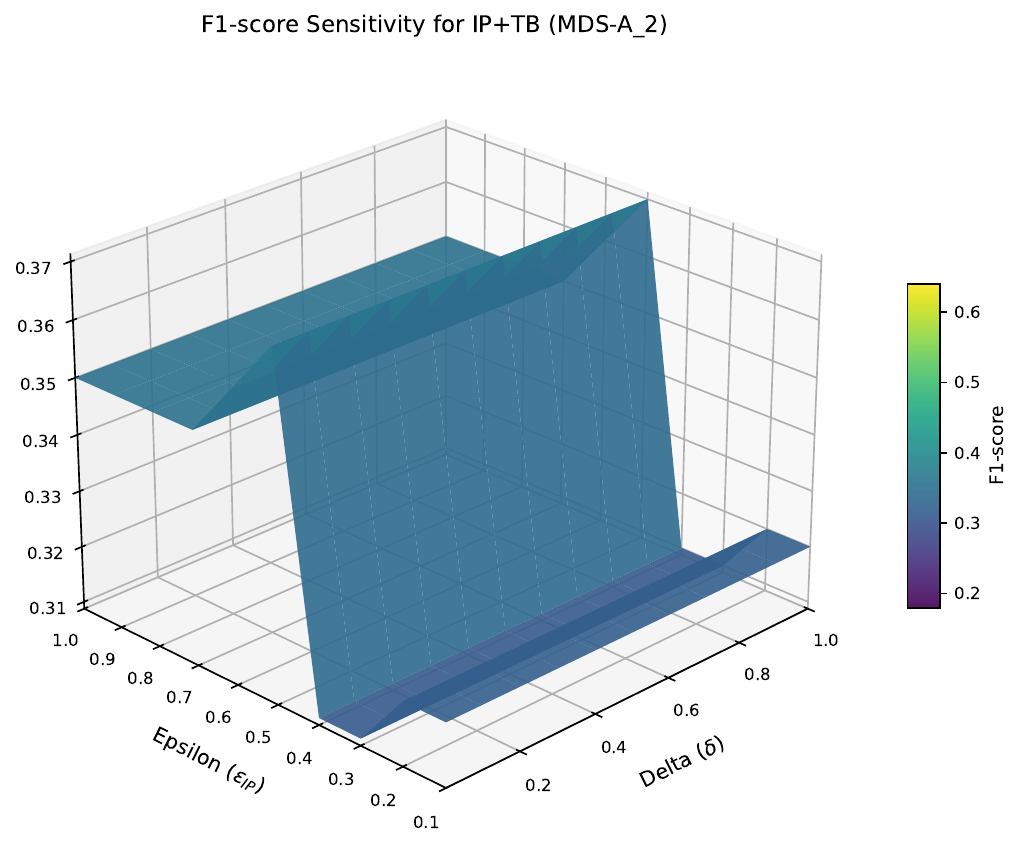}
    \end{subfigure}
    \hfill
    \begin{subfigure}[b]{0.19\textwidth}
        \centering
        \includegraphics[width=\linewidth]{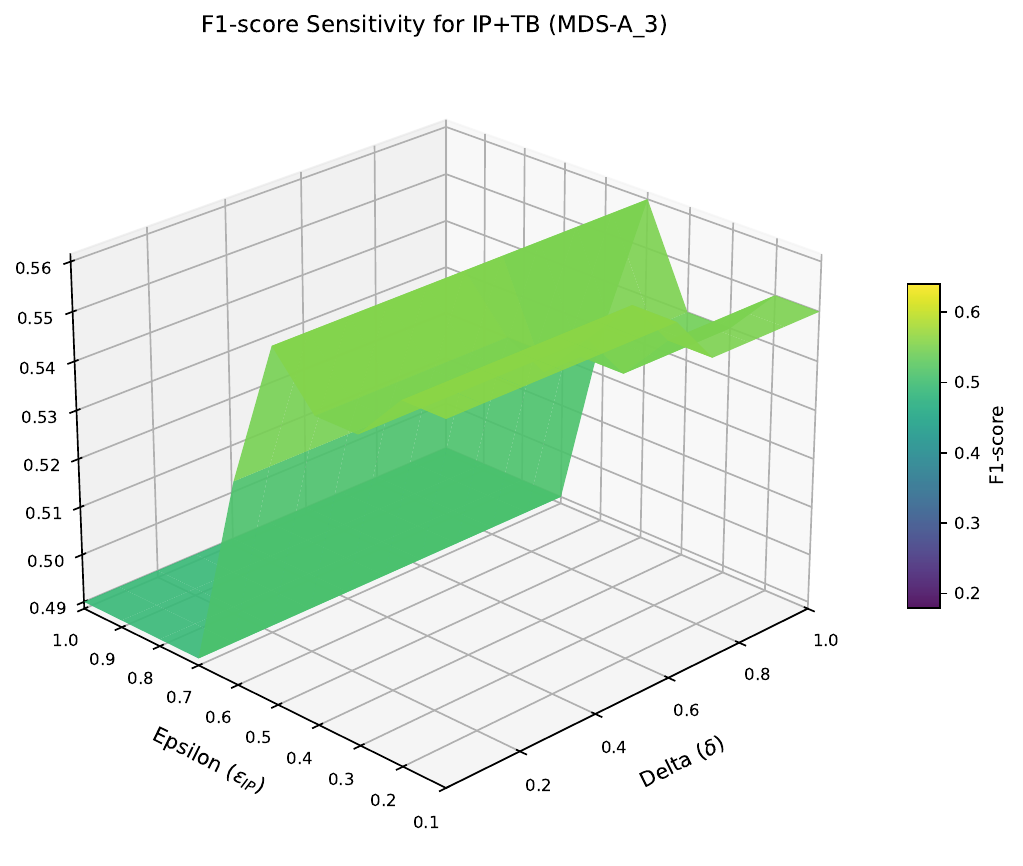}
    \end{subfigure}
    \hfill
    \begin{subfigure}[b]{0.19\textwidth}
        \centering
        \includegraphics[width=\linewidth]{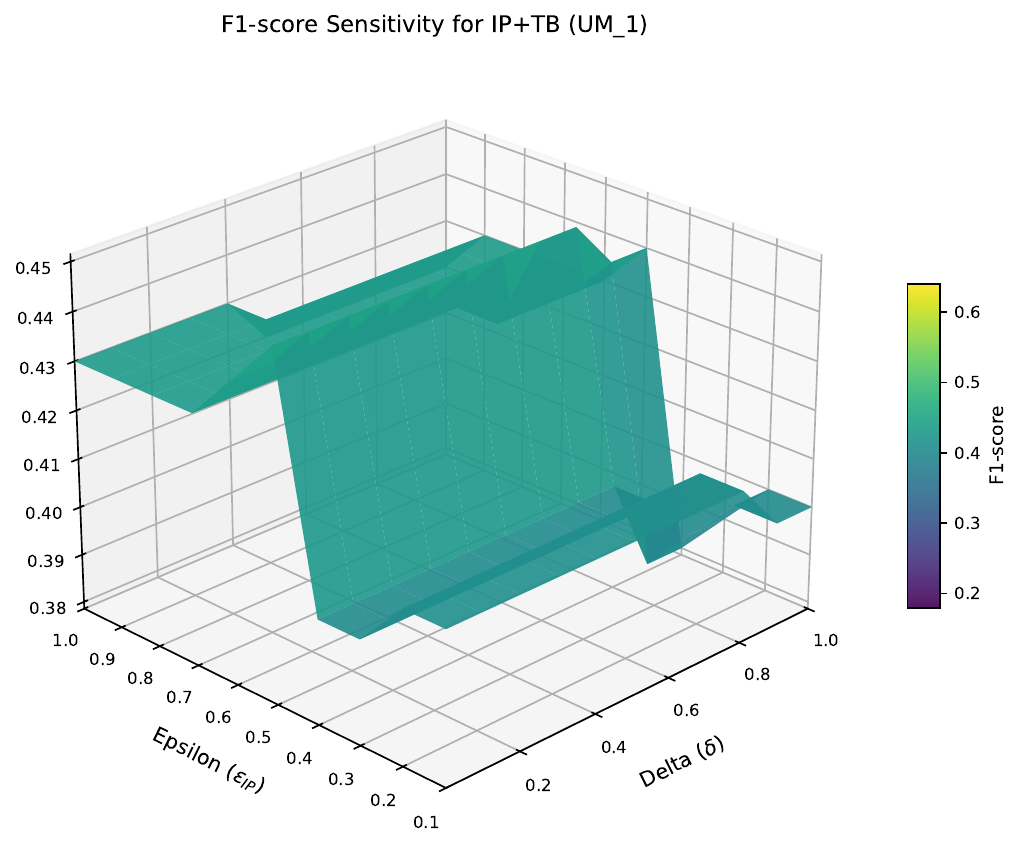}
    \end{subfigure}
    \hfill
    \begin{subfigure}[b]{0.19\textwidth}
        \centering
        \includegraphics[width=\linewidth]{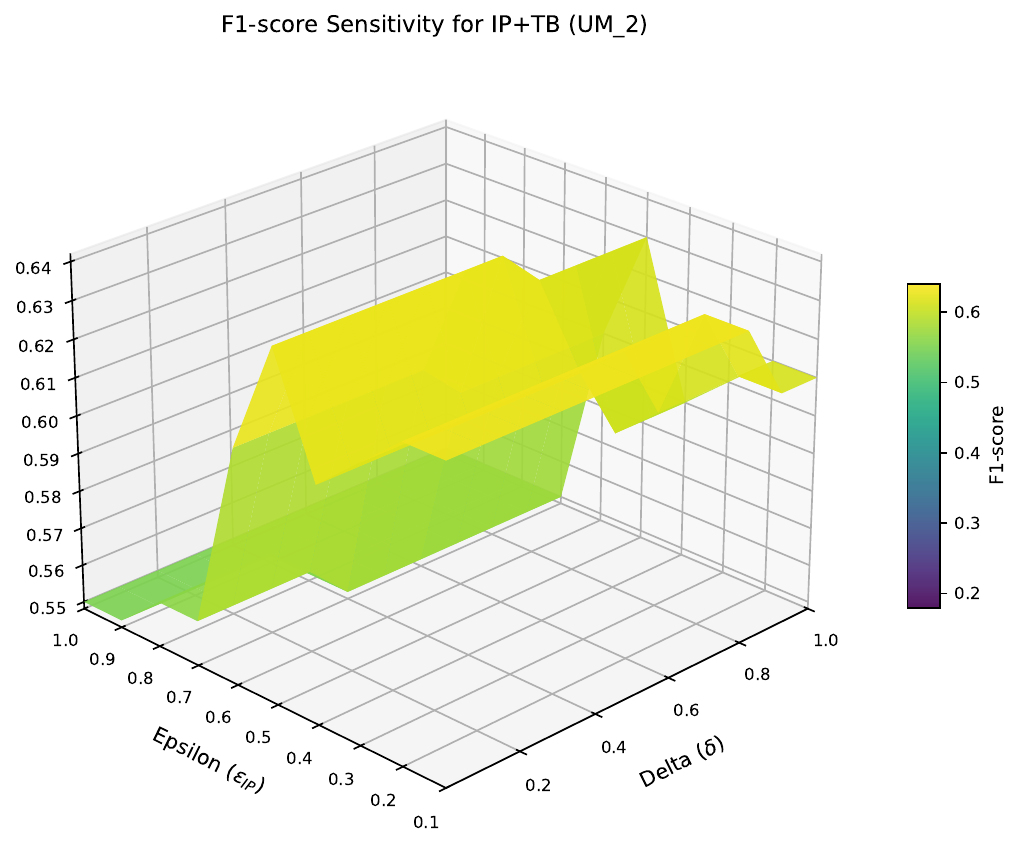}
    \end{subfigure}

    \vspace{\baselineskip}

    \begin{subfigure}[b]{0.19\textwidth}
        \centering
        \includegraphics[width=\linewidth]{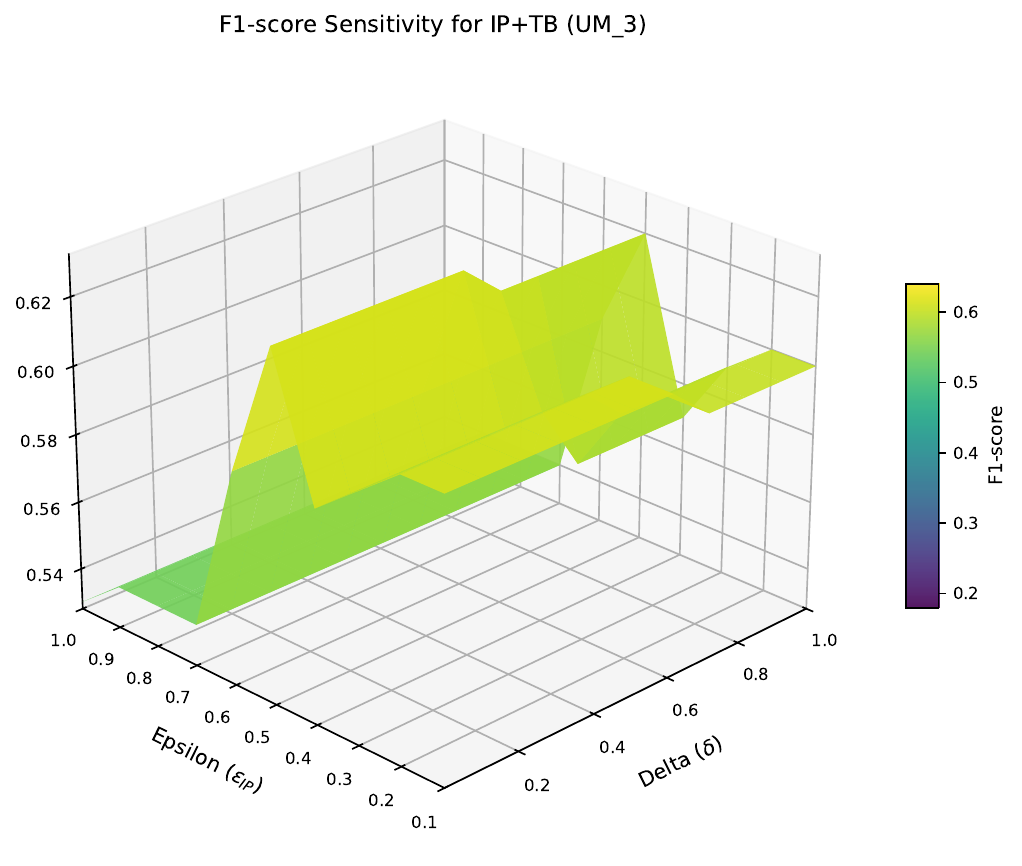}
    \end{subfigure}
    \hfill
    \begin{subfigure}[b]{0.19\textwidth}
        \centering
        \includegraphics[width=\linewidth]{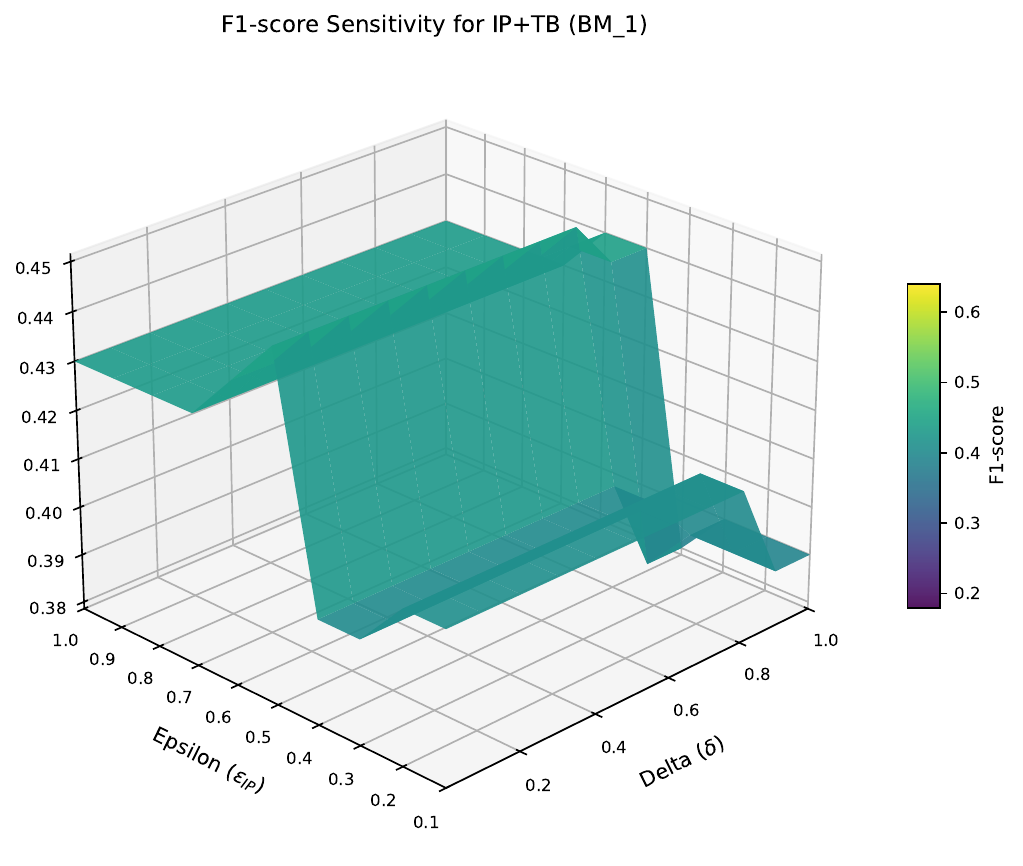}
    \end{subfigure}
    \hfill
    \begin{subfigure}[b]{0.19\textwidth}
        \centering
        \includegraphics[width=\linewidth]{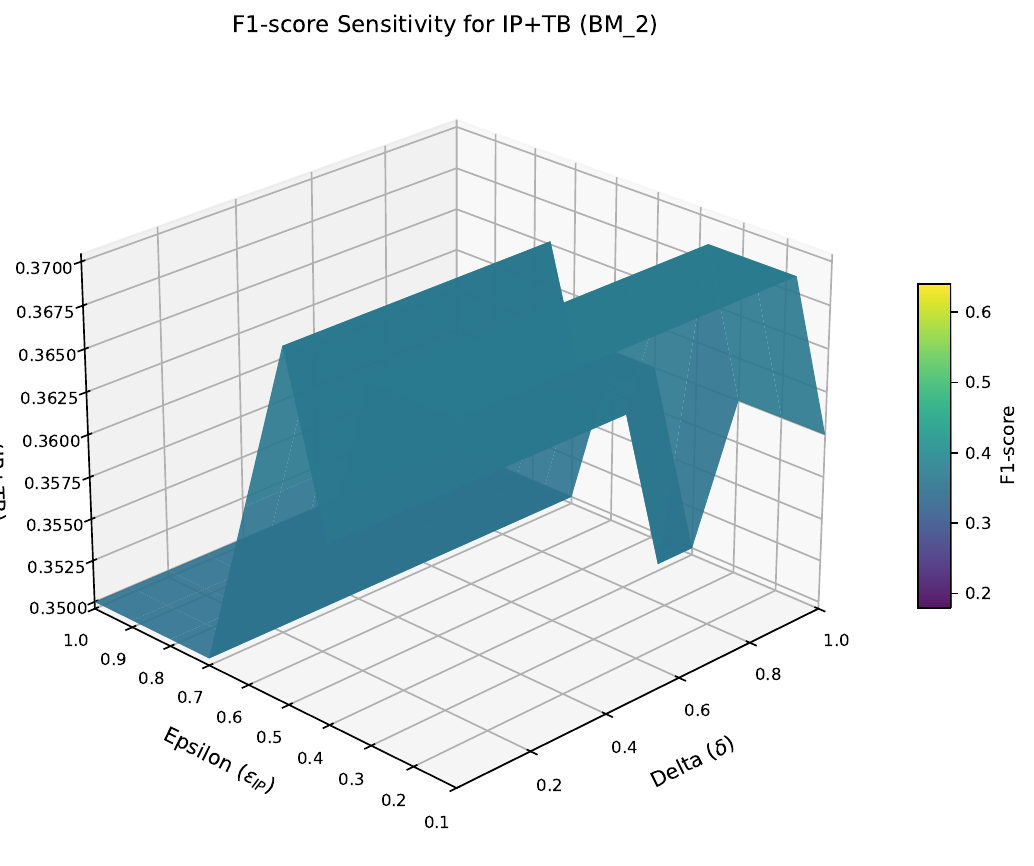}
    \end{subfigure}
    \hfill
    \begin{subfigure}[b]{0.19\textwidth}
        \centering
        \includegraphics[width=\linewidth]{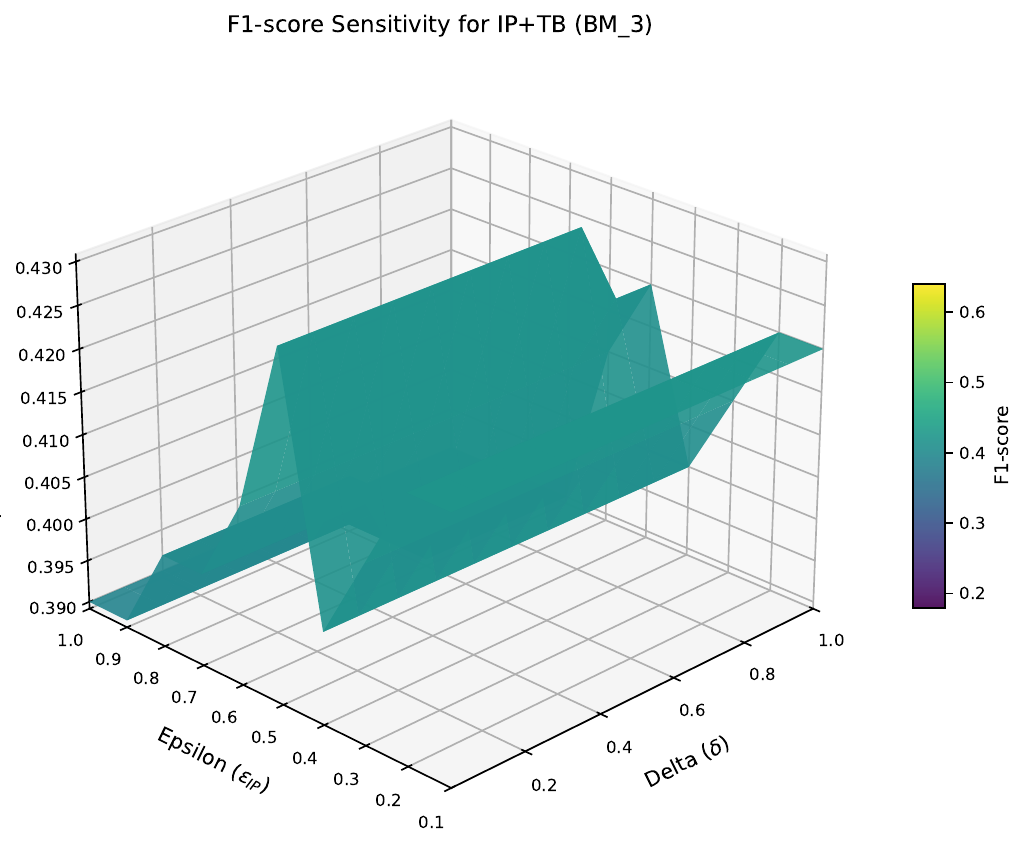}
    \end{subfigure}
    \hfill
    \begin{subfigure}[b]{0.19\textwidth}
        \centering
        \includegraphics[width=\linewidth]{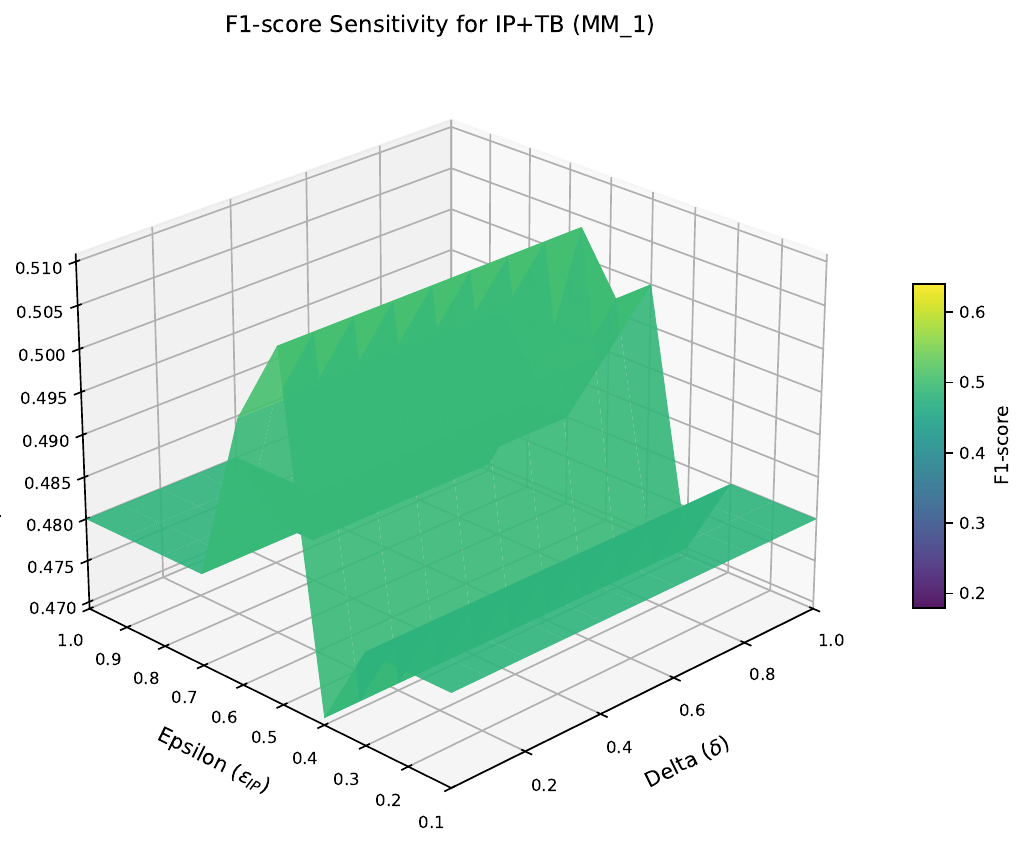}
    \end{subfigure}

    \vspace{\baselineskip}

    \begin{subfigure}[b]{0.19\textwidth}
        \centering
        \includegraphics[width=\linewidth]{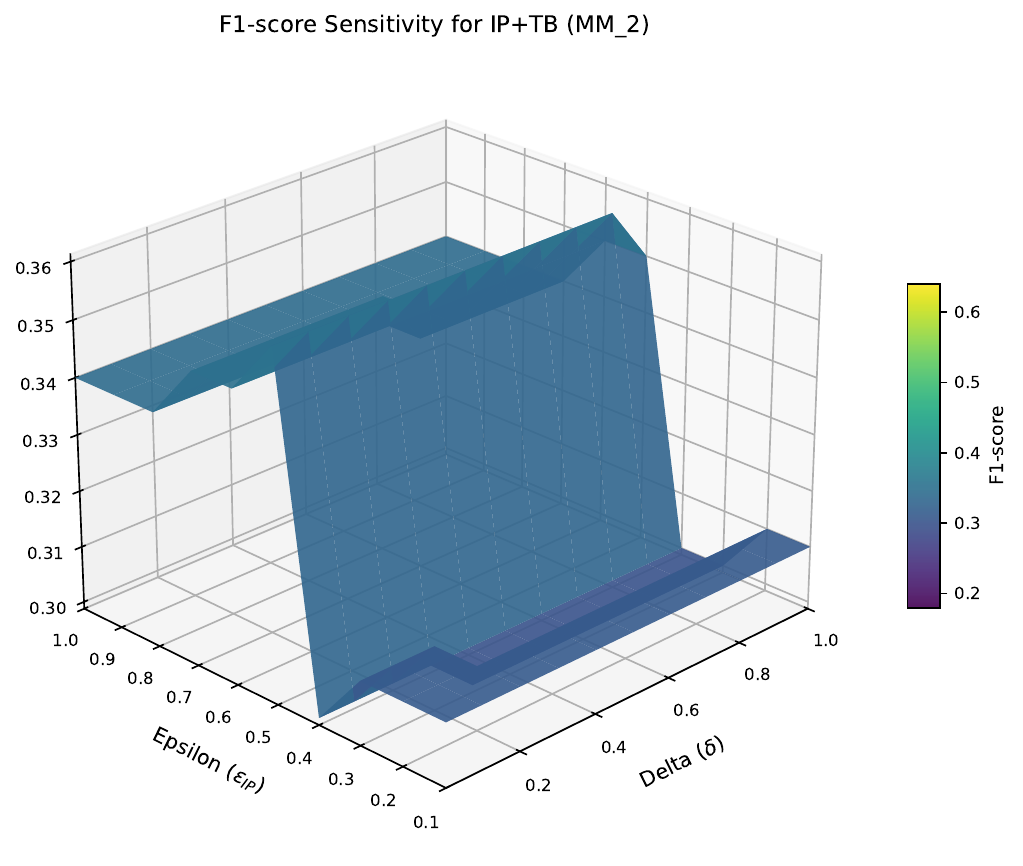}
    \end{subfigure}
    \hfill
    \begin{subfigure}[b]{0.19\textwidth}
        \centering
        \includegraphics[width=\linewidth]{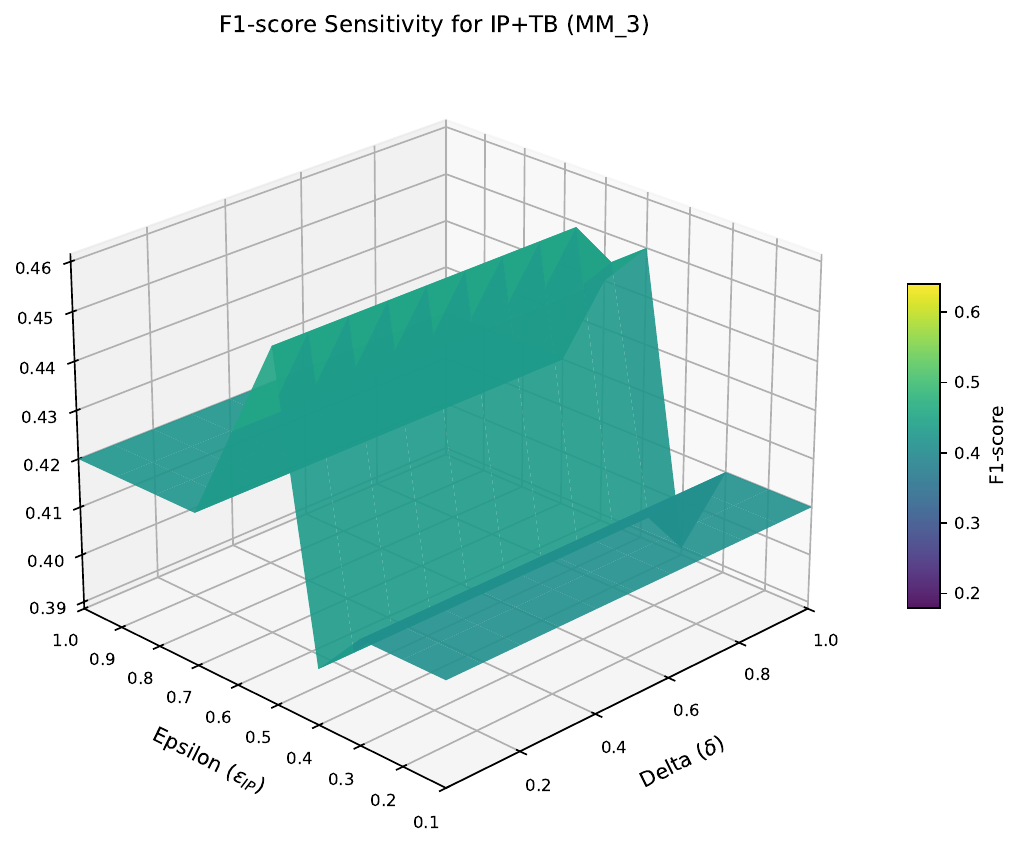}
    \end{subfigure}
    \hfill
    \begin{subfigure}[b]{0.19\textwidth}
        \centering
        \includegraphics[width=\linewidth]{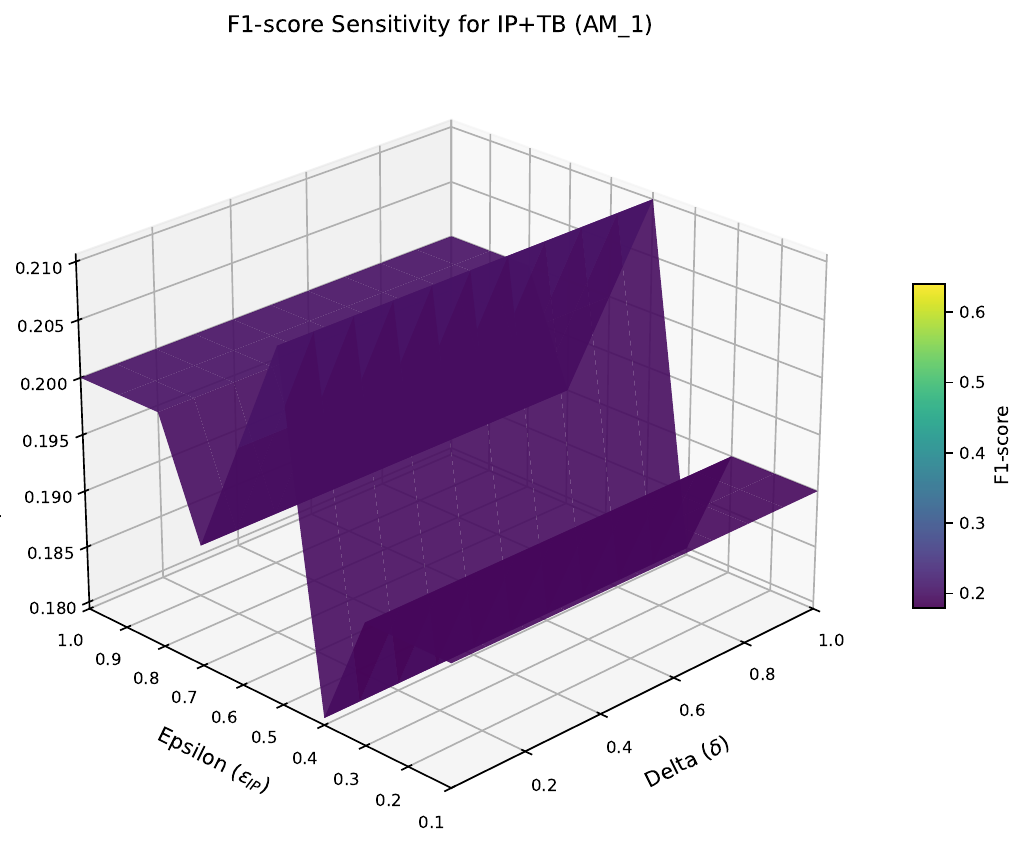}
    \end{subfigure}
    \hfill
    \begin{subfigure}[b]{0.19\textwidth}
        \centering
        \includegraphics[width=\linewidth]{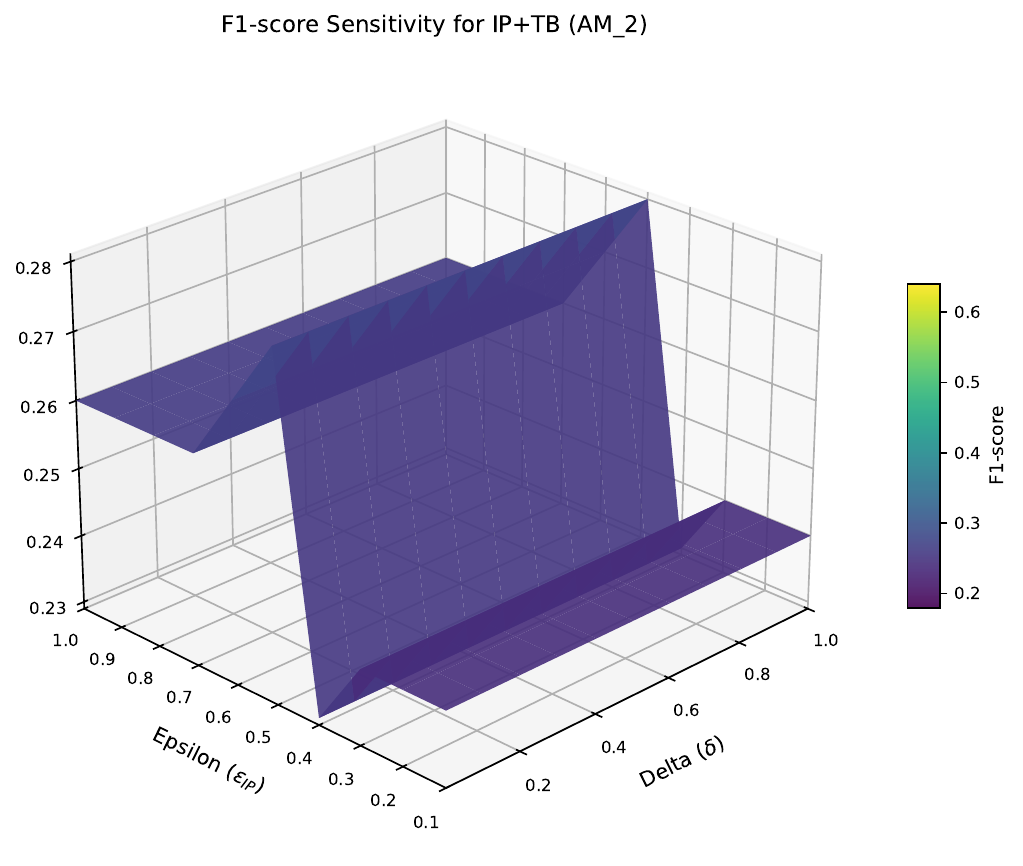}
    \end{subfigure}
    \hfill
    \begin{subfigure}[b]{0.19\textwidth}
        \centering
        \includegraphics[width=\linewidth]{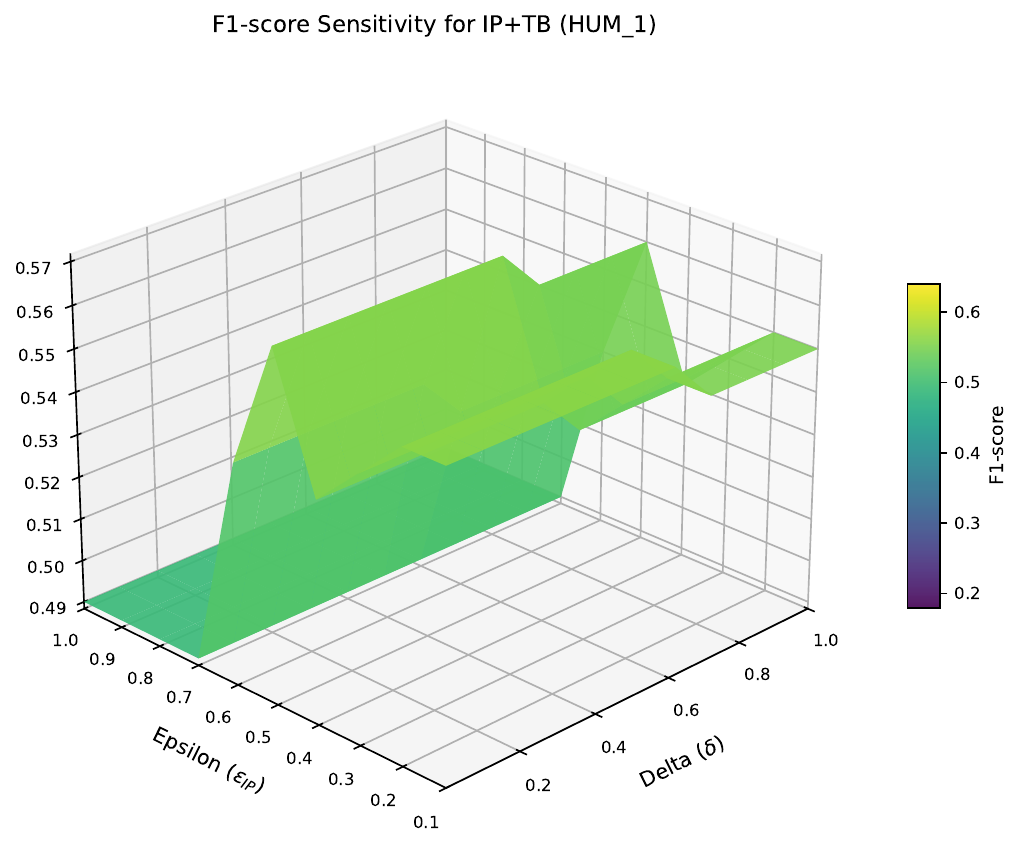}
    \end{subfigure}
    \caption{Detailed IP+TB Epsilon-Delta sensitivity analysis per test set for F1-score} 
    \label{fig:eps_delta_sens_f1}
\end{figure}

\begin{figure}[htbp!]
    \centering

    \begin{subfigure}[b]{0.19\textwidth}
        \centering
        \includegraphics[width=\linewidth]{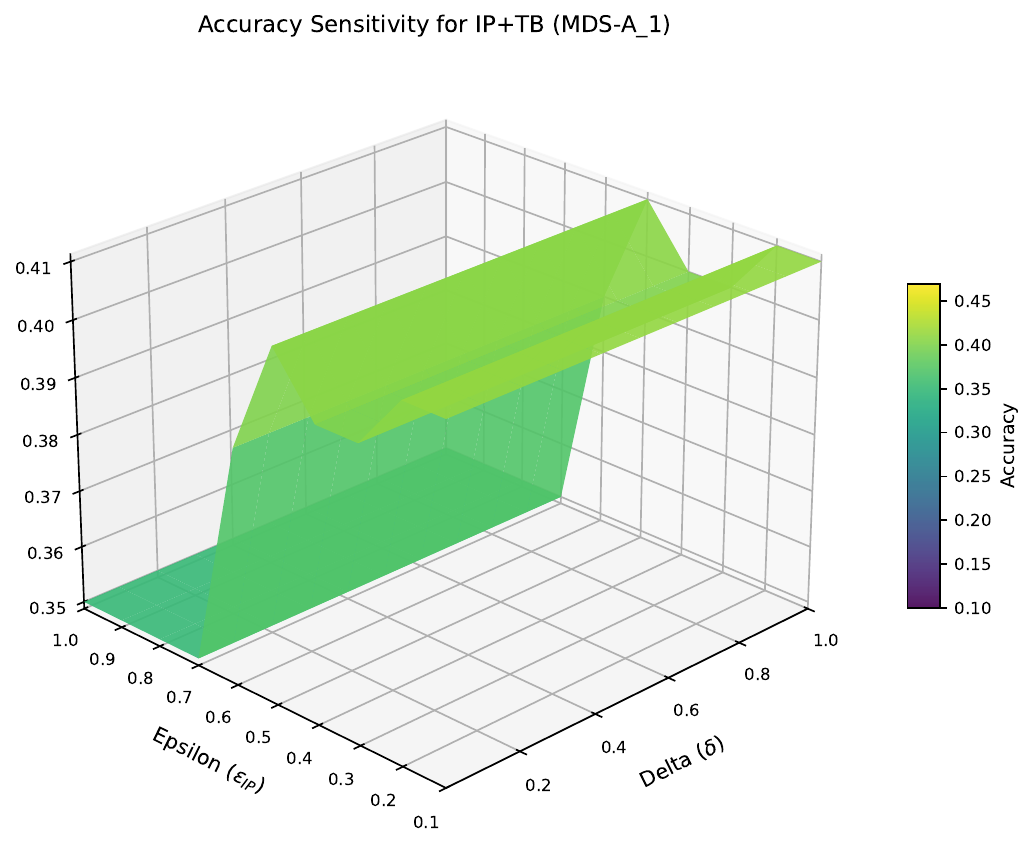}
    \end{subfigure}
    \hfill
    \begin{subfigure}[b]{0.19\textwidth}
        \centering
        \includegraphics[width=\linewidth]{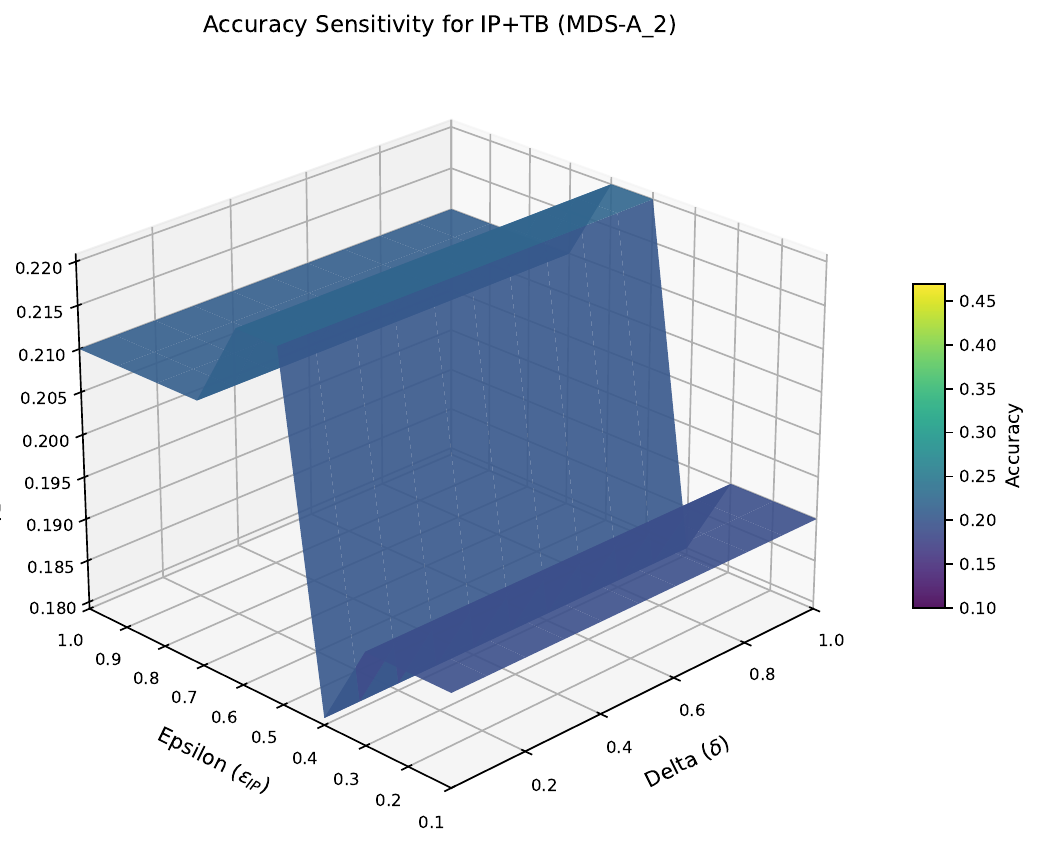}
    \end{subfigure}
    \hfill
    \begin{subfigure}[b]{0.19\textwidth}
        \centering
        \includegraphics[width=\linewidth]{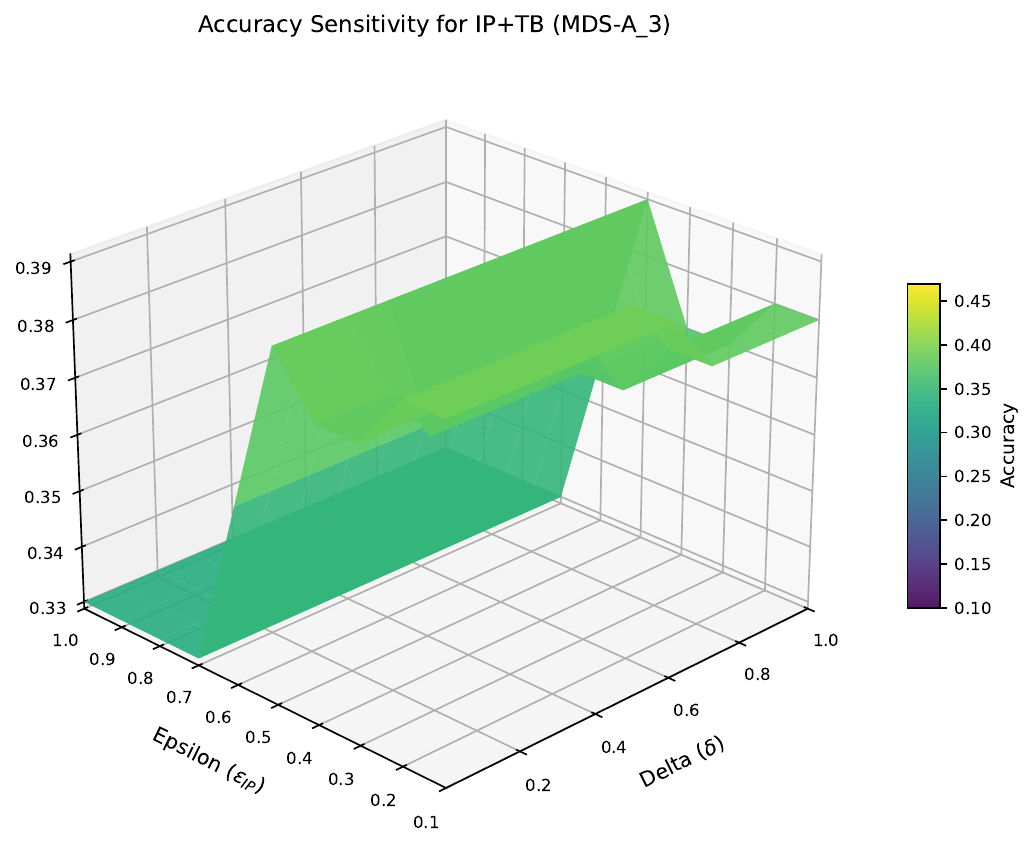}
    \end{subfigure}
    \hfill
    \begin{subfigure}[b]{0.19\textwidth}
        \centering
        \includegraphics[width=\linewidth]{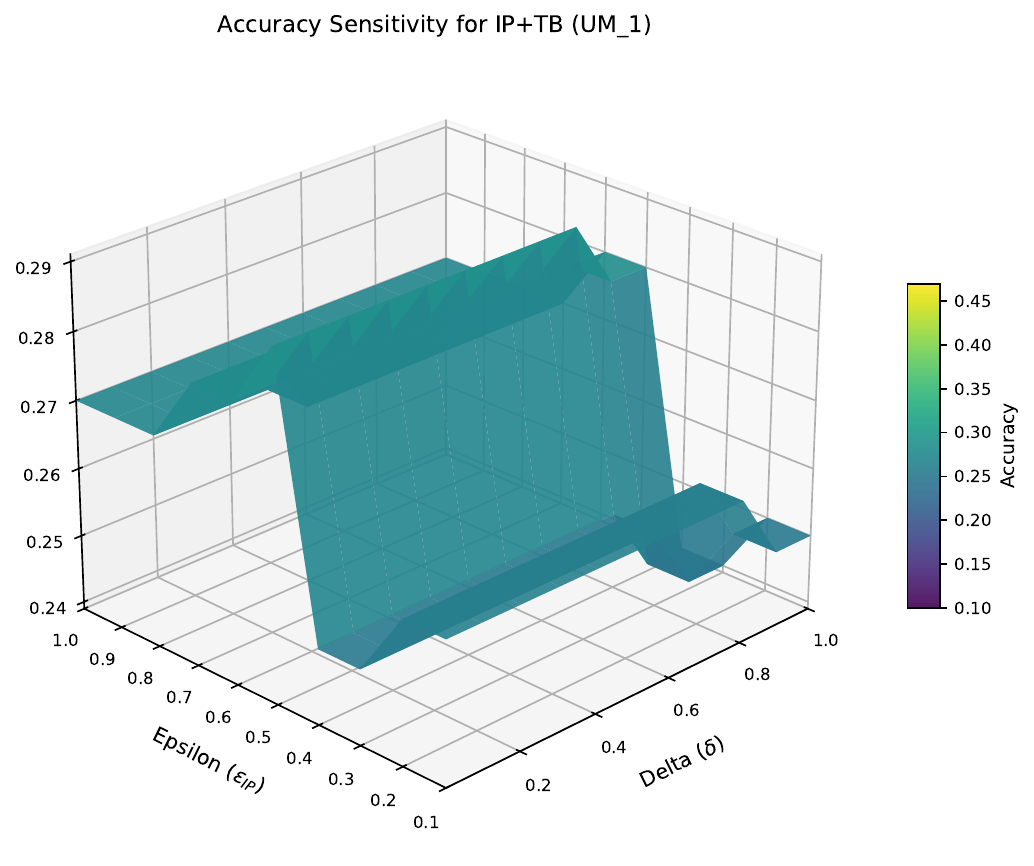}
    \end{subfigure}
    \hfill
    \begin{subfigure}[b]{0.19\textwidth}
        \centering
        \includegraphics[width=\linewidth]{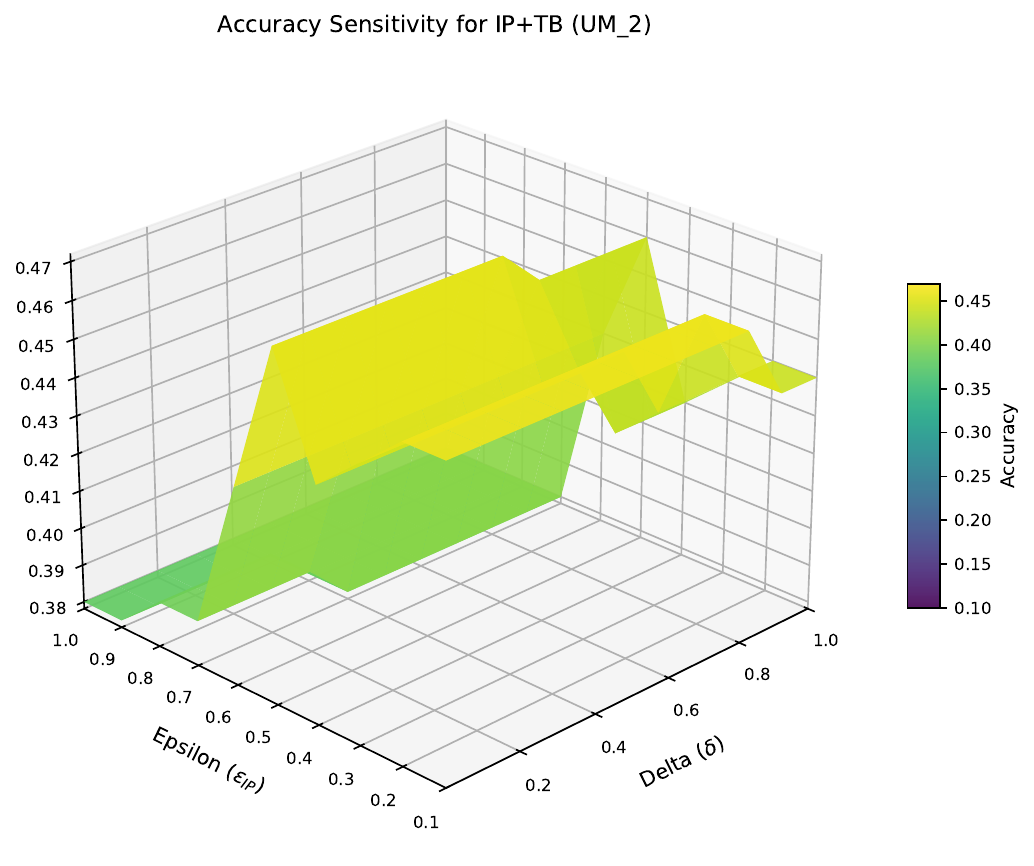}
    \end{subfigure}

    \vspace{\baselineskip}

    \begin{subfigure}[b]{0.19\textwidth}
        \centering
        \includegraphics[width=\linewidth]{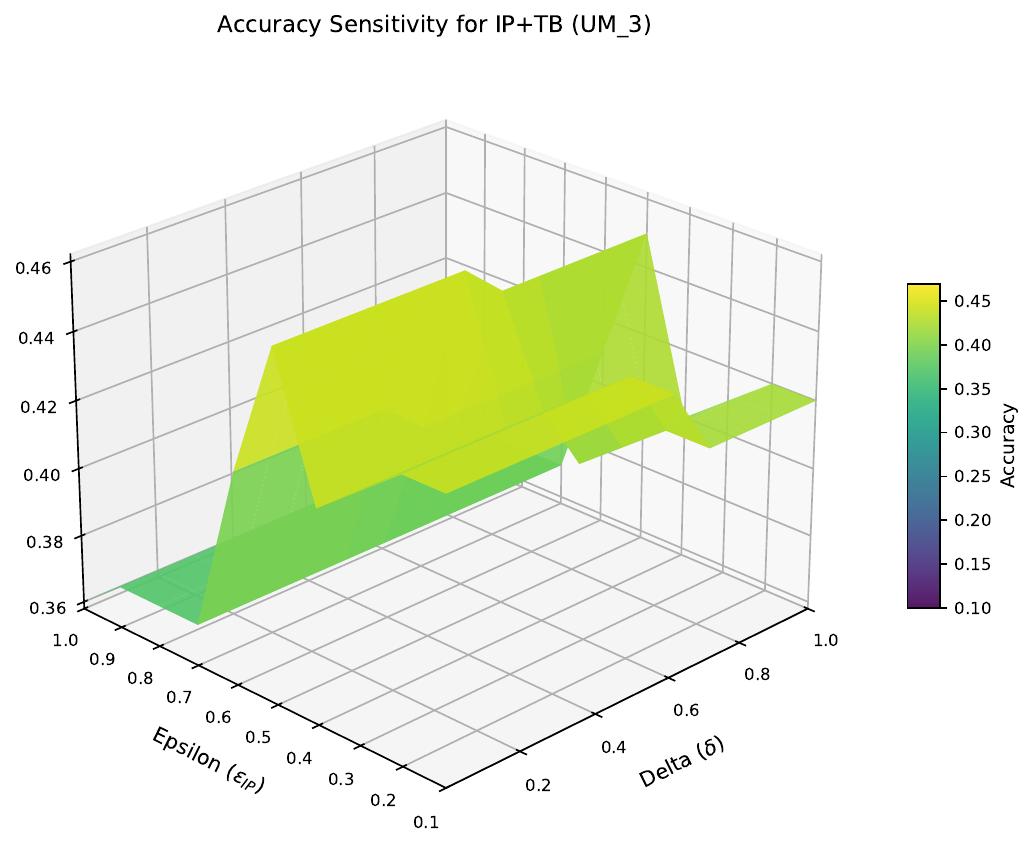}
    \end{subfigure}
    \hfill
    \begin{subfigure}[b]{0.19\textwidth}
        \centering
        \includegraphics[width=\linewidth]{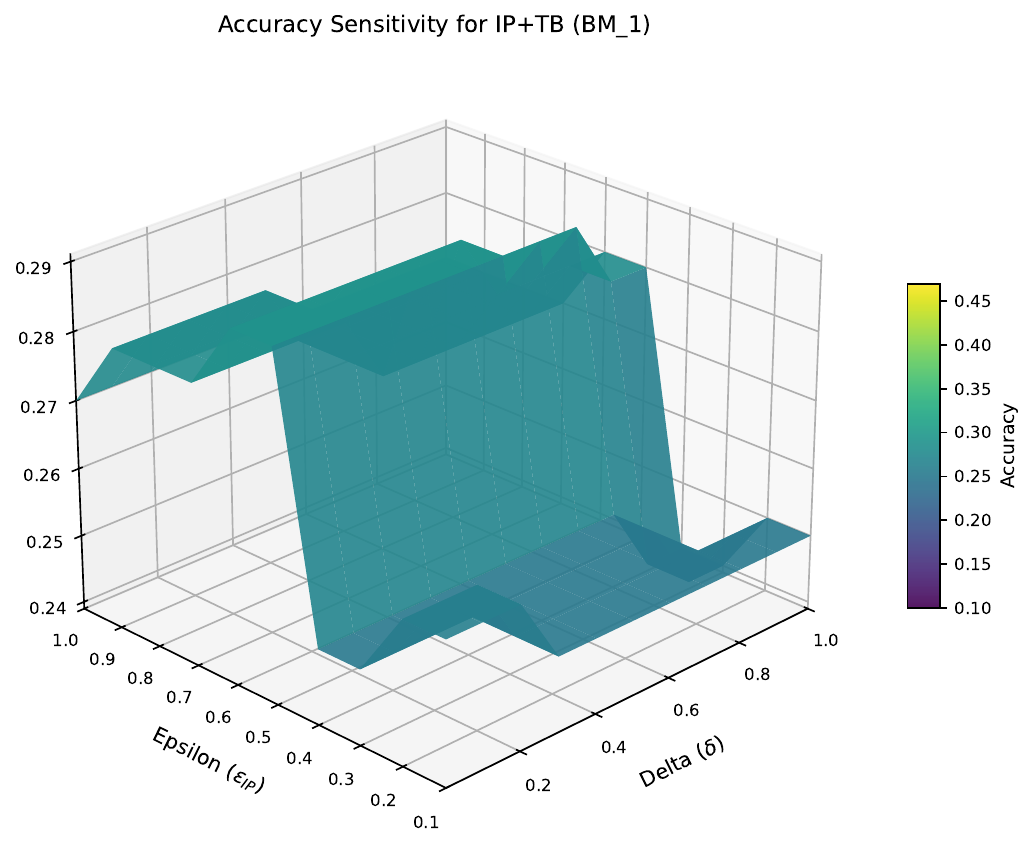}
    \end{subfigure}
    \hfill
    \begin{subfigure}[b]{0.19\textwidth}
        \centering
        \includegraphics[width=\linewidth]{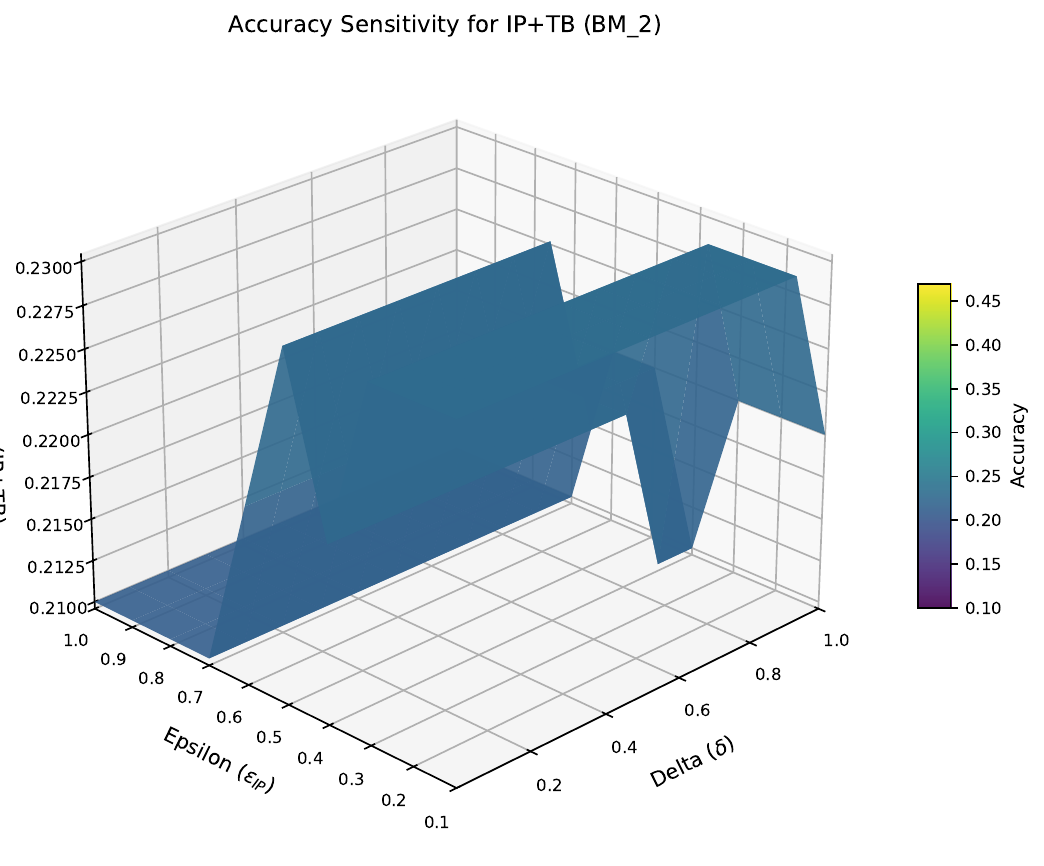}
    \end{subfigure}
    \hfill
    \begin{subfigure}[b]{0.19\textwidth}
        \centering
        \includegraphics[width=\linewidth]{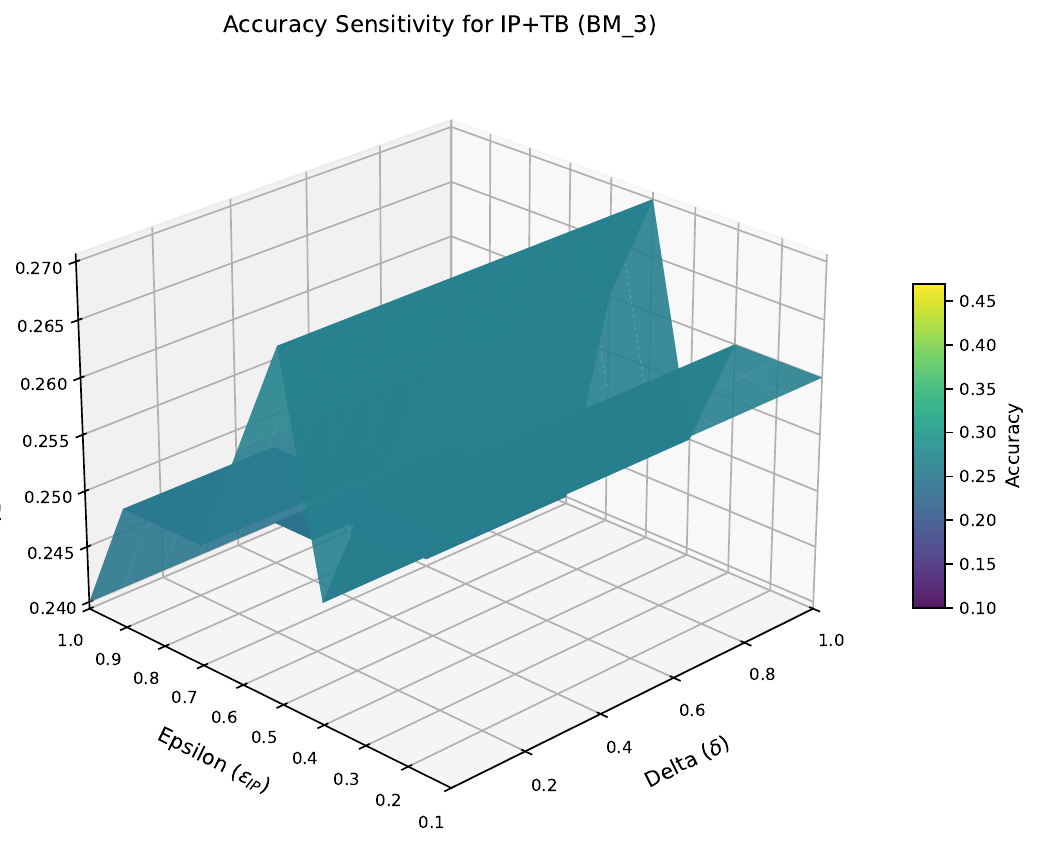}
    \end{subfigure}
    \hfill
    \begin{subfigure}[b]{0.19\textwidth}
        \centering
        \includegraphics[width=\linewidth]{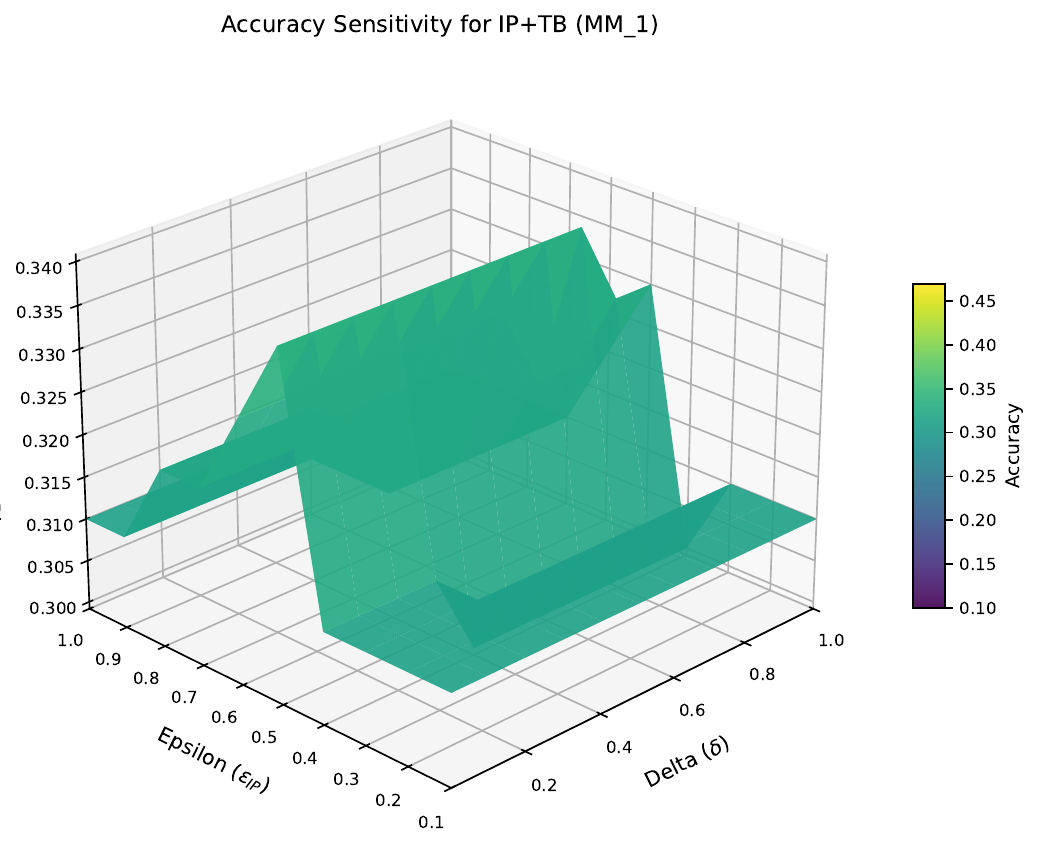}
    \end{subfigure}

    \vspace{\baselineskip}

    \begin{subfigure}[b]{0.19\textwidth}
        \centering
        \includegraphics[width=\linewidth]{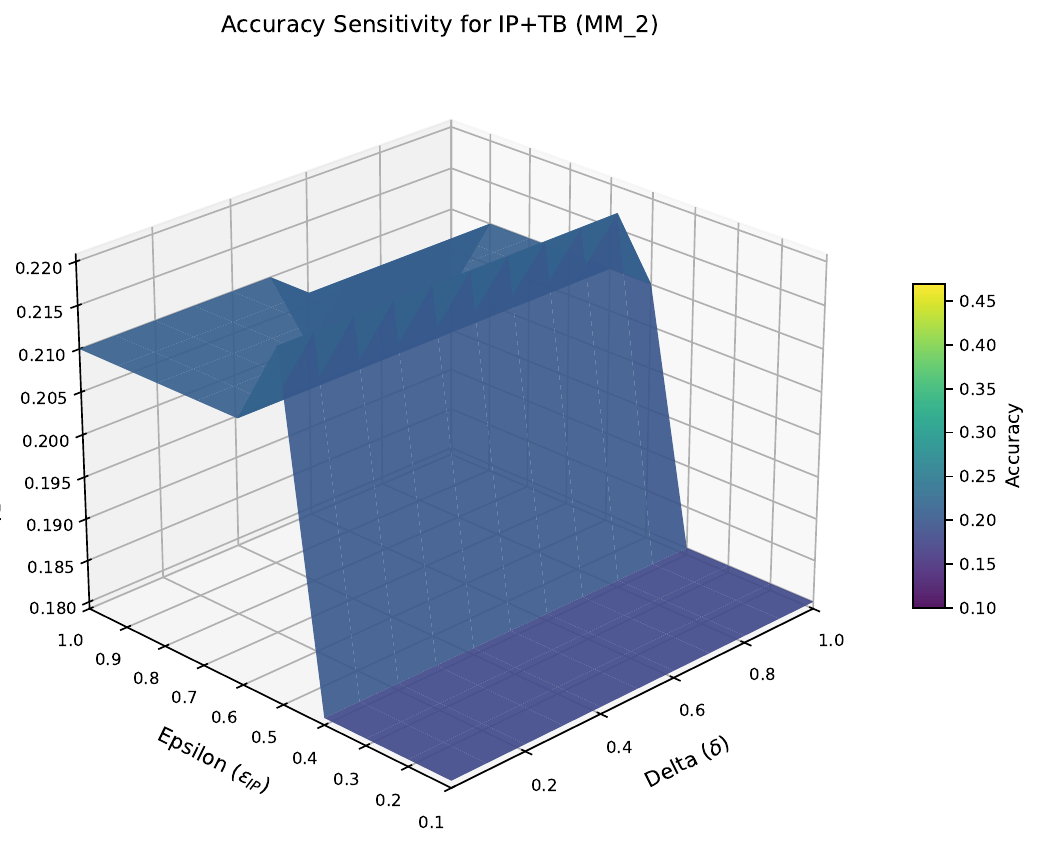}
    \end{subfigure}
    \hfill
    \begin{subfigure}[b]{0.19\textwidth}
        \centering
        \includegraphics[width=\linewidth]{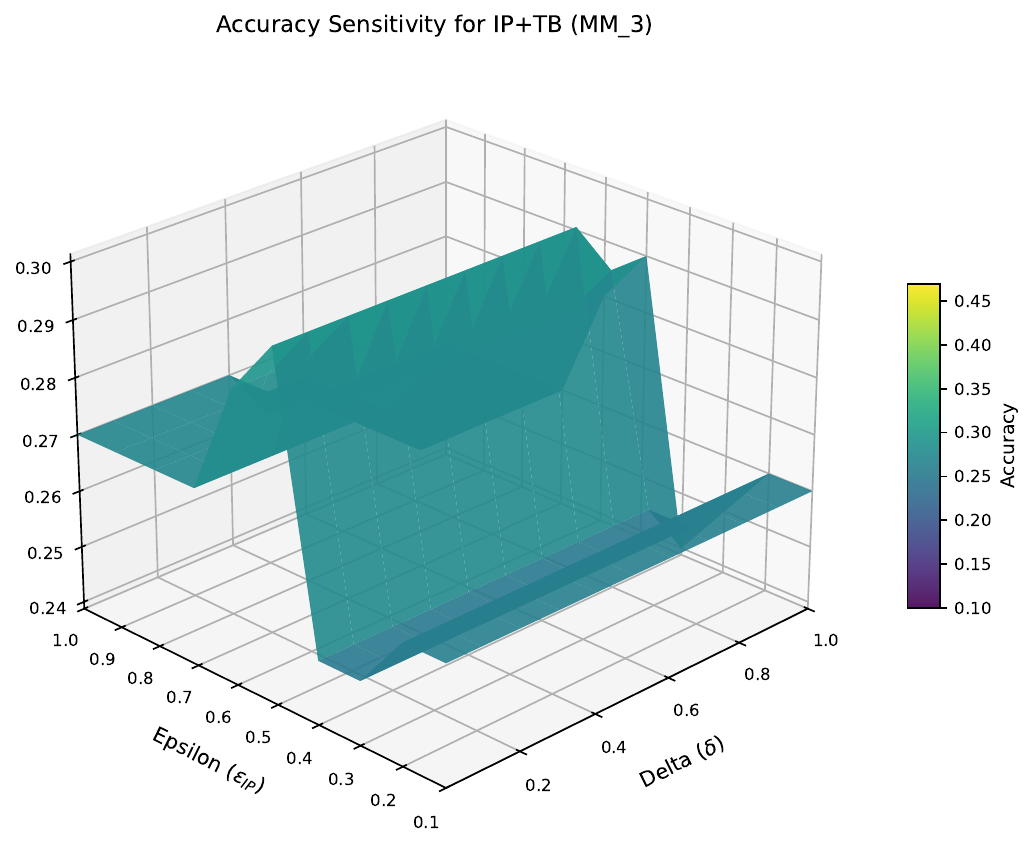}
    \end{subfigure}
    \hfill
    \begin{subfigure}[b]{0.19\textwidth}
        \centering
        \includegraphics[width=\linewidth]{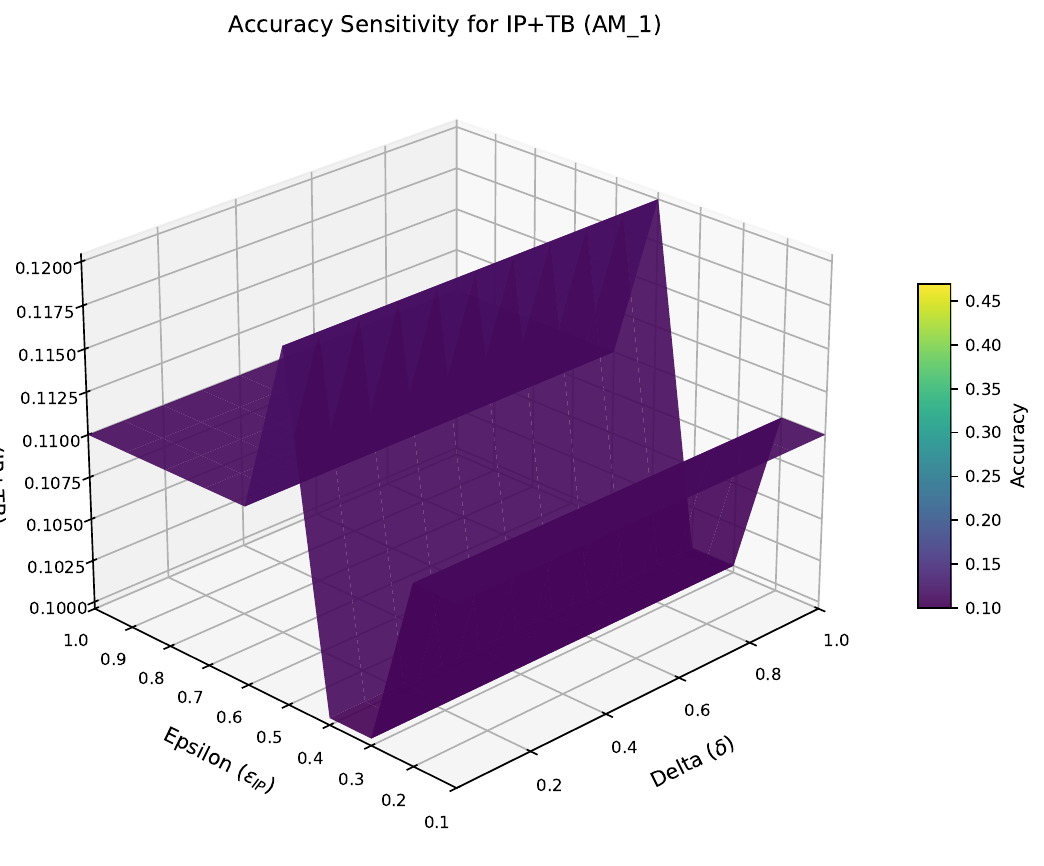}
    \end{subfigure}
    \hfill
    \begin{subfigure}[b]{0.19\textwidth}
        \centering
        \includegraphics[width=\linewidth]{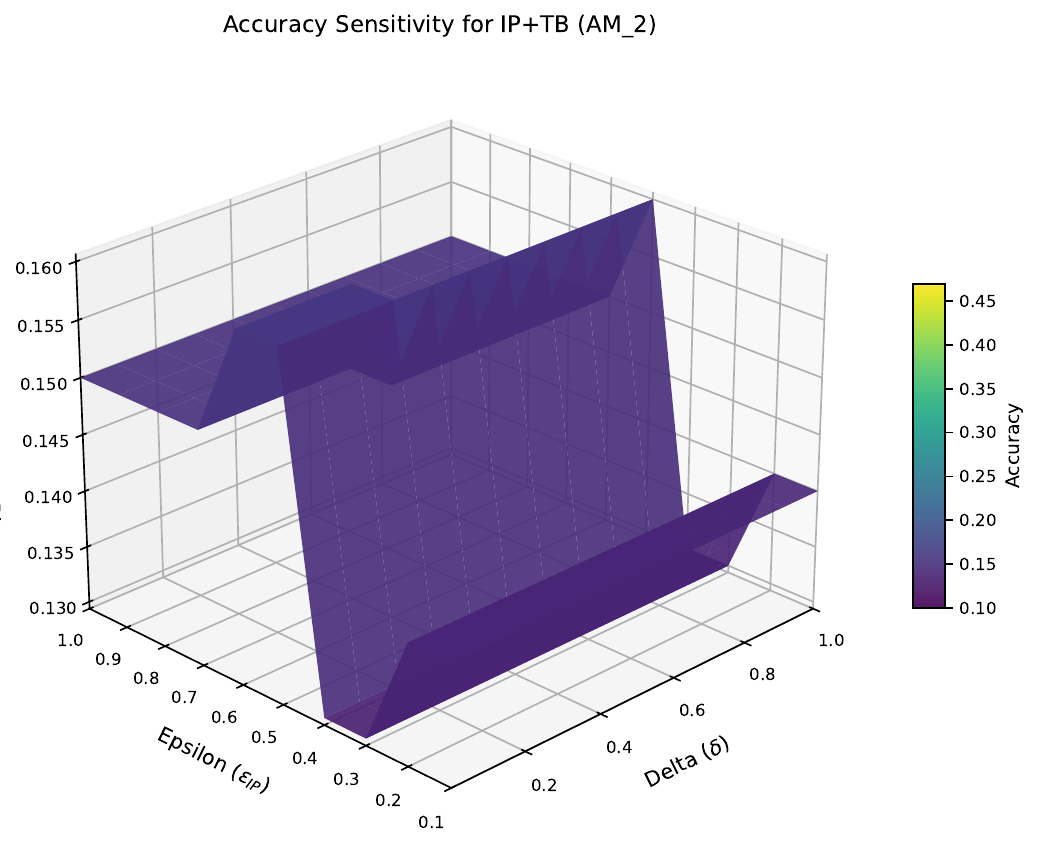}
    \end{subfigure}
    \hfill
    \begin{subfigure}[b]{0.19\textwidth}
        \centering
        \includegraphics[width=\linewidth]{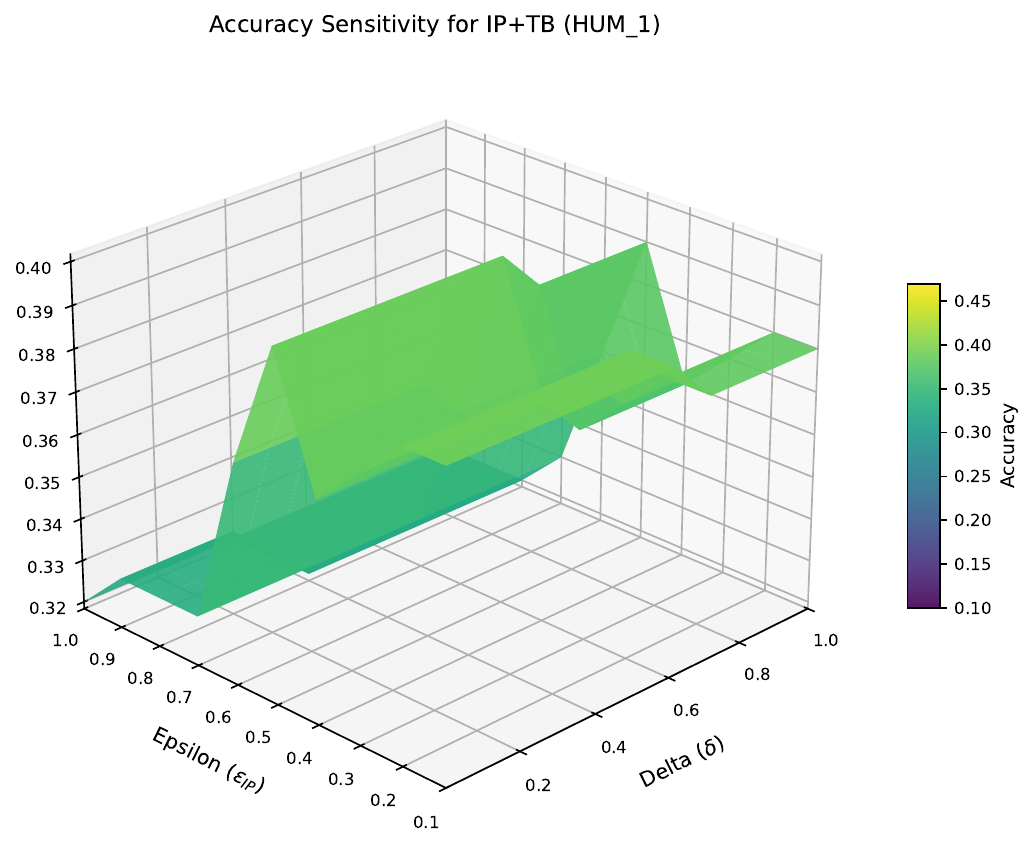}
    \end{subfigure}
    \caption{Detailed IP+TB Epsilon-Delta sensitivity analysis per test set for Accuracy} 
    \label{fig:eps_delta_sens_accuracy}
\end{figure}

This section presents the sensitivity of the Integer Programming with Tie-Breaker (IP+TB) method to both the maximum inconsistency threshold $\delta$ and its internal optimization parameter $\epsilon_{IP}$. For each of the 15 test datasets, two surface plots are provided: one for f1-score (Figura~\ref{fig:eps_delta_sens_f1}) and one for Accuracy (Figura~\ref{fig:eps_delta_sens_accuracy}). These detailed plots support the analysis presented for a representative dataset in Figure 5 (bottom panels) of the main paper.

\clearpage
\section{Object Detection Models Hyperparameters and IP Solver}
\label{sec:model_hyperparameters}
\subsection{Model Hyperparameters}
This subsection shows the hyperparameters used to train the object-detection models including the learning rate, batch size, and number of training epochs.

\begin{table}[h!]
\centering
\begin{tabular}{|l|l|}
\hline
\textbf{Hyperparameter} & \textbf{Value} \\
\hline
\texttt{Number of Training Epoch} & 500 \\
\texttt{Batch Size} & 64 \\
\texttt{Learning Rate} & 5e-5 \\
\texttt{Learning Rate Scheduler Type} & Cosine \\
\texttt{Weight Decay} & 1e-4 \\
\texttt{Maximum Gradient Norm} & 0.01 \\
\texttt{Metric for Best Model Selection} & mAP \\
\hline
\end{tabular}
\caption{Training hyperparameters used for object-detection model training}
\label{tab:training-hyperparameters}
\end{table}

\subsection{IP Optimization Solver}
The Integer Programming (IP) models for our IP+TB approach were implemented in Python using the 
PuLP library.\footnote{PuLP is an open-source linear programming package for Python. For documentation, see \url{https://coin-or.github.io/pulp/}.} PuLP serves as a high-level modeling interface that allows for the algebraic 
representation of linear and integer programming problems. For solving the IP instances formulated 
with PuLP, we utilized the COIN-OR Branch and Cut (CBC) solver,\footnote{The COIN-OR Branch and 
Cut (CBC) solver is an open-source mixed integer programming solver provided by the COIN-OR 
Foundation. More information can be found at \url{https://projects.coin-or.org/Cbc} or 
\url{https://github.com/coin-or/Cbc}.} which is invoked via PuLP. The CBC solver is a robust tool 
for tackling mixed integer linear programs.

\section{URLs of the datasets and the anonymized code.}
\begin{itemize}
    \item Test Sets: The full test set used is an extension of the dataset presented in Ngu et al. 2025. This extension consists of new test sets with additional variations in weather conditions (shown in Figure 2 of the main paper).
    \item Code: github.com/lab-v2/EDCR\_PyReason\_AirSim
\end{itemize}
\end{document}